\documentclass[nohyperref]{article}

\usepackage{microtype}
\usepackage{graphicx}
\usepackage{subfigure}
\usepackage{booktabs} %
\usepackage{multirow}
\usepackage{hyperref}

\usepackage[accepted]{icml2023}

\usepackage{amsmath}
\usepackage{amssymb}
\usepackage{mathtools}
\usepackage{amsthm}
\usepackage{bm}
\usepackage{enumitem}
\usepackage[title]{appendix}
\usepackage[capitalize,noabbrev]{cleveref}

\theoremstyle{plain}

\theoremstyle{definition}

\theoremstyle{remark}

\usepackage{xspace}
\newcommand{\figref}[1]{\figurename~\ref{#1}}
\newcommand{\tblref}[1]{Table~\ref{#1}}
\newcommand{\secref}[1]{Section~\ref{#1}}

\newcommand{\appref}[1]{Appendix~\ref{#1}}
\newcommand{\actor}{\textbf{Actor}\xspace}
\newcommand{\plearner}{\textbf{P-learner}\xspace}
\newcommand{\vlearner}{\textbf{V-learner}\xspace}
\newcommand{\shadow}{\textit{Shadow Hand}\xspace}
\newcommand{\allegro}{\textit{Allegro Hand}\xspace}
\newcommand{\franka}{\textit{Franka Cube Stacking}\xspace}
\newcommand{\anymal}{\textit{ANYmal}\xspace}
\newcommand{\ant}{\textit{Ant}\xspace}
\newcommand{\humanoid}{\textit{Humanoid}\xspace}
\newcommand{\ball}{\textit{Ball Balancing}\xspace}
\newcommand{\dcalw}{\textit{DClaw Hand}\xspace}
\newcommand{\method}{\textbf{PQL}\xspace}

\usepackage[textsize=tiny]{todonotes}
\hypersetup{
breaklinks=true,colorlinks,citecolor=blue
}

\icmltitlerunning{Parallel $Q$-Learning: Scaling Off-policy Reinforcement Learning under Massively Parallel Simulation}

\begin{document}

\twocolumn[
\icmltitle{Parallel $Q$-Learning: Scaling Off-policy Reinforcement Learning under Massively Parallel Simulation}

\icmlsetsymbol{equal}{*}

\begin{icmlauthorlist}
\icmlauthor{Zechu Li}{equal,mit}
\icmlauthor{Tao Chen}{equal,mit}
\icmlauthor{Zhang-Wei Hong}{mit}
\icmlauthor{Anurag Ajay}{mit}
\icmlauthor{Pulkit Agrawal}{mit}
\end{icmlauthorlist}

\icmlaffiliation{mit}{Improbable AI Lab, Massachusetts Institute of Technology, Cambridge, USA}

\icmlcorrespondingauthor{Zechu Li}{zechu@mit.edu}
\icmlcorrespondingauthor{Tao Chen}{taochen@mit.edu}
\icmlcorrespondingauthor{Pulkit Agrawal}{pulkitag@mit.edu}

\icmlkeywords{Machine Learning, ICML}

\vskip 0.3in
]

\printAffiliationsAndNotice{\icmlEqualContribution} %

\begin{abstract}
Reinforcement learning is time-consuming for complex tasks due to the need for large amounts of training data. Recent advances in GPU-based simulation, such as Isaac Gym, have sped up data collection thousands of times on a commodity GPU. Most prior works used on-policy methods like PPO due to their simplicity and ease of scaling. Off-policy methods are more data efficient but challenging to scale, resulting in a longer wall-clock training time. This paper presents a Parallel $Q$-Learning (PQL) scheme that outperforms PPO in wall-clock time while maintaining superior sample efficiency of off-policy learning. PQL achieves this by parallelizing data collection, policy learning, and value learning. Different from prior works on distributed off-policy learning, such as Apex, our scheme is designed specifically for massively parallel GPU-based simulation and optimized to work on a single workstation. In experiments, we demonstrate that $Q$-learning can be scaled to \textit{tens of thousands of parallel environments} and investigate important factors affecting learning speed.  The code is available at \href{https://github.com/Improbable-AI/pql}{https://github.com/Improbable-AI/pql}.
\end{abstract}

\section{Introduction}
\label{sec:intro}
Reinforcement learning (RL) has achieved impressive results on many real-world problems. A primary challenge in using RL is the need for large amounts of real-world data. There are two main strategies to tackle this problem. One is to improve the sample efficiency of RL algorithms~\citep{mnih2015human, lillicrap2015continuous} to make better use of available data. The other is to reduce the need for real-world data collection by training policies in simulation and deploying them in the real world~\citep{hwangbo2019learning, andrychowicz2020learning, margolis2022rapid, miki2022learning, chen2022visual}. In sim-to-real pipelines, the training wall-clock time matters more than the sample efficiency --- faster training can speed up the experiment cycle and unlock the potential for addressing a broader range of complex problems. 

The community has widely recognized the need for faster training, leading to the development of several distributed frameworks~\citep{horgan2018distributed, espeholt2018impala}. However, these frameworks usually operate at a \textit{server scale} that requires hundreds or thousands of computers in a cluster, making them impractical for many researchers and practitioners. In these frameworks, most computers run multiple simulator instances in parallel to speed up data collection. Recent advances in GPU-based simulation, such as Isaac Gym~\citep{makoviychuk2021isaac}, mitigated the need for a large cluster by enabling the parallel simulation of \textit{tens of thousands} of environments on a single GPU. A natural question is: what RL algorithm achieves the best wall-clock time when training uses massively parallel simulation on GPUs? Many prior works~\citep{allshire2021transferring, rudin2022learning, chen2022system} use on-policy algorithms like PPO~\citep{schulman2017proximal} for training in Isaac Gym due to its simplicity and easy-to-scale nature. 

It is known that on-policy methods have lower data efficiency than off-policy methods. Intuitively, by virtue of requiring less data than on-policy algorithms, off-policy algorithms ($Q$-learning methods, in particular) should reduce the wall-clock time of training. 
However, better sample efficiency will not lead to shorter training time if the algorithm cannot efficiently use parallel environments. 
Some prior works \cite{nair2015massively,horgan2018distributed} have developed distributed frameworks for off-policy methods to leverage parallel environments. However, these frameworks have only shown successful scaling with hundreds of parallel environments (for example, a maximum of 256 environments in \cite{horgan2018distributed}). Now that GPU-based simulation enables \textit{tens of thousands} of parallel environments on a single GPU, it remains unclear whether off-policy methods can work efficiently in this case. For instance, if there are $10,000$ parallel environments and we still use the typical replay buffer capacity (say $1$M samples), the entire replay buffer is refreshed every $100$ environment steps, making the data in the replay buffer more like the data collected from an on-policy method. Do off-policy methods still retain their data efficiency in this scenario? Increasing the replay buffer capacity is not always an option due to the memory size limits of the hardware. 

In this work, we investigate how to scale up $Q$-learning to tens of thousands of environments. We present our approach, \textbf{P}arallel $\mathbf{Q}$-\textbf{L}earning (\method), which can be deployed on a workstation. The learning speed in PQL is boosted by parallelizing the data collection, policy function learning, and value function learning on a single workstation. This allows for collecting more simulation data and updating value/policy functions more times in a given time window, leading to an improvement in the training wall clock time. Achieving such parallelization would be non-trivial for on-policy algorithms, as the policy update requires on-policy interaction data, which means that data collection and policy updates need to happen in sequence.

Our main contributions are summarized as follows:
\begin{itemize}
    \item We present a scheme for time-efficient reinforcement learning, \method, that can efficiently leverage tens of thousands of parallel environments on a workstation.
    \item We thoroughly investigate the effect of important hyperparameters such as the speed ratio on different processes that control the resource allocation and provide empirical guidelines for tuning these values to scale up $Q$-learning.
    \item We deploy different exploration strategies in parallel environments, which leads to robust exploration and mitigates the hassle of tuning the exploration noise.
    \item We demonstrate the effectiveness of our method on six Isaac Gym benchmark tasks~\citep{makoviychuk2021isaac} and show its superiority over state-of-the-art (SOTA) on-/off-policy algorithms. Our method \method achieves both faster learning in wall-clock time and better sample efficiency. Empirically, we also found that DDPG performs better than SAC in a massively parallel environment setting.
\end{itemize}

\section{Related Work}

\paragraph{Massively Parallel Simulation}
Simulation has been an important tool in various research fields, such as robotics, drug discovery, and physics. In the past, researchers have used simulators like MuJoCo~\citep{todorov2012mujoco} and PyBullet~\citep{coumans2016pybullet} for rigid body simulation. Recently, there has been a new wave of development in GPU-based simulation, e.g., Isaac Gym~\citep{makoviychuk2021isaac}. GPU-based simulation has substantially improved simulation speed by allowing massive amounts of parallel simulation on a single commodity GPU. It has been used in various challenging robotics control problems, including quadruped locomotion~\citep{rudin2022learning, margolis2022rapid} and dexterous manipulation~\citep{allshire2021transferring, chen2022system, chen2022visual}. With fast simulation, one can obtain much more environment interaction data in the same training time as before. This poses a challenge to RL algorithms in making the best use of the massive amount of data. A straightforward way is to use on-policy algorithms such as PPO, which can be easily scaled up and is also the default algorithm used by researchers in Isaac Gym. However, on-policy algorithms are less data-efficient. In our work, we investigate how to scale up off-policy algorithms to achieve higher sample efficiency and shorter wall-clock training time under massively parallel simulation. 

\paragraph{Distributed Reinforcement Learning} There have been numerous distributed reinforcement learning frameworks to speed up learning. One line of work focuses on $Q$-learning methods. Gorila~\citep{nair2015massively} distributes DQN agents to many machines where each machine has its local environment, replay buffer, and value learning, and uses asynchronous SGD for a centralized $Q$ function learning. Similarly, \citet{popov2017data} apply asynchronous SGD to the DDPG algorithm~\citep{lillicrap2015continuous}. Combining with prioritized replay~\citep{schaul2015prioritized}, $n$-step returns~\citep{sutton1988learning}, and double-Q learning~\citep{hasselt2010double}, \citet{horgan2018distributed} (Ape-X) parallelize the actor thread (environment interactions) for data collection and use a centralized learner thread for policy and value function learning. Built upon Ape-X, \citet{kapturowski2018recurrent} adapt the distributed prioritized experience replay for RNN-based DQN agents.

Another line of work improves the training speed on policy gradient methods. A3C~\citep{mnih2016asynchronous} uses asynchronous SGD across many CPU cores, with each running an actor learner on a single machine. \citet{babaeizadeh2016reinforcement} develop a hybrid CPU/GPU implementation of A3C, but it can have poor convergence due to the stale off-policy data being used for the on-policy update. \citet{espeholt2018impala} (IMPALA) introduce an off-policy correction scheme (V-trace) to mitigate the lagging issue between the actors and learners in distributed on-policy settings. \citet{espeholt2019seed} further improve the IMPALA training speed by moving the policy inference from the actor to the learner. \citet{clemente2017efficient} parallelize the environments for synchronous advantage actor-critic. \citet{heess2017emergence} propose a distributed version of PPO~\citep{schulman2017proximal} for training various locomotion skills in a diverse set of environments. \citet{wijmans2019dd} develop a decentralized version of distributed PPO to mitigate the synchronization overhead between different actor processes and applies it to a point-goal navigation task.

\begin{figure}[!tb]
    \centering
    \includegraphics[width=\linewidth]{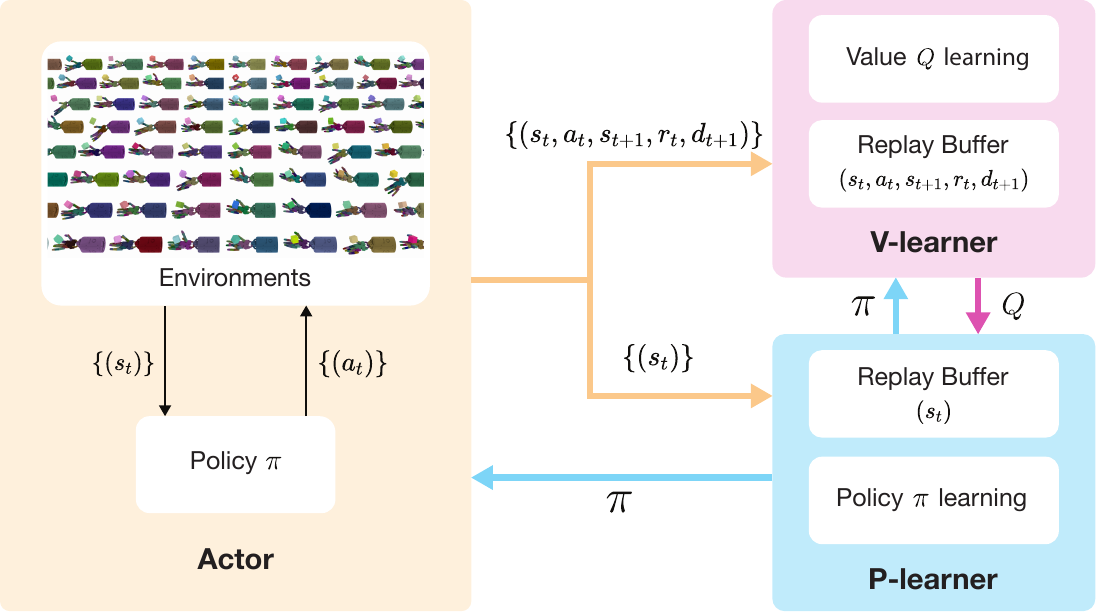}
    \caption{Overview of Parallel $Q$ Learning (PQL). We have three concurrent processes running: \actor, \plearner, \vlearner. \actor collects interaction data. \plearner updates the policy network. \vlearner updates the $Q$ functions.}
    \label{fig:arch}
\end{figure}

Our scheme is most closely related to Ape-X~\citep{horgan2018distributed} but has a number of key differences. \textbf{First}, our scheme is specifically designed for massively ($>>1000$) parallel GPU-based simulation. Our scheme is optimized for a single-machine setup, which can help democratize large-scale RL research. \textbf{Second}, we further decouple and parallelize the learning with two separate learners for policy function learning and $Q$-function learning, respectively. \textbf{Third}, we allocate a local replay buffer for each learner. This can reduce the communication cost between the replay buffer and the learners. \textbf{Fourth}, working with a single machine presents new challenges in balancing the computing resource between different parallel processes. Our scheme offers a mechanism to balance the computing resource among different processes.

\section{Method}

We developed a parallel off-policy training scheme, \textbf{P}arallel $\mathbf{Q}$-\textbf{L}earning (\method), for massively parallel GPU-based simulation, where thousands of environments can be simulated simultaneously on a single GPU. In a typical actor-critic $Q$-learning method, three components run sequentially: a policy function, a $Q$-value function, and an environment. Agents roll out the policy in the environments and collect interaction data, which is added to a replay buffer; then, the value function is updated to minimize the Bellman error, after which the policy function is updated to maximize the $Q$ values. This sequential scheme slows down the training, as each component needs to wait for the other two to finish before proceeding. To maximize the learning speed and reduce the waiting time, we parallelize the computation of all three components. This allows for more network updates per data point, which can improve the utilization of the massive amount of data and lead to better training speed, as demonstrated in the experiments. Off-policy RL methods are well-suited for parallelization as the interaction data in a replay buffer does not need to come from the latest policy. In contrast, on-policy methods such as PPO require using the rollout data from the latest policy (on-policy data) to update the policy, thus making it non-trivial to parallelize the data collection and policy/value function update.

Our scheme is optimized for training speed in terms of wall-clock time and can be readily applied on a workstation. It is built upon DDPG~\citep{lillicrap2015continuous}, but can be easily extended to other off-policy algorithms such as SAC~\citep{haarnoja2018soft} (see Appendix \ref{appsec:extra_exps}). Our scheme also incorporates common techniques used to improve Q learning performance, such as double $Q$ learning~\citep{hasselt2010double} and $n$-step returns~\citep{sutton1988learning}. Furthermore, we experimented with adding distributional RL~\cite{bellemare2017distributional} to PQL, which we refer to as \textbf{PQL-D}. While it improves performance on challenging manipulation tasks, it leads to a slight decrease in the convergence speed of the RL agent. In this paper, we use the following notation: at time step $t$, $s_t$ represents observation data, $a_t$ represents action command, $r_t$ represents the reward, $d_t$ represents whether the environment terminates, $\pi(s_t)$ represents the policy network, $Q(s_t,a_t)$ represents the $Q$ network, $Q^\prime(s_t, a_t)$ represents the target $Q$ network, and $N$ represents the number of parallel environments.

\subsection{Scheme Overview}
\label{subsec:framework}
PQL parallelizes data collection, policy learning, and value learning into three processes, as shown in \figref{fig:arch}. We refer to them as \actor, \plearner, and \vlearner, respectively. 

\begin{itemize}[wide, labelindent=0pt, labelwidth=0pt]
    \item \actor: We collect a batch of interaction data using parallel environments. We use Isaac Gym~\citep{makoviychuk2021isaac} as our simulation engine, which supports massively parallel simulation. Note that we do not make any Isaac-Gym-specific assumptions, and PQL is optimized for any GPU-based simulator that supports a large number of parallel environments. In the \actor process, the agent interacts with tens of thousands of environments according to an exploration policy. Therefore, we maintain a local policy network $\pi^a(s_t)$, which is periodically synchronized with the policy network $\pi^p(s_t)$ in \plearner (which we explain below).
    \item \vlearner: We create a dedicated process for training value functions, which allows for continuous updates without being interrupted by data collection or policy network updates. To compute the Bellman error, we need the policy network to estimate the optimal action and the replay buffer to sample a batch of I.I.D. training data. Since we use a dedicated process for updating value functions, \vlearner must frequently query the policy network and sample data from the replay buffer. To reduce the communication overhead of the policy and data across processes, we maintain a local version of the policy network $\pi^v(s_t)$ and the replay buffer in \vlearner. $\pi^v(s_t)$ is synced with $\pi^p(s_t)$ in \plearner periodically. When the GPU memory is sufficiently large to host the entire replay buffer, which is usually the case when observations are not images, we construct the replay buffer on the GPU to avoid the CPU-GPU data transfer bottleneck. 
    \item \plearner: We use another dedicated process for updating the policy network $\pi^p(s_t)$, which is optimized to maximize the $Q^p(s_t, \pi^p(s_t))$. We also maintain a local replay buffer of $\{(s_t)\}$ and a value function $Q^p(s_t, a_t)$ in \plearner to reduce communication overhead across processes. $Q^p(s_t, a_t)$ is periodically updated with $Q^v(s_t, a_t)$ in \vlearner. 
\end{itemize}
 We use Ray~\citep{moritz2017ray} for parallelization. The pseudo-code is in Algorithm \ref{alg:actor}, \ref{alg:plearner}, and \ref{alg:vlearner} in the Appendix \ref{appsec:code}.

\paragraph{Data Transfer}
Suppose there are $N$ parallel environments in the \actor process. At each step, the \actor rolls out the policy $\pi_a(s_t)$ and generates $N$ pairs of $(s_t, a_t, s_{t+1}, r_t, d_{t+1})$.  Then the \actor sends the entire batch of interaction data $\{(s_t, a_t, s_{t+1}, r_t, d_{t+1})\}$ to the \vlearner (see \figref{fig:arch}). Since policy update in \plearner only needs state information, \actor only sends $\{(s_t)\}$ to the \plearner.

\paragraph{Network Transfer}
The \vlearner periodically sends the parameters of the $Q^v(s_t, a_t)$
to \plearner, which updates the local $Q^p(s_t, a_t)$ in \plearner. The \plearner sends the policy network $\pi^{p}(s_t)$ to both the \actor and \vlearner.

Both the data and policy network transfer happen concurrently.

\subsection{Balance between \actor, \plearner, and \vlearner}
\label{subsec:ratio_balance}

Our scheme allows the \actor, \plearner, and \vlearner to run concurrently. However, if each process ran as fast as possible, independent of each other, it can make training unstable.
Thus, we explicitly control the update frequencies of the three processes using the following two ratios:
\begin{align*}
    \beta_{a:v}:=\frac{f_a}{f_v}\qquad \text{and} \qquad \beta_{p:v}:=\frac{f_p}{f_v},
\end{align*}
where $f_a$ is the number of rollout steps per environment in \actor per unit time, $f_v$ is the number of $Q$ function updates in \vlearner per unit time, $f_p$ is the number of policy updates in \plearner per unit time. $\beta_{a:v}$ determines how many steps \actor rolls out the policy with $N$ environments when one $Q$ function update is performed in the \vlearner. $\beta_{p:v}$ decides how many $Q$ function updates are performed in \vlearner when \plearner updates the policy once. Once the ratios are set, we monitor the progress of each process and dynamically adjust the speed of \actor and \plearner by letting the process wait if necessary.

Controlling the three processes via $\beta_{a:v}, \beta_{p:v}$ provides three major advantages. \textbf{First}, it allows us to balance the resource allocation of each process and reduce the variance of PQL's performance. Given a fixed amount of computing resources, the ability to let some of the processes wait enables other processes to use the GPU resource more. This is particularly important when working with limited resources. If there is only one GPU, and all three processes run freely on it, simulation with a large number of environments can cause very high GPU utilization, which slows down the \plearner and \vlearner and leads to worse performance. Note that such control was not examined in prior studies, such as Ape-X~\cite{horgan2018distributed}, where a computer cluster was used for both the simulation and network training ---  the phenomenon of competing for limited computing resources (all three processes on one GPU) did not occur. On the other hand, leaving each process running freely creates more variance in the training speed and learning performance as the simulation speed and network training speed are heavily dependent on the task complexity, network size, computer hardware, etc. For example, simulation for contact-rich tasks can be slower than others; some tasks might require a deeper policy network or $Q$ networks; even the GPUs on a machine might have different running conditions at different times, leading to different speeds across processes and further leading to different learning performances. 

\textbf{Second}, ratio control can improve convergence speed. For example, prior works~\citep{fujimoto2018addressing} show that updating the policy network less frequently than the $Q$ functions leads to better learning. 

\textbf{Third}, the ratio $\beta_{p:v}$ can be interpreted as the frequency of the target policy network update. One may notice that we use a lagged policy to update the $Q$ function and synchronize it according to the above ratio. Therefore, we do not create a target policy network explicitly, but every synchronization can be considered as a hard update of the policy network.

\begin{figure}[!tb]
    \centering
    \resizebox{\linewidth}{!}{%
    \includegraphics[height=0.33\linewidth]{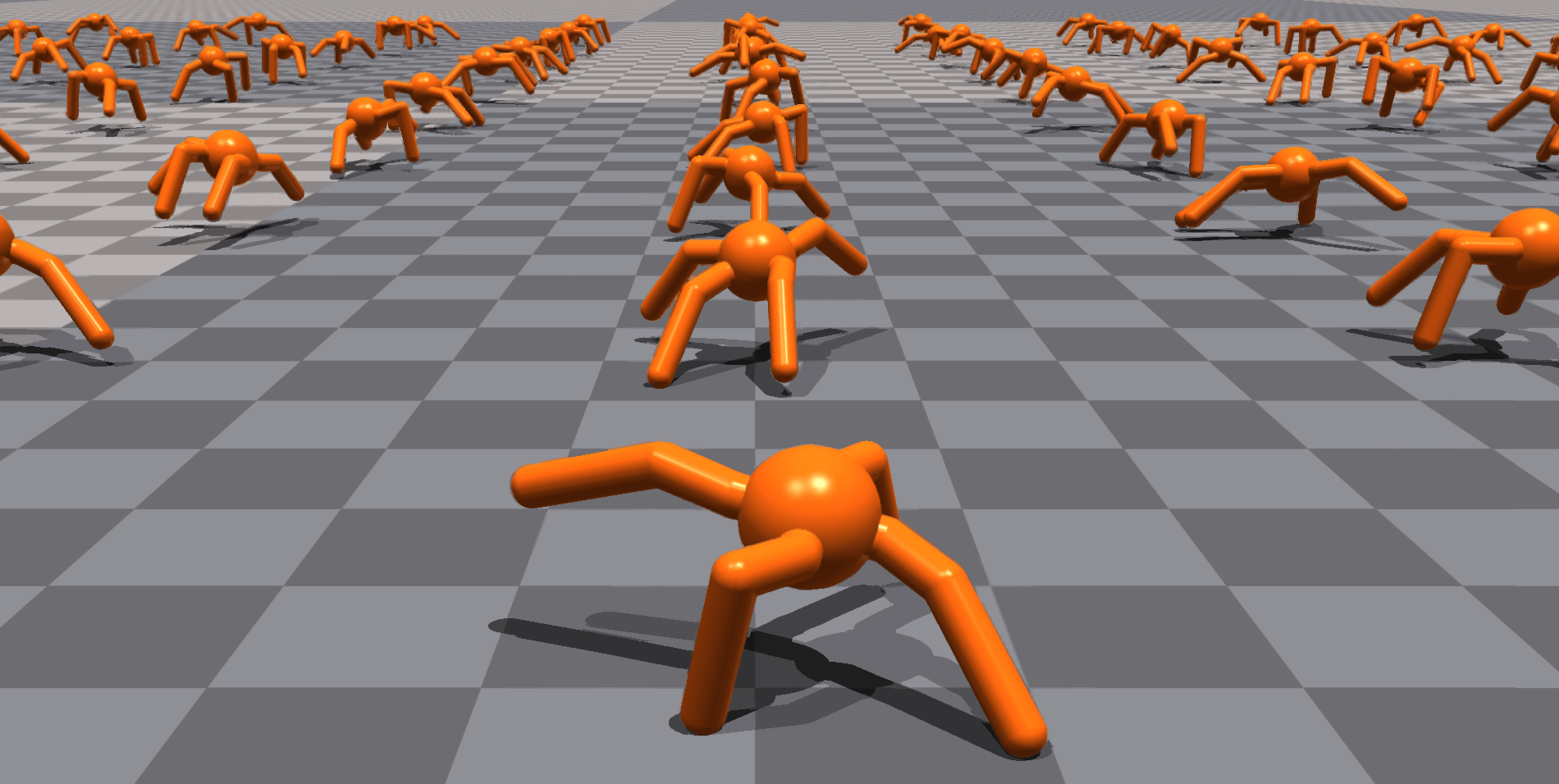}
    \includegraphics[height=0.33\linewidth]{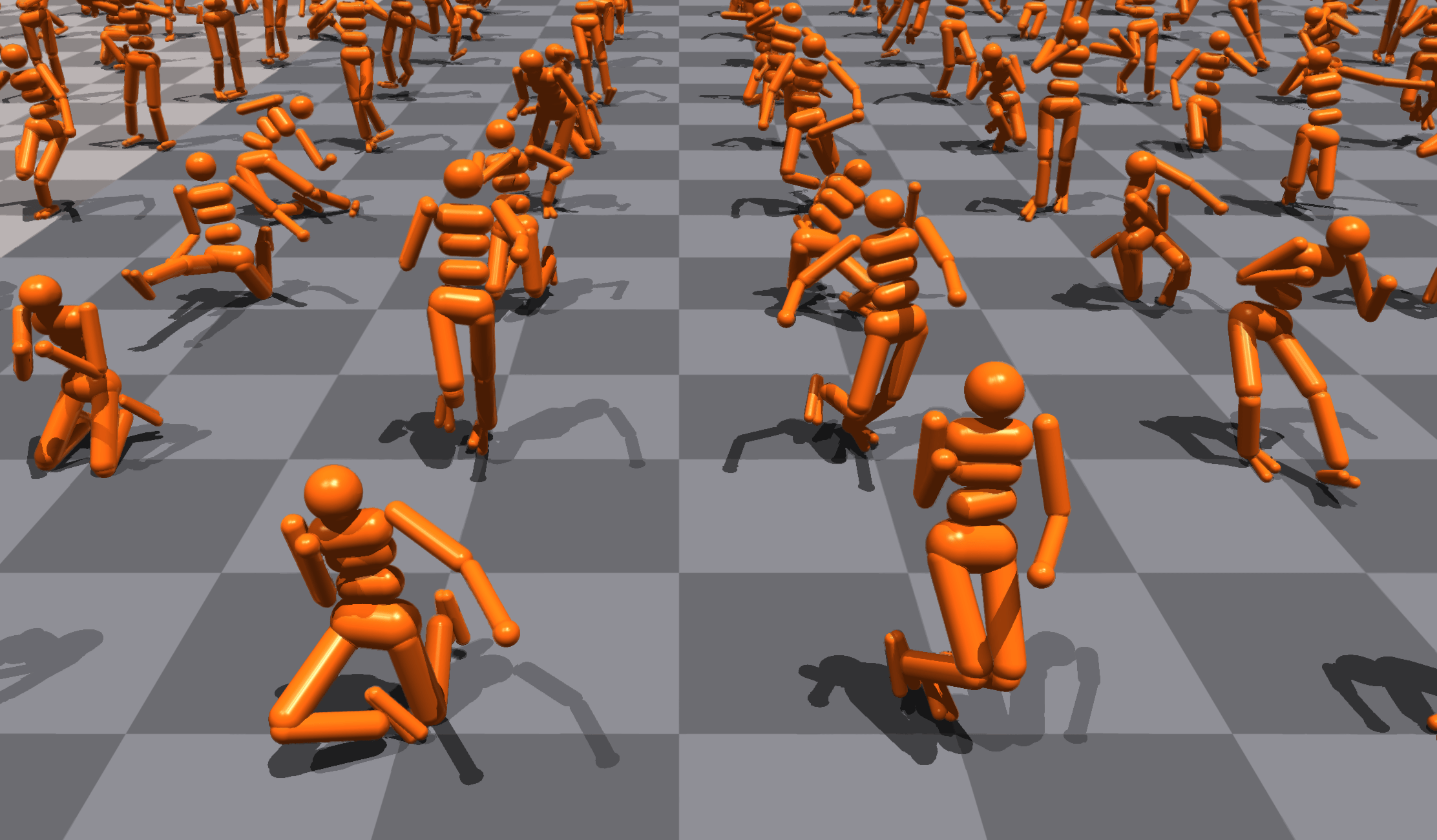}
    \includegraphics[height=0.33\linewidth]{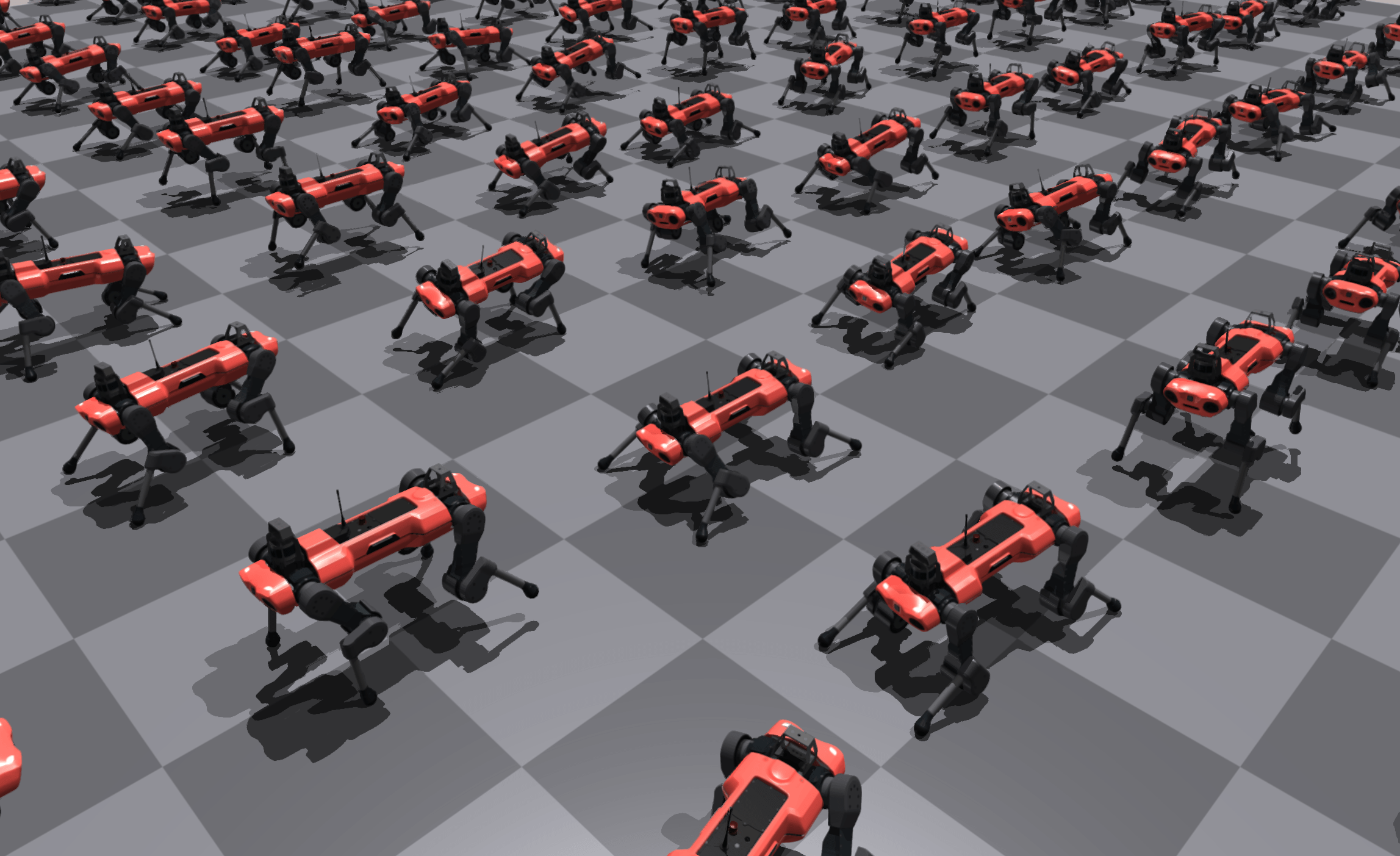}}\\
    \resizebox{\linewidth}{!}{%
    \includegraphics[height=0.33\linewidth]{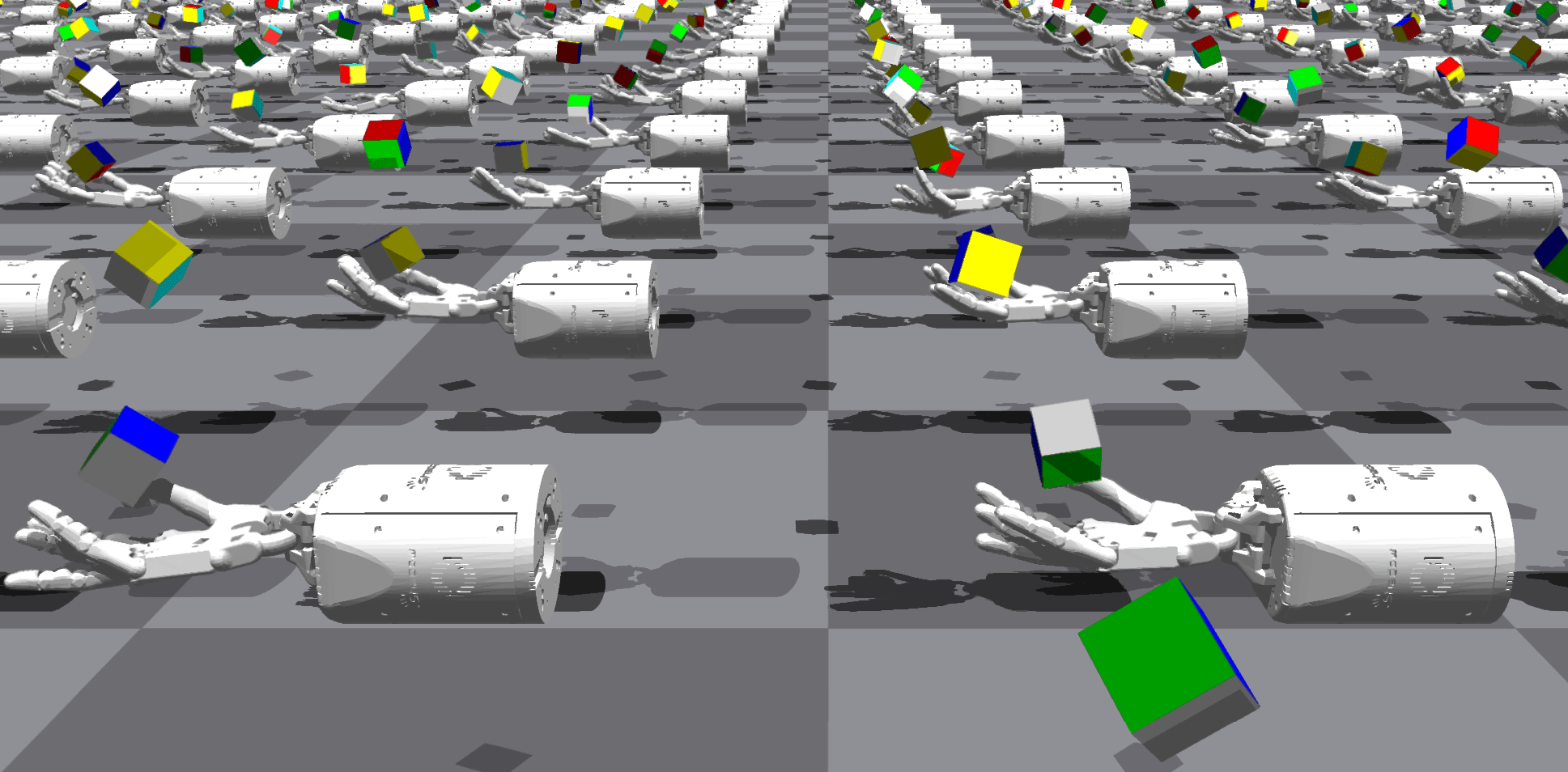}
    \includegraphics[height=0.33\linewidth]{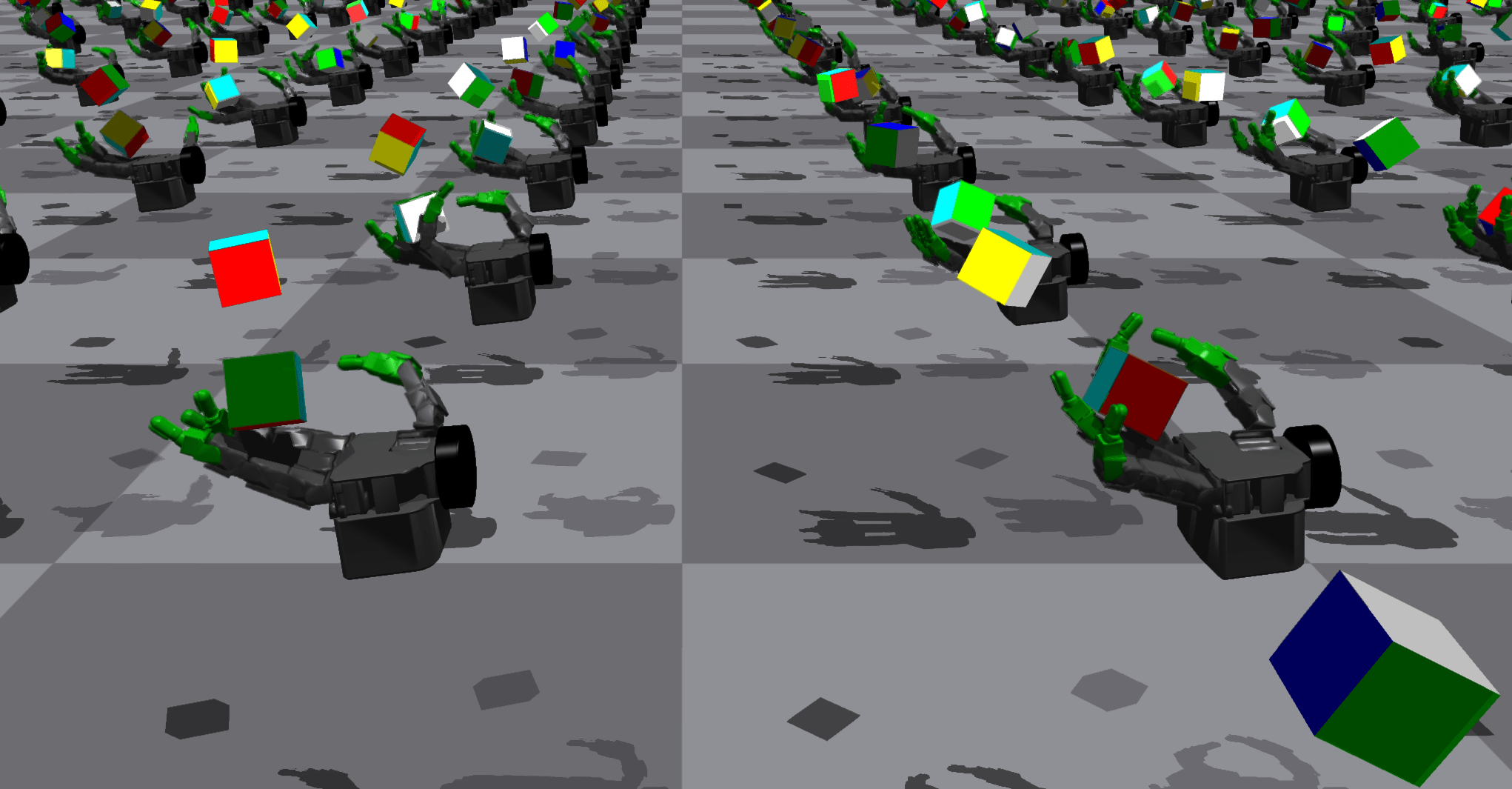}
    \includegraphics[height=0.33\linewidth]{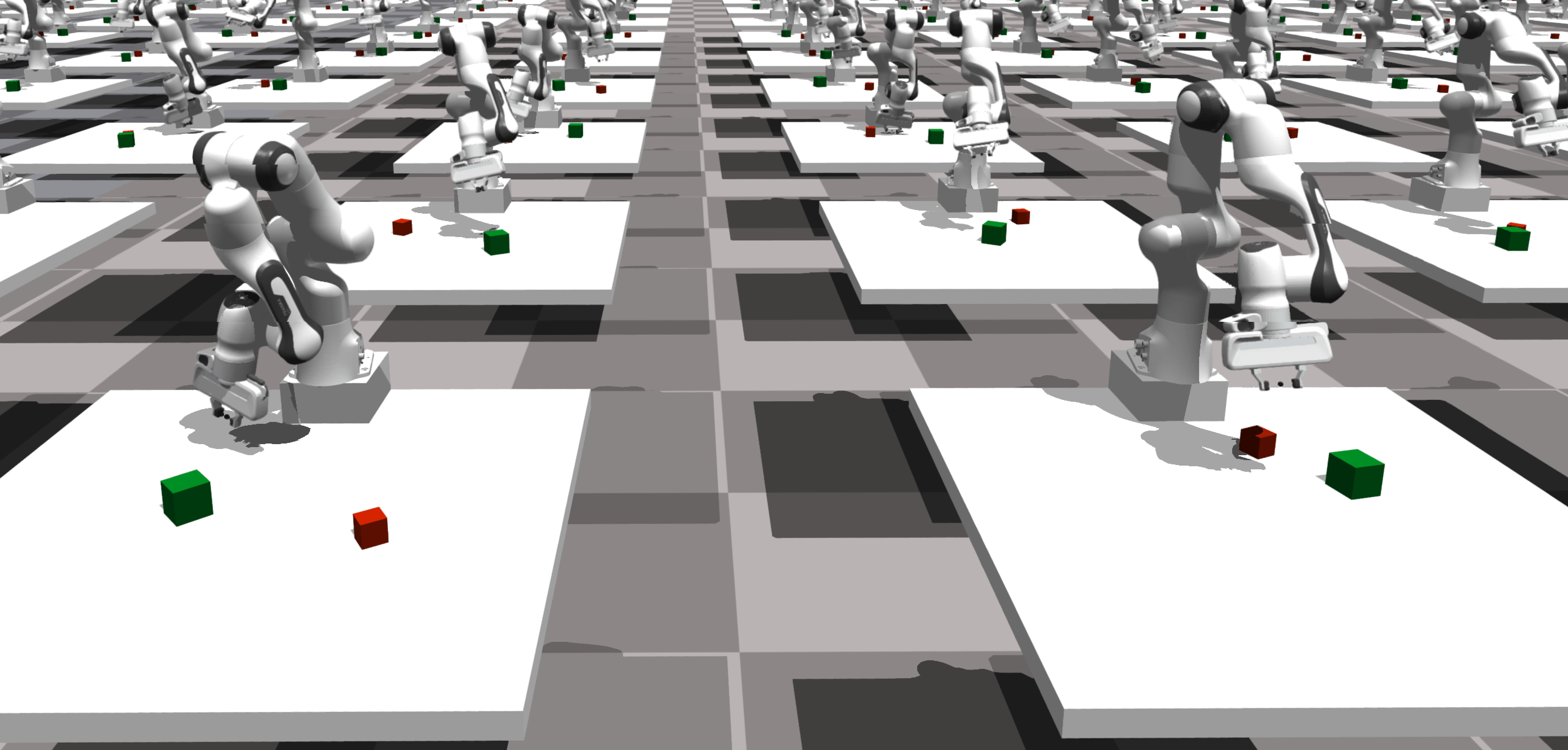}
     }
    \caption{We experiment on six Isaac Gym tasks: \ant, \humanoid, \anymal, \shadow, \allegro, \franka.}
    \label{fig:tasks}
    \vspace{-1cm}
\end{figure}

\subsection{Mixed Exploration}
\label{subsec:mix_exploration}
We can achieve improvement in convergence by having a good exploration strategy. Too much exploration can make agents fail to latch onto useful experience and learn a good policy quickly, while too little exploration does not give the agent enough good interaction data to improve the policy. Balancing the exploration and exploitation often requires extensive hyper-parameter tuning or complex scheduling mechanisms. In DDPG, one common practice to control exploration is to set the standard deviation $\sigma$ of the uncorrelated and zero-mean Gaussian noise that is being added to the deterministic policy output ($a_t =\max(\min(\pi(s_t)+\mathcal{N}(0, \sigma), a_u), a_l)$~\citep{SpinningUp2018, JMLR:v22:20-376, yang2022overcoming} where $a_t\in [a_l, a_u]$). Since it is difficult to predict how much exploration noise is appropriate, one typically needs to tune $\sigma$ for each task. Can we mitigate the hassle of tuning $\sigma$? Our idea is that instead of finding the best $\sigma$ value, we can try out different $\sigma$ values altogether, which we call \textbf{mixed exploration}. Even if some $\sigma$ values lead to bad exploration at a certain training stage, others can still generate good exploration data. This strategy is easily implemented thanks to the massively parallel simulation, as we can use different $\sigma$ values in different parallel environments. Similar ideas have been used in prior works~\citep{horgan2018distributed, mnih2016asynchronous}. In our work, we uniformly generate the noise levels in the range of $[\sigma_{\min}, \sigma_{\max}]$. For the $i^{th}$ environment out of $N$ environments, $\sigma_i=\sigma_{\min}+\frac{i-1}{N-1}(\sigma_{\max}-\sigma_{\min})$ where $i\in\{1,2,...,N\}$. We use $\sigma_{\min}=0.05, \sigma_{\max}=0.8$ for all the tasks in our experiments.

\section{Experiments}

In this section, we demonstrate the effectiveness of our method compared to SOTA baselines, analyze the effects of key hyper-parameters, and provide empirical guidelines for setting their values. All experiments are carried out on a single workstation with a few GPUs. We run each experiment with five random seeds and plot their mean and standard error.

\subsection{Setup} 
\paragraph{Tasks} We evaluate our method on six Isaac Gym benchmark tasks \citep{makoviychuk2021isaac}: \ant, \humanoid, \anymal, \shadow, \allegro, and \franka (see \figref{fig:tasks}). For more details about these tasks, please refer to \citep{makoviychuk2021isaac}. Additionally, we provide two more tasks in \secref{sec:vision-based}: (1) a vision-based \ball task and (2) a contact-rich dexterous manipulation task that requires learning to reorient hundreds of different objects using a \dcalw with a single policy~\citep{chen2022visual}. Note that we use the four-finger hand version and do not include any domain randomization.

\paragraph{Baselines} We consider the following baselines: (1) \textbf{PPO}~\citep{schulman2017proximal}, which is the default algorithm used by many prior works~\citep{makoviychuk2021isaac, chen2022system, allshire2021transferring} that use Isaac Gym for simulation, (2) \textbf{DDPG(n)}: DDPG~\citep{lillicrap2015continuous} implementation with double Q learning and $n$-step returns, (3) \textbf{SAC(n)}: SAC~\citep{haarnoja2018soft} implementation with $n$-step returns. Hyper-parameters are available in Appendix \ref{appsubsec:hyper}.

\paragraph{Hardware} We use NVIDIA GeForce RTX $3090$ GPUs as our default GPUs for the experiments unless otherwise specified. More details are shown in \tblref{tbl:hardware} in the appendix.

\subsection{PQL learns faster than baselines}
\begin{figure}[!tb]
    \centering
    \subfigure[]{\label{fig:ant_baseline}\includegraphics[width=0.45\linewidth]{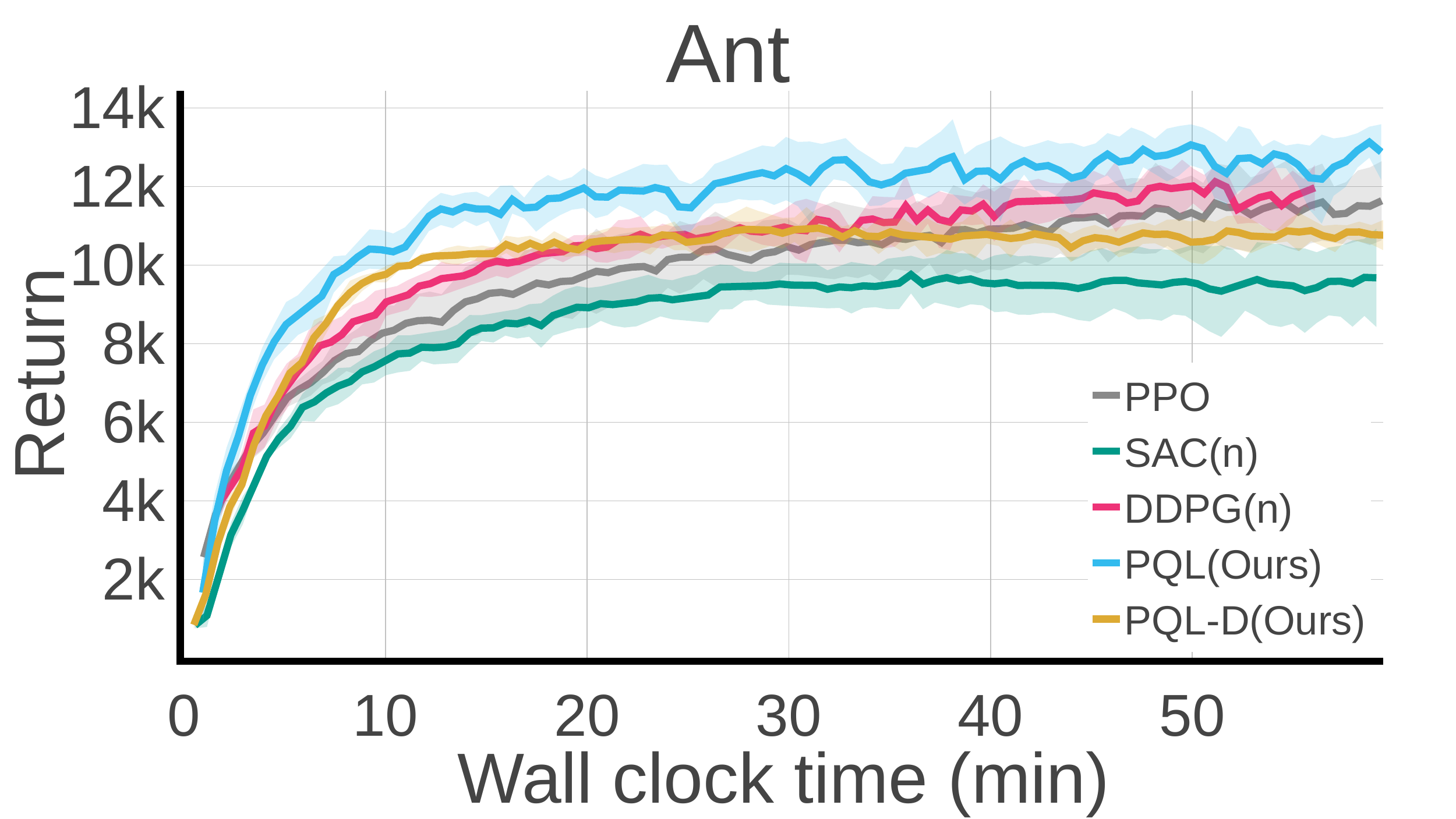}}
    \subfigure[]{\label{fig:humanoid_baseline}\includegraphics[width=0.45\linewidth]{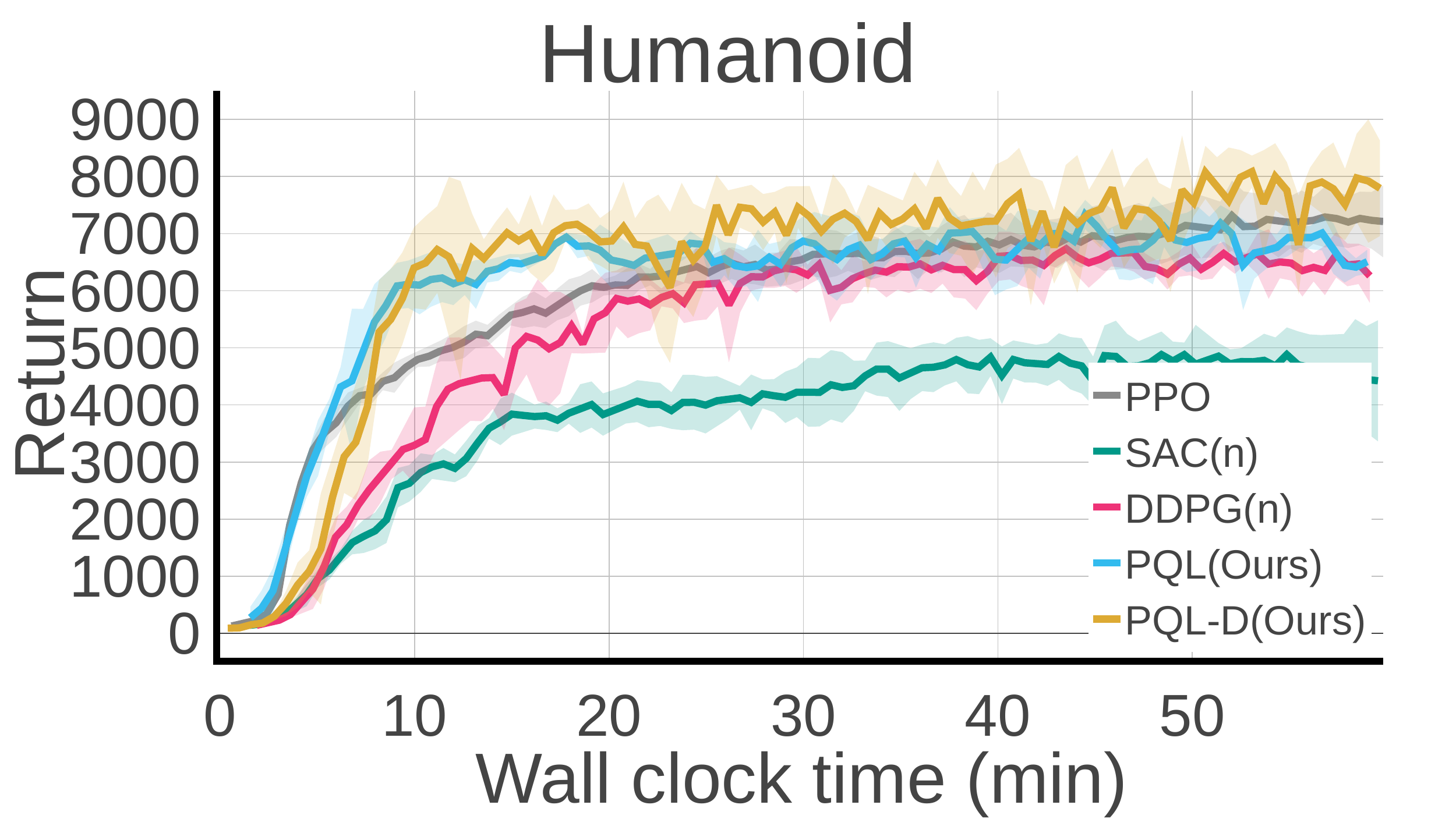}}\\
    \subfigure[]{\label{fig:anymal_baseline}\includegraphics[width=0.45\linewidth]{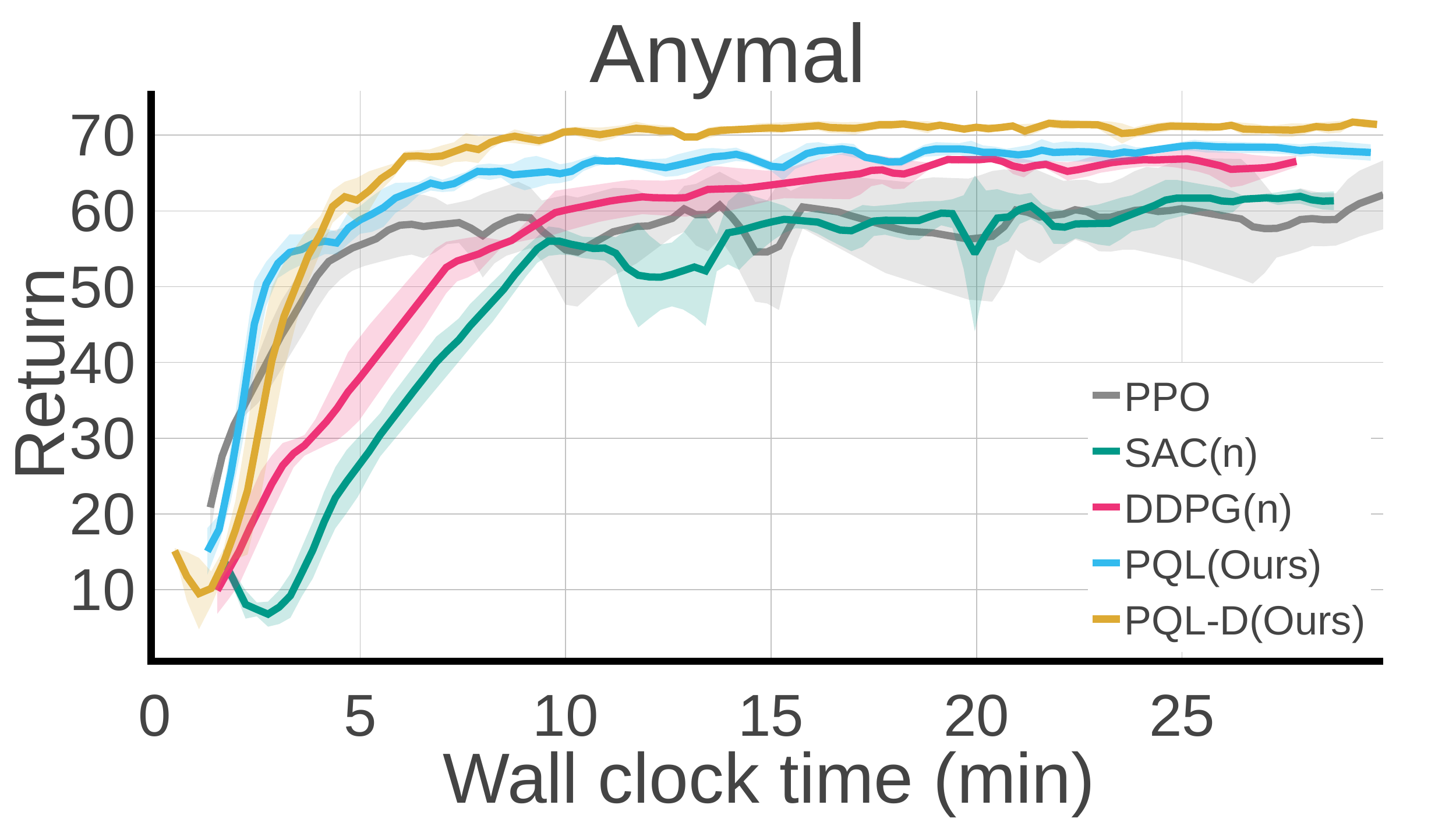}}
    \subfigure[]{\label{fig:franka_baseline}\includegraphics[width=0.45\linewidth]{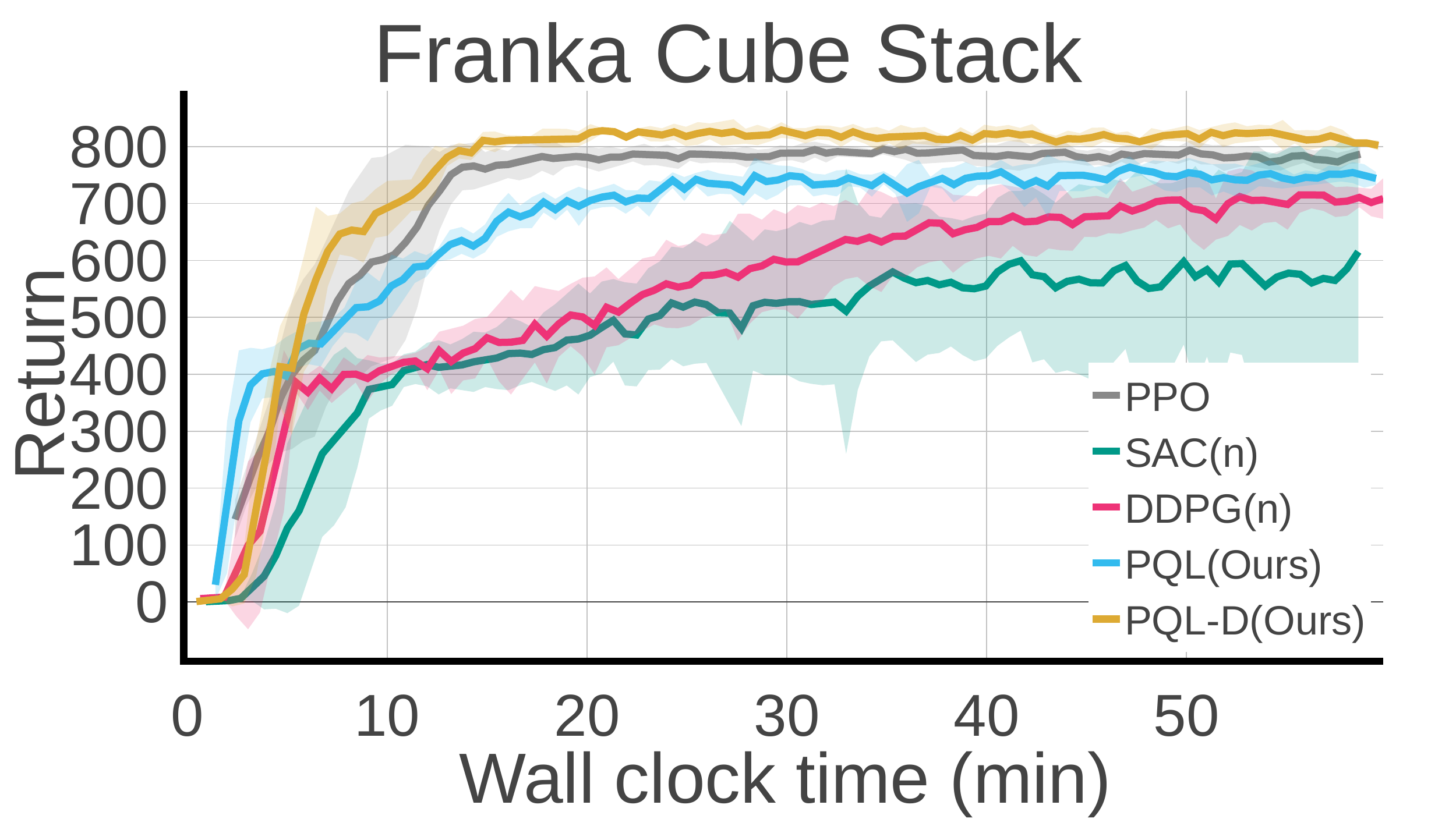}}\\
    \subfigure[]{\label{fig:allegro_baseline}\includegraphics[width=0.45\linewidth]{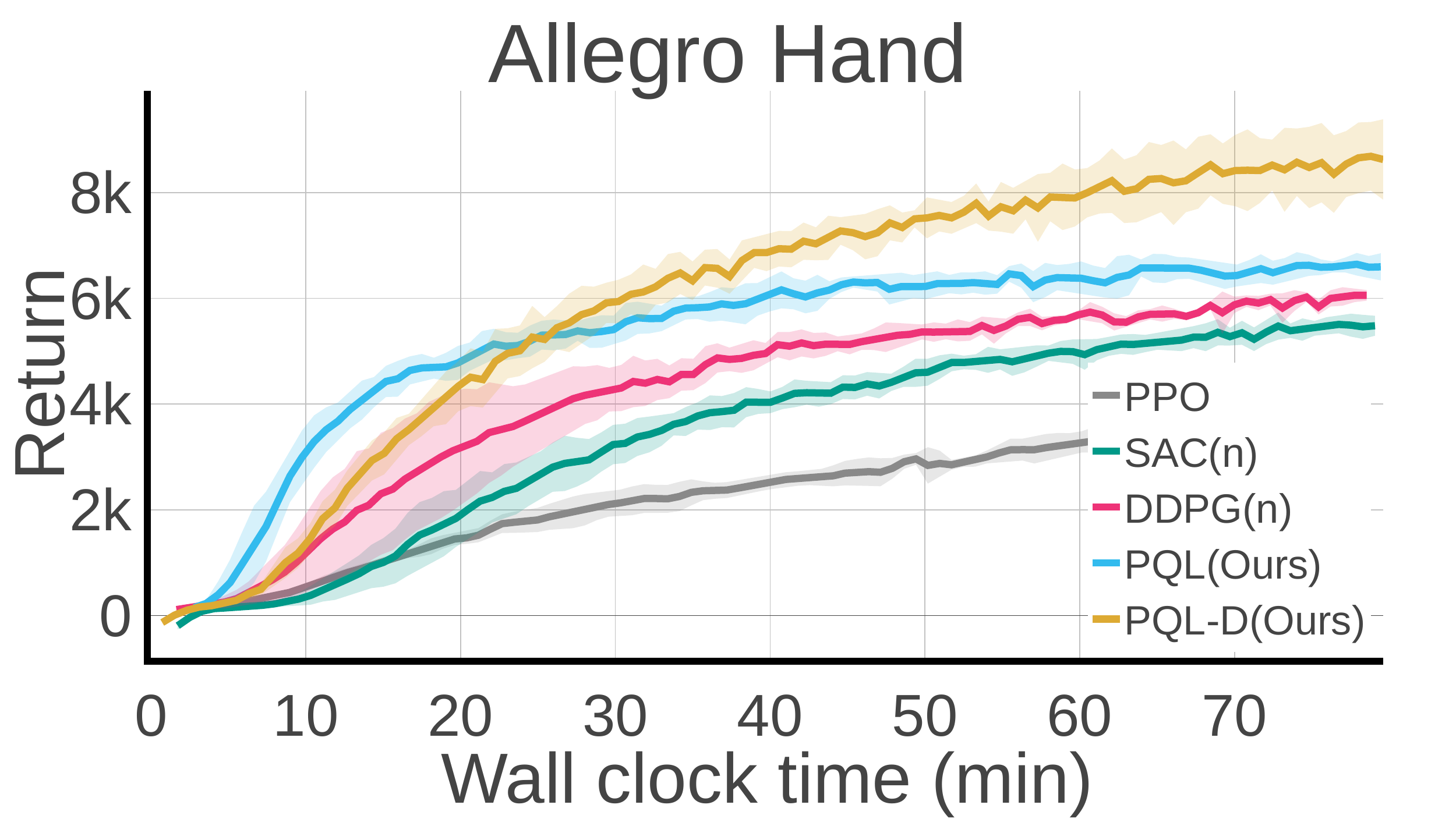}}
    \subfigure[]{\label{fig:shadow_baseline}\includegraphics[width=0.45\linewidth]{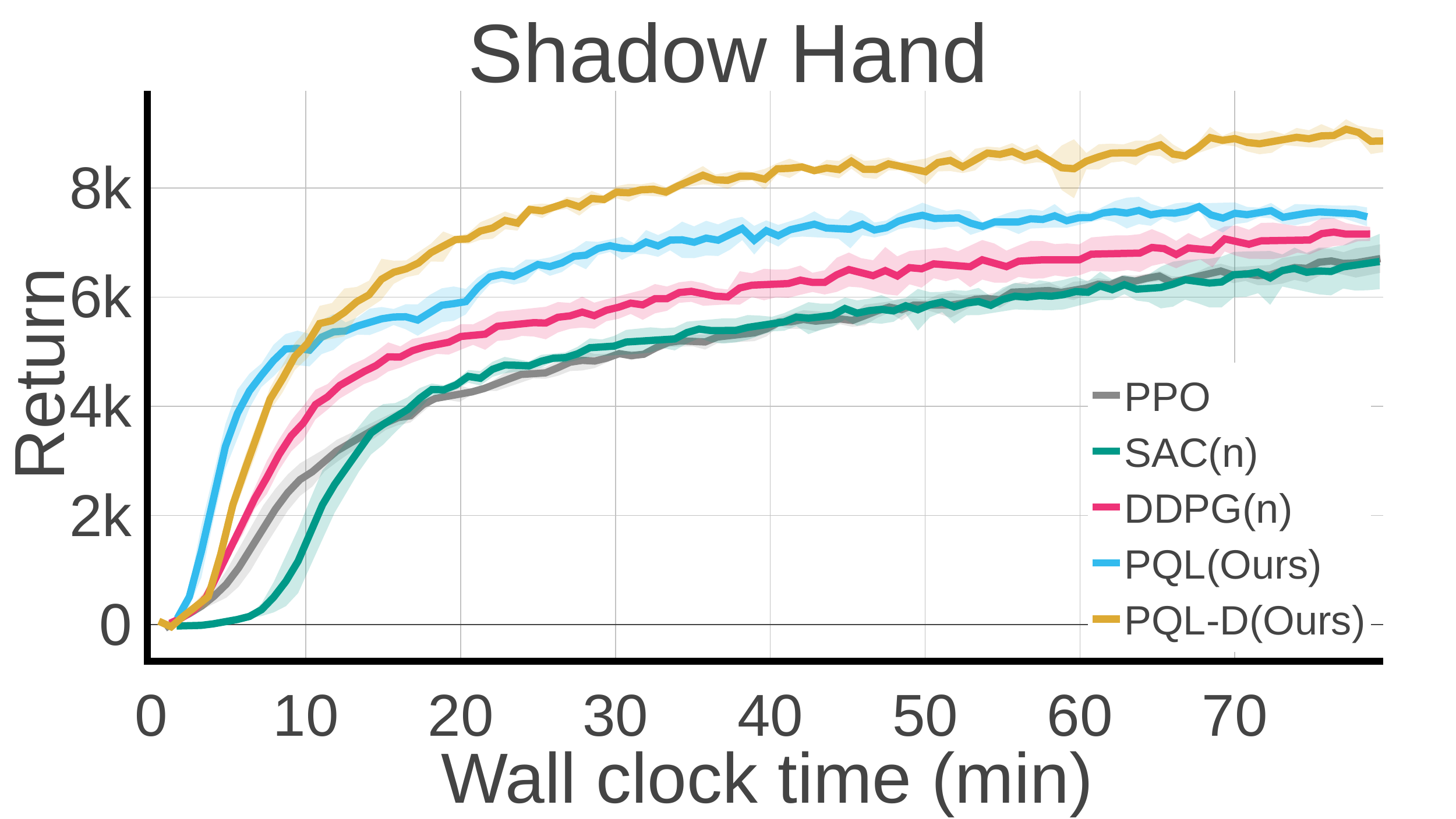}}
    \caption{We compare our methods to the SOTA RL algorithms (PPO, SAC with $n$-step returns, DDPG with $n$-step returns). We use $4096$ environments for training in all tasks except the PPO baseline on \shadow and \allegro tasks, where we use $16384$ as it gives the best performance for PPO on these two tasks as shown in \figref{fig:shadow_ppo_num_envs}. Our methods achieve the fastest learning speed in almost all tasks. }
    \label{fig:baselines}
\end{figure}

The first and most important question to answer is whether PQL leads to faster learning than SOTA baselines. To answer this, we compared the learning curves of PQL and PQL-D (PQL with distributional RL) with baselines on six benchmark tasks. As shown in \figref{fig:baselines}, our method (PQL, PQL-D) achieves the fastest policy learning in five out of six tasks compared to all baselines. Moreover, we observed that adding distributional RL to PQL can further boost learning speed.  \figref{fig:baselines} shows that in five out of six tasks, PQL-D achieves wall-clock time faster than, or at least on par with, PQL. The improvements are most salient on the two challenging contact-rich manipulation tasks (\shadow and \allegro). Additionally, the faster learning of PQL than DDPG(n) demonstrates the advantage of using a parallel scheme for data collection and network updates. We also found that DDPG(n) outperforms SAC(n) in all tasks. This could be due to the fact that the exploration scheme in DDPG can scale up better than the one in SAC. In DDPG, we apply the same mixed exploration as in PQL, while the exploration of SAC solely comes from sampling in the stochastic policy distribution, which can be heavily affected by the quality of the policy distribution.

\subsection{How well does mixed exploration perform?}
\label{subsec:mixed_exploration_results}

\begin{figure}[!tb]
    \centering
    \subfigure[]{\label{fig:ant_noise}\includegraphics[width=0.49\linewidth]{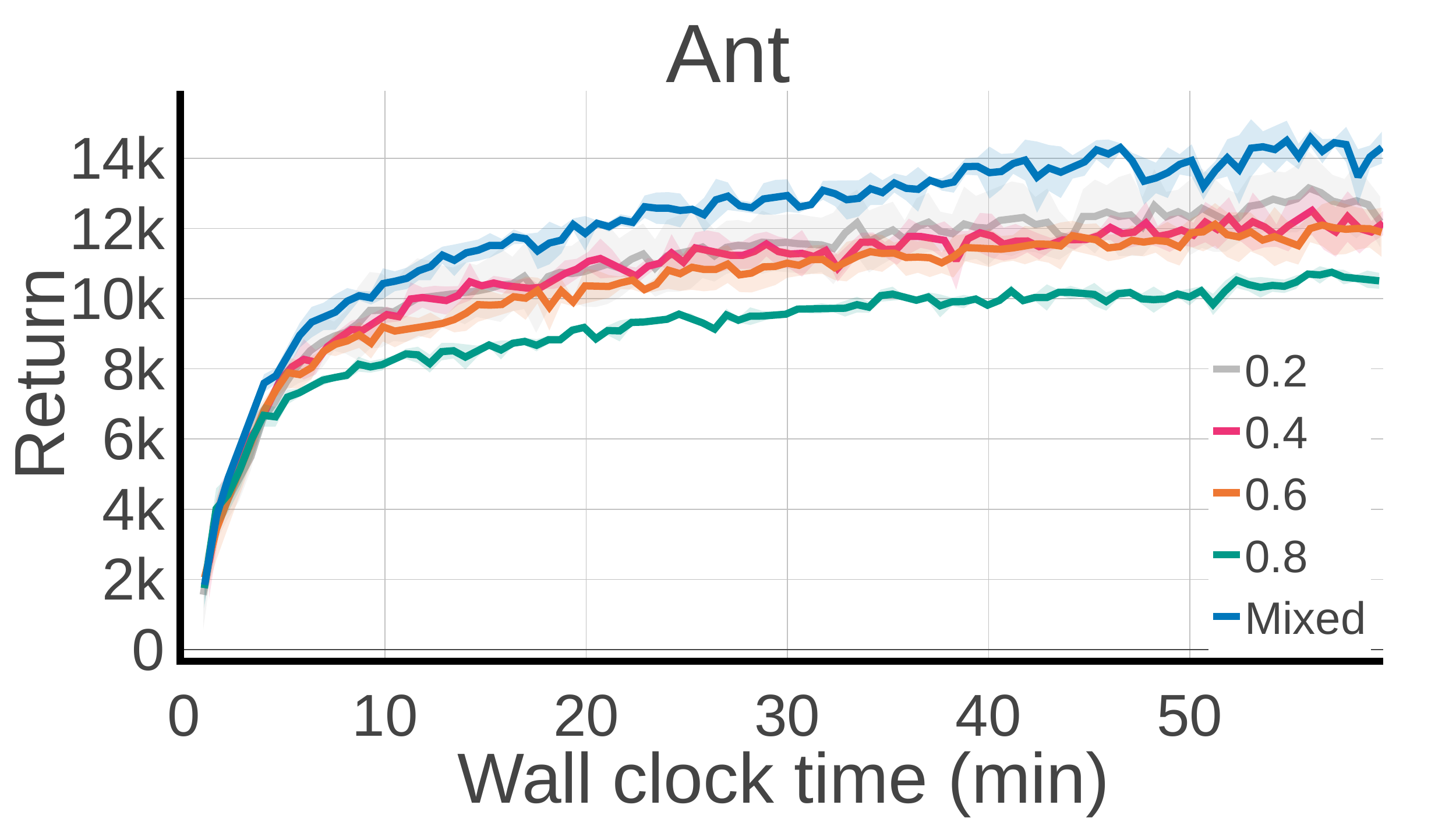}}
    \subfigure[]{\label{fig:humanoid_noise}\includegraphics[width=0.49\linewidth]{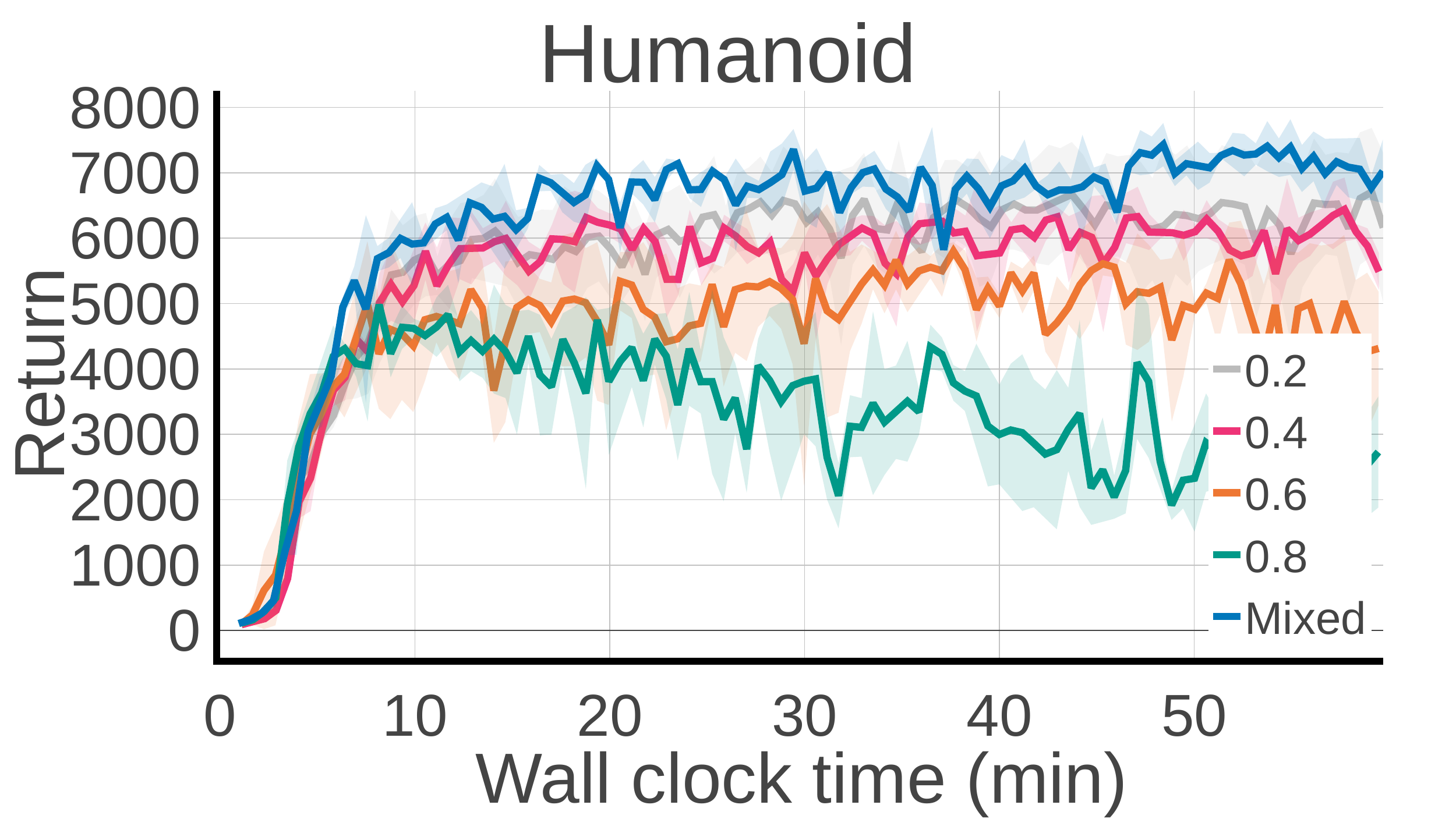}}\\
    \subfigure[]{\label{fig:franka_noise}\includegraphics[width=0.49\linewidth]{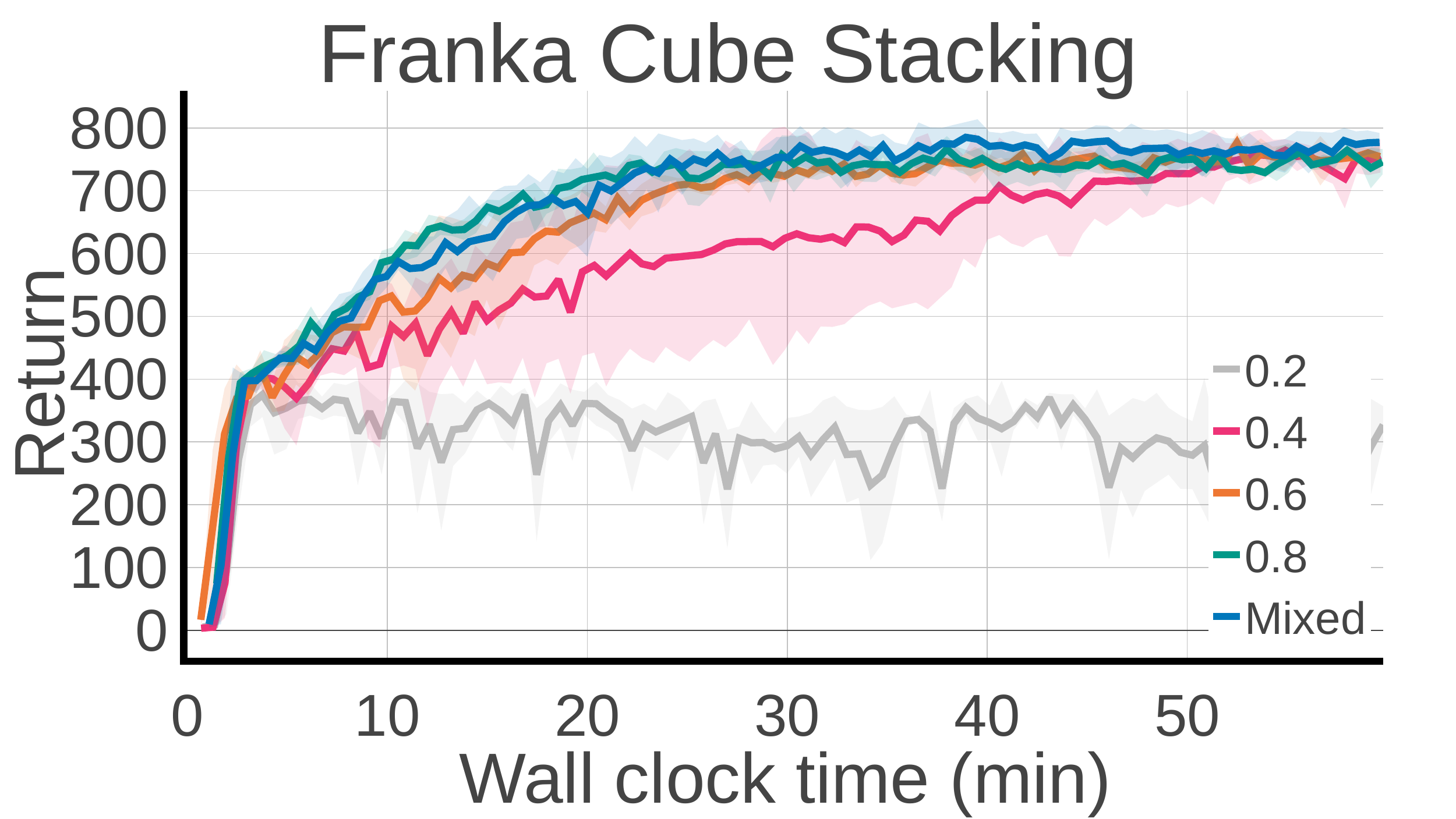}}
    \subfigure[]{\label{fig:shadow_noise}\includegraphics[width=0.49\linewidth]{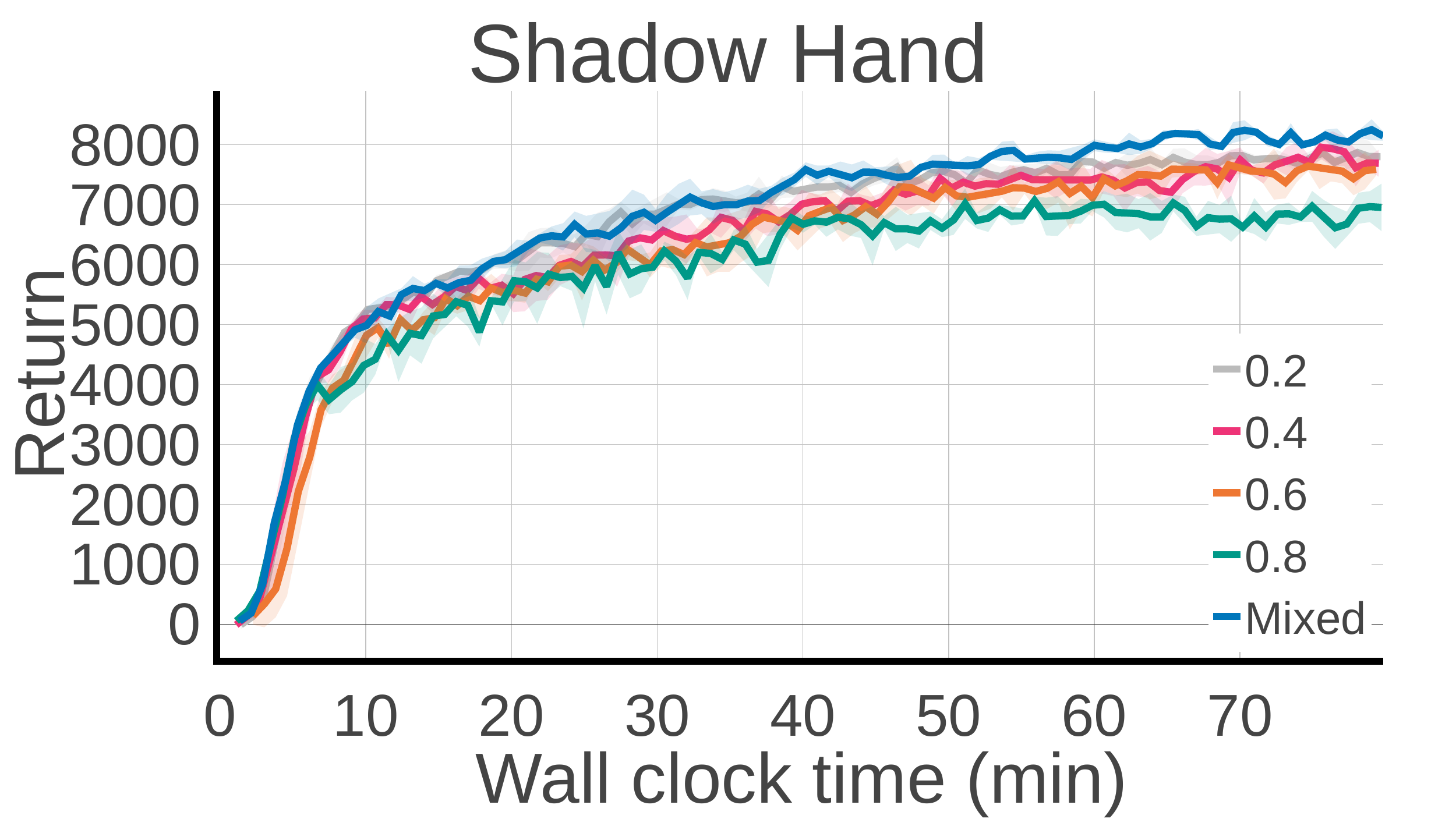}}
    \vspace{-2mm}
    \caption{We compared our proposed mixed exploration scheme by applying different constant maximum noise values. We can see that the mixed exploration scheme either outperforms or is on par with other schemes, which can save the tuning effort on the noise level.}
    \label{fig:noise}
\end{figure}

As discussed in \secref{subsec:mix_exploration}, massively parallel simulation enables us to deploy different exploration strategies in different environments to generate more diverse exploration trajectories. We use a simple mixed exploration strategy, as described in \secref{subsec:mix_exploration}, and compare its effectiveness to cases where all the environments use the same exploration capacity (the same $\sigma$ values). We experimented with $\sigma\in\{0.2, 0.4, 0.6, 0.8\}$. As shown in \figref{fig:noise}, the learning performance is significantly affected by the choice of $\sigma$ value. If we use the same $\sigma$ value for all parallel environments, then we need to tune $\sigma$ for each task. In contrast, the mixed exploration strategy, where each environment uses a different $\sigma$ value, outperforms (learns faster or at least as fast as) all other fixed $\sigma$ values. This implies that using the mixed exploration strategy can reduce the tuning effort needed for $\sigma$ values per task.

\subsection{Effects of different hyper-parameters}
In this section, we investigate the effects of the number of environments, $\beta_{p:v}$, $\beta_{a:v}$, batch size, replay buffer size, and the number of GPUs. These hyper-parameters are of particular interest given the massively parallel simulation ($N>>1000$) and our parallel scheme. Lastly, GPU hardware can also impact learning speed. To explore this, we conduct experiments using four different GPU models and analyze the effect of GPU hardware on performance in Appendix \ref{appsec:extra_exps}. Overall, PQL works robustly across different GPU models. 

\subsubsection{How does the number of environments $N$ affect policy learning?}
\begin{figure}[!tb]
    \centering
    \subfigure[]{\label{fig:ant_num_envs}\includegraphics[width=0.49\linewidth]{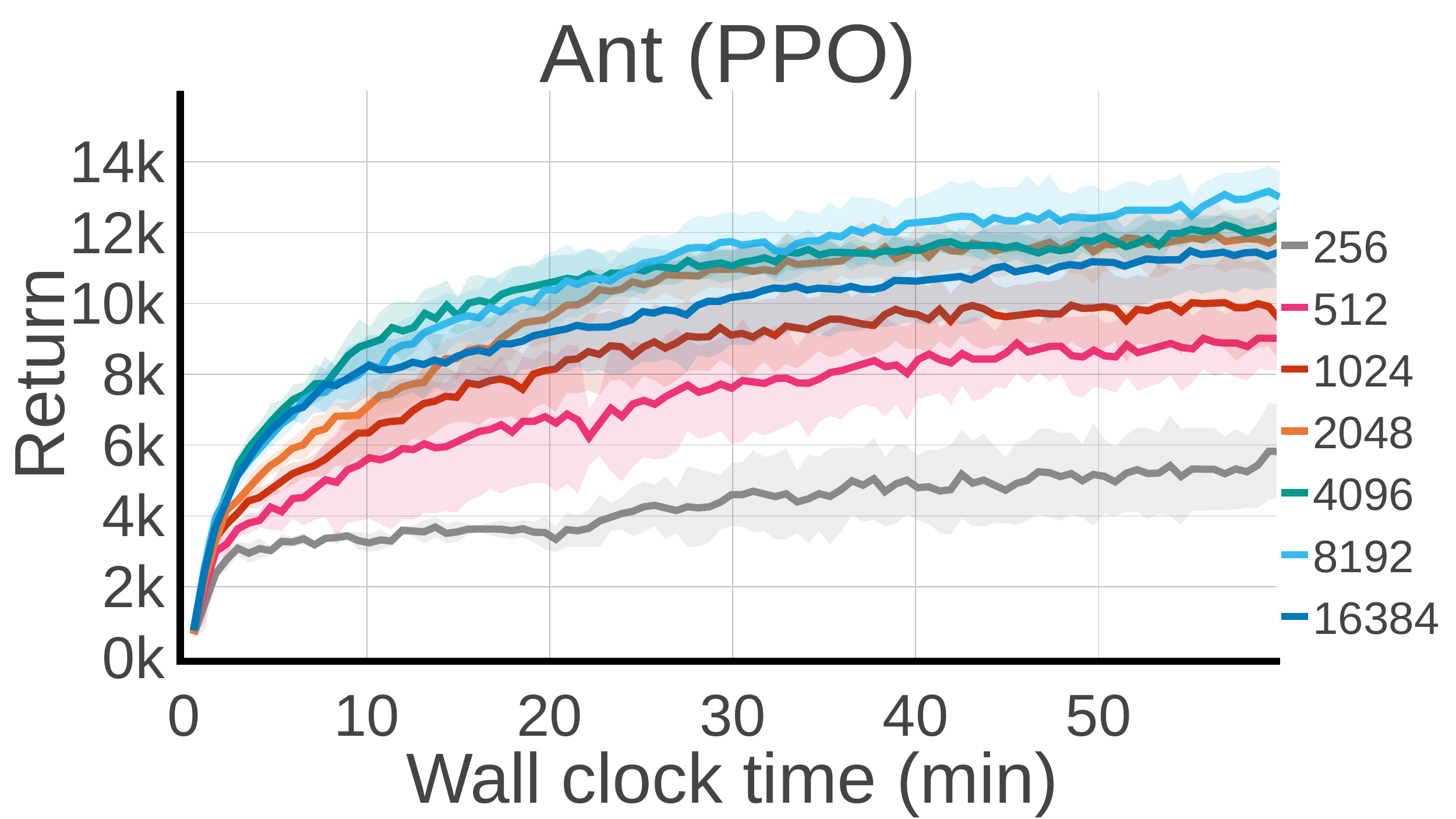}}
    \subfigure[]{\label{fig:ant_ddpg_num_envs}\includegraphics[width=0.49\linewidth]{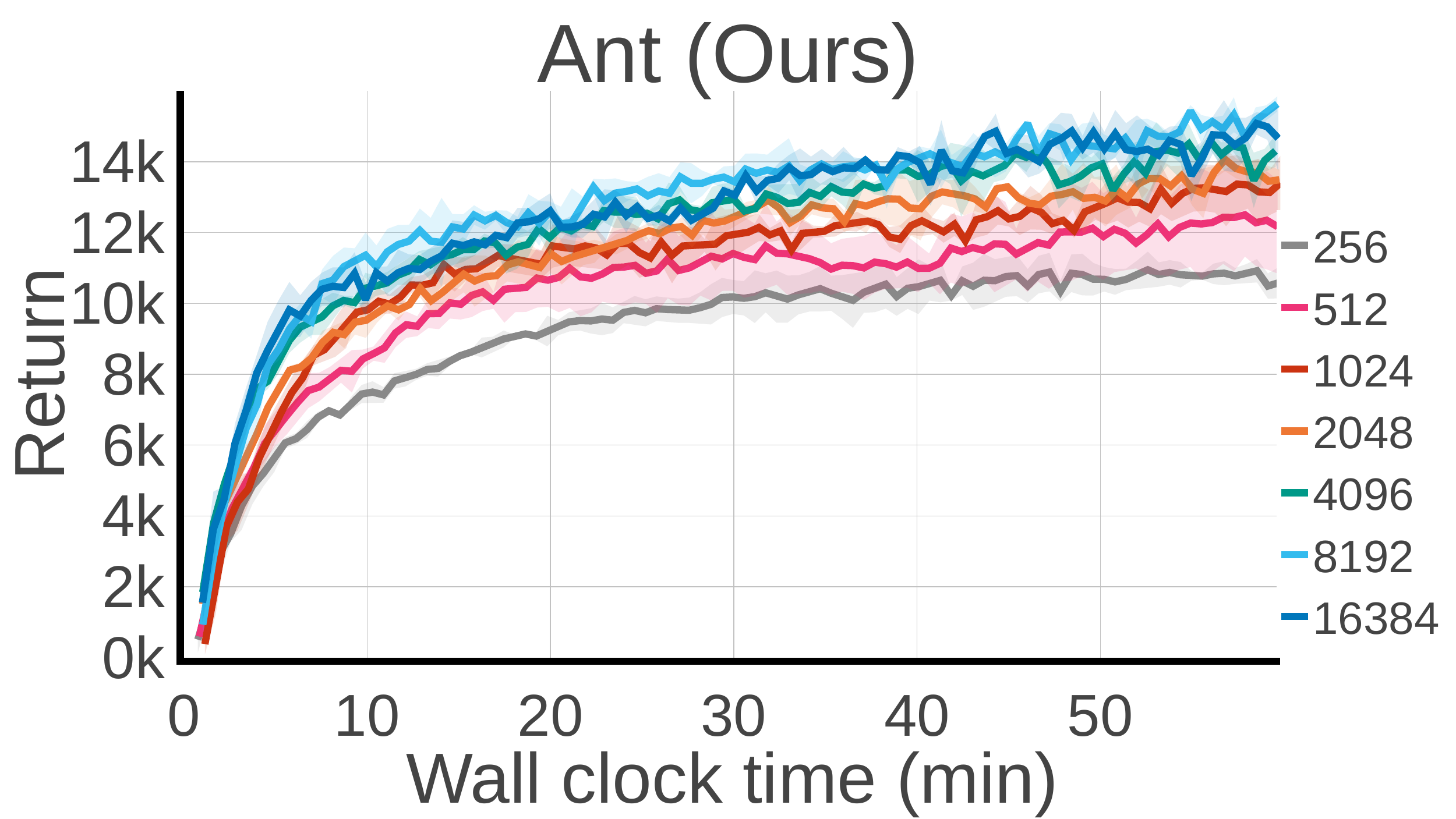}}\\
    \subfigure[]{\label{fig:shadow_ppo_num_envs}\includegraphics[width=0.49\linewidth]{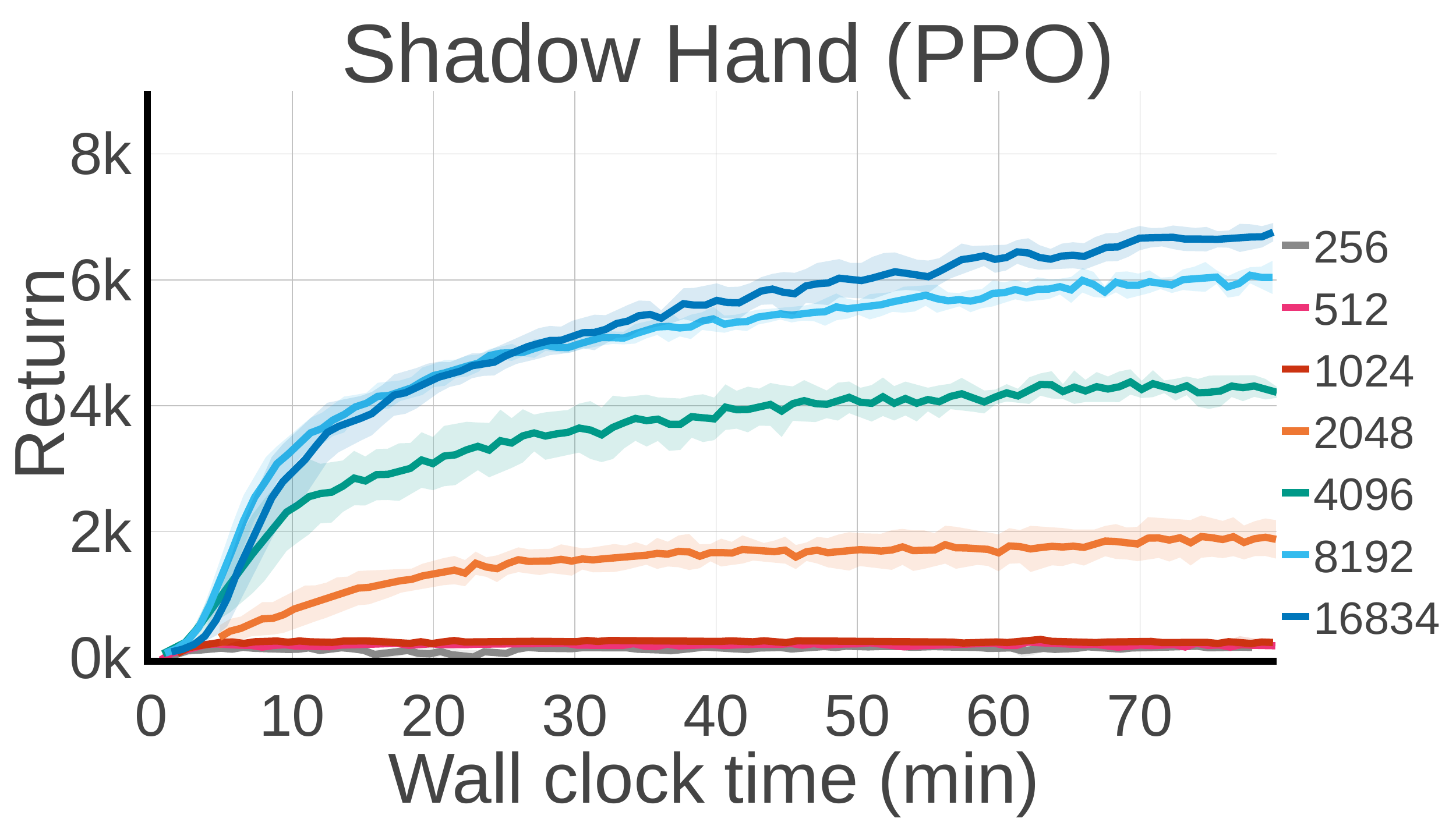}}
    \subfigure[]{\label{fig:shadow_ddpg_num_envs}\includegraphics[width=0.49\linewidth]{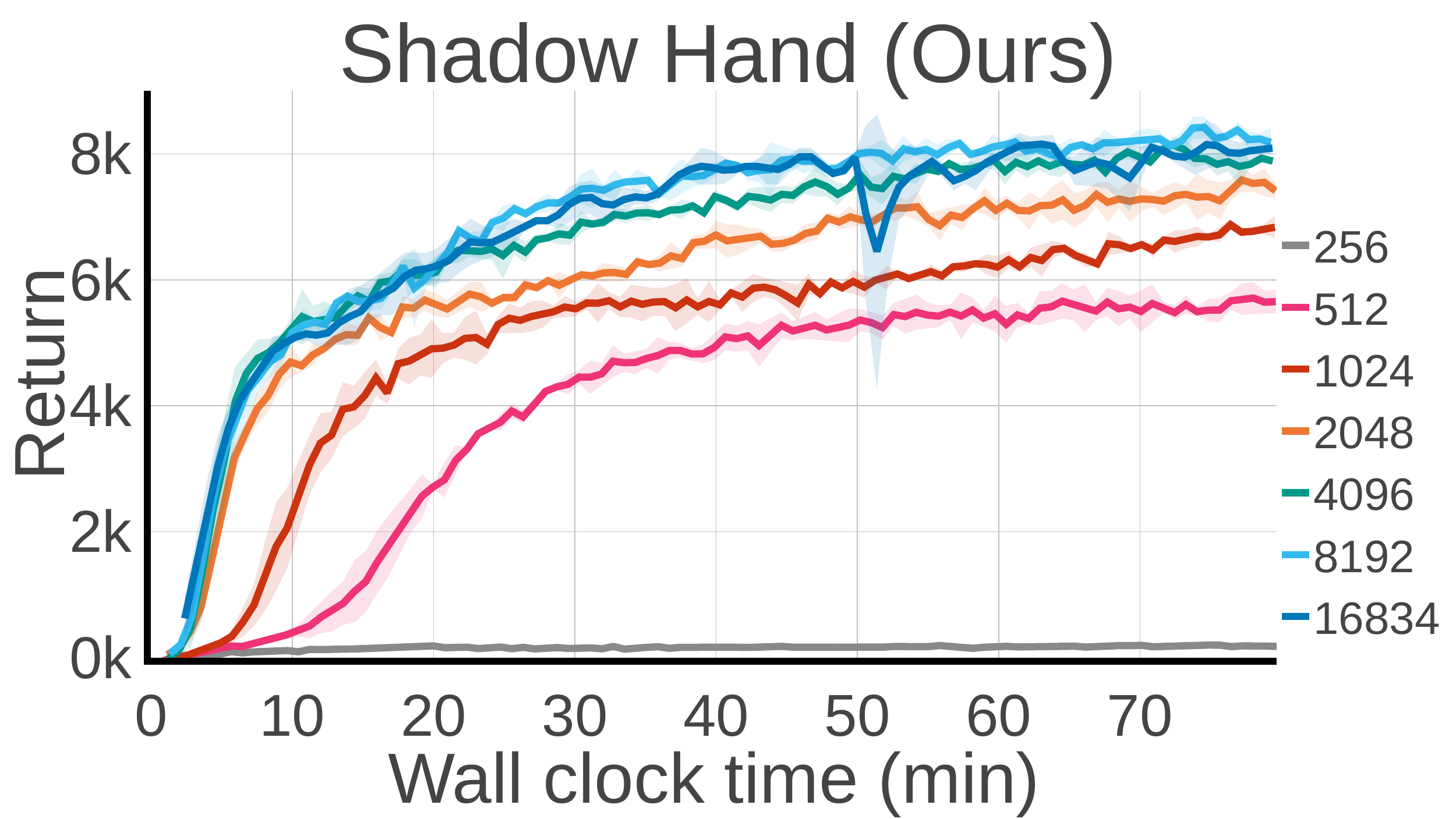}}
    \vspace{-2mm}
    \caption{We sweep over different numbers of environments ($N$) on both PPO and PQL (our method). Overall, PQL is less sensitive to the number of environments than PPO on both tasks.}
    \label{fig:num_envs}
\end{figure}

Previous works on distributed frameworks for RL~\cite{horgan2018distributed,espeholt2018impala} have shown how the learning performance is affected by the number of parallel environments $N$, with $N$ in the order of hundreds. GPU simulation enables running thousands of environments in parallel on a single workstation, and we anticipate that this will only improve with time. However, more parallel environments will only be beneficial if RL algorithms can exploit such data, i.e. if performance scales with more data. We, therefore, investigated how different algorithms scale with the number of environments ($N>>1000$, the biggest $N$ we experimented with is $16,384$). As shown in \figref{fig:num_envs}, both PQL and PPO benefit from using more environments in parallel. Moreover, the learning performance of PQL is relatively less sensitive to $N$ on the simple task (\ant), while on the hard task (\shadow), PPO's learning performance substantially drops as we decrease the number of environments. In contrast, our method (PQL) demonstrates stable and similar learning with all the different numbers of environments except when $N$ is very small ($N=256$) on \shadow, suggesting that PQL is more robust to changes in $N$.

\subsubsection{Effect of $\beta_{p:v}$ and $\beta_{a:v}$}
\label{subsec:effect_beta}
As discussed in \secref{subsec:ratio_balance}, explicitly controlling the $\beta_{a:v}$ and $\beta_{p:v}$ can help improve the learning performance and reduce the variance under different training conditions, such as fluctuated hardware utilization. If $\beta_{p:v}$ is larger, the policy updates more frequently than the value functions, potentially leading to policy overfitting to the stale value function, which in turn leads to poor exploration. If the policy updates much slower than the value function, the policy might lag behind the value function a lot, which hurts the learning speed. Similarly, if $\beta_{a:v}$ is larger, the \vlearner might need to wait for \actor to collect enough data, since the simulation speed cannot be changed, leading to slower learning. If $\beta_{a:v}$ is smaller, the value function updates more given the generated rollout data.

\begin{figure}[!tb]
    \centering
    \subfigure{\includegraphics[width=0.49\linewidth]{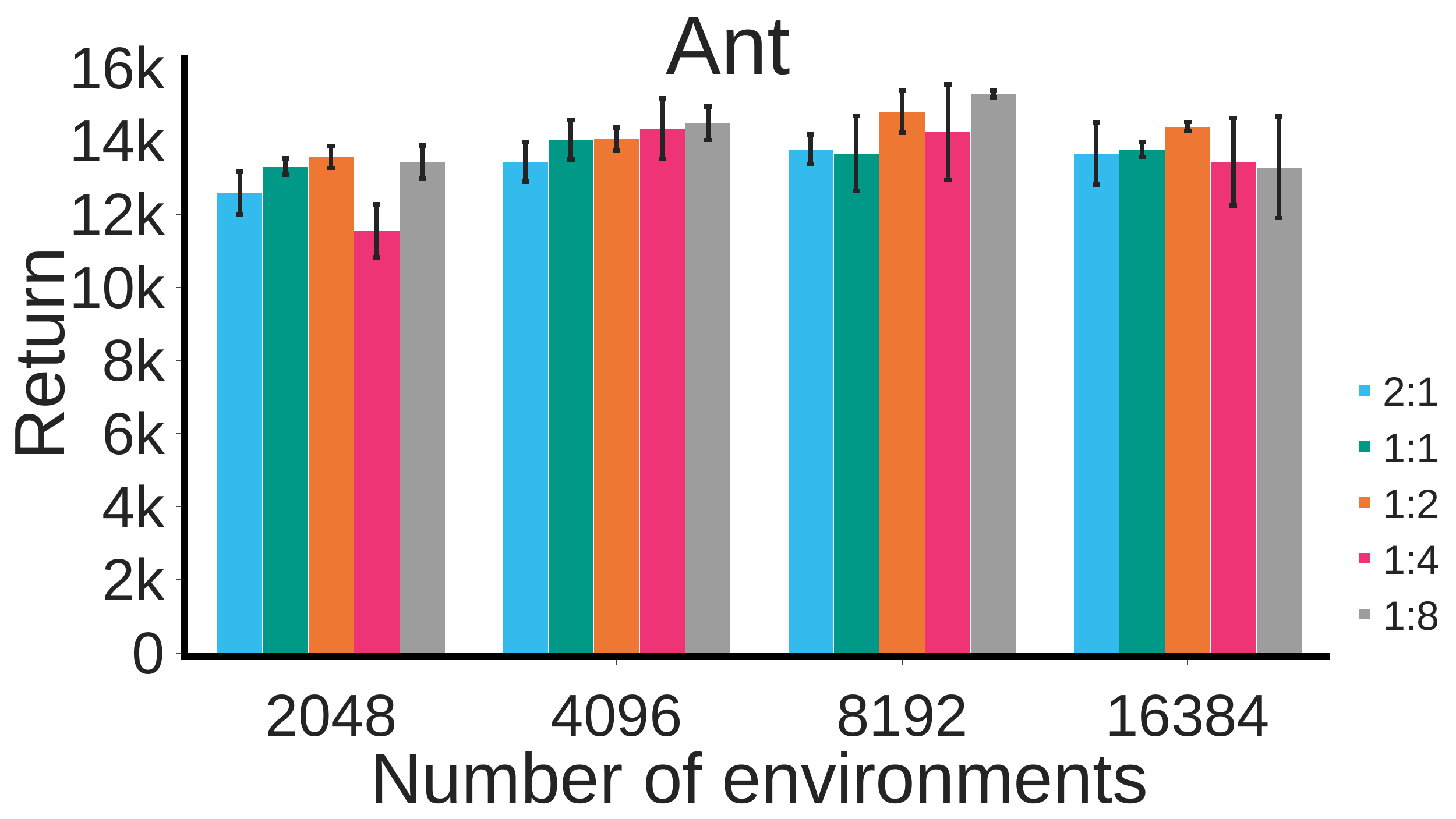}}
    \subfigure{\includegraphics[width=0.49\linewidth]{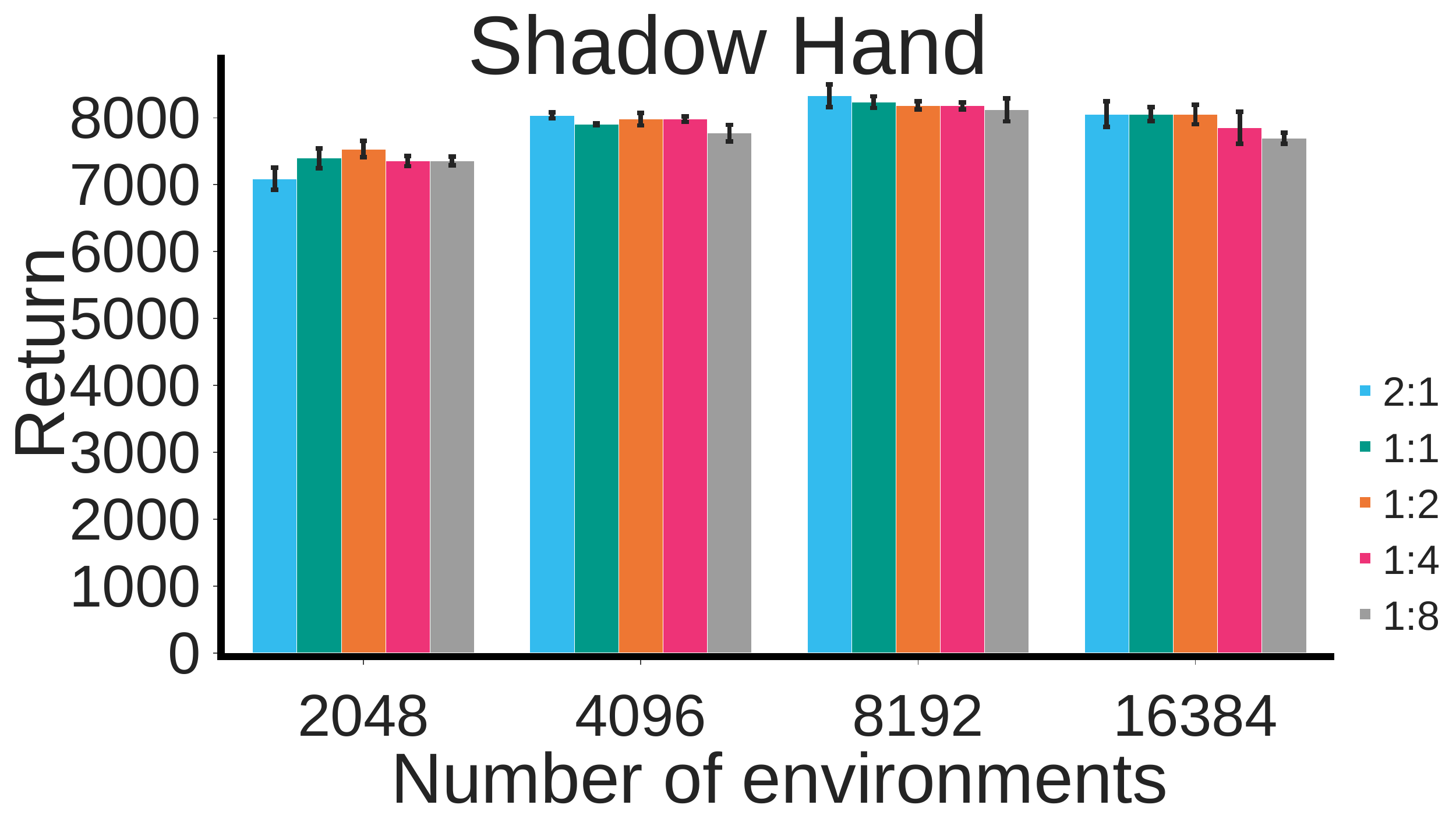}}
    \vspace{-2mm}
    \caption{We show the averaged returns in evaluation after a fixed amount of training time $\Delta T$. Across the set of different numbers of environments we experimented with ($2048, 4096, 8192, 16384$), we found that setting $\beta_{p:v}=1:2$ generally works well. $\Delta T=60$ mins for \ant, and $\Delta T=80$ mins for \shadow. The complete learning curves are in \figref{fig:app_actor_critic_ratio}. }
    \label{fig:actor_critic_ratio}
\end{figure}

\begin{figure}[!tb]
\vspace{-0.4cm}
    \centering
    \subfigure{\includegraphics[width=0.49\linewidth]{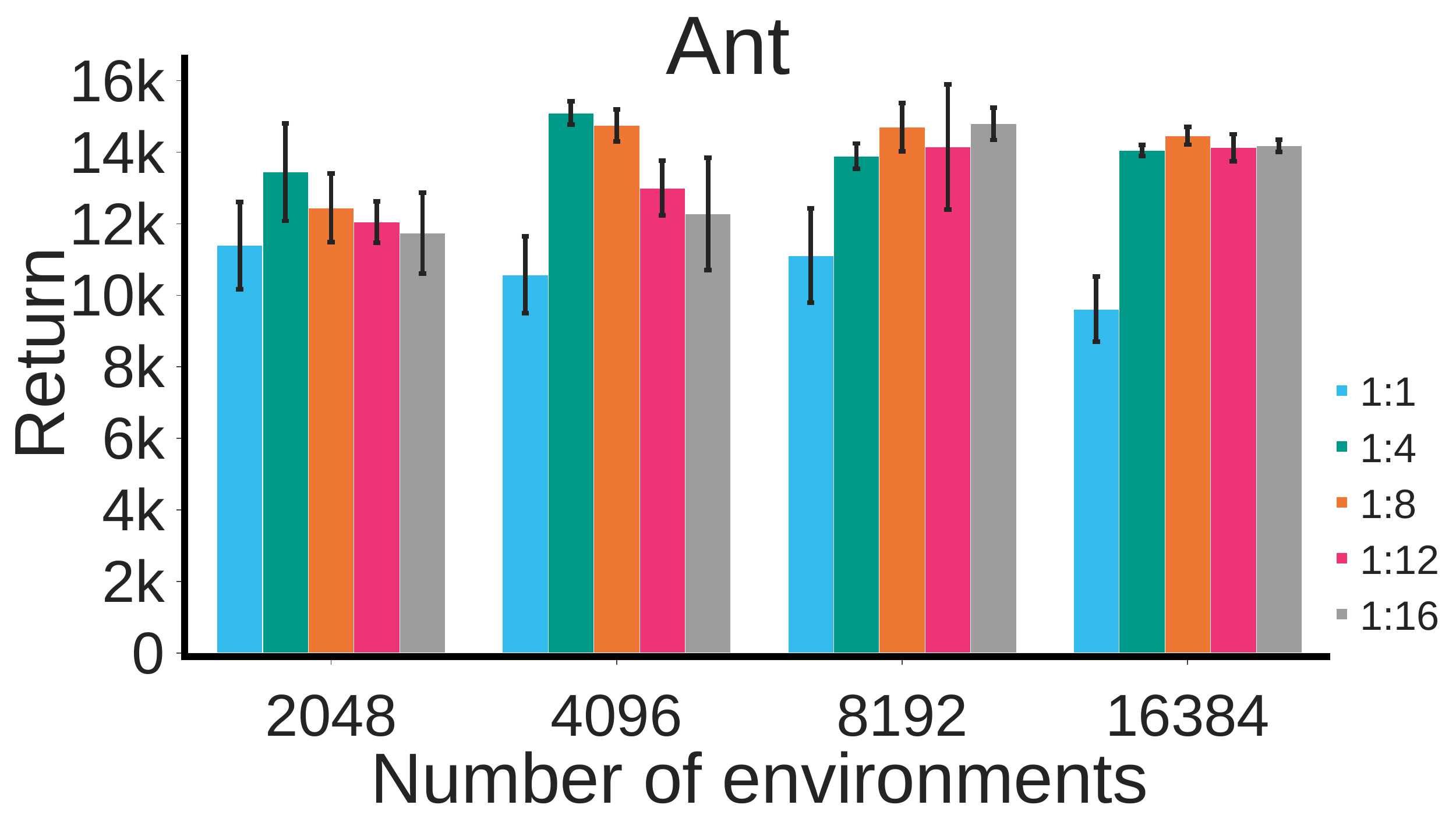}}
    \subfigure{\includegraphics[width=0.49\linewidth]{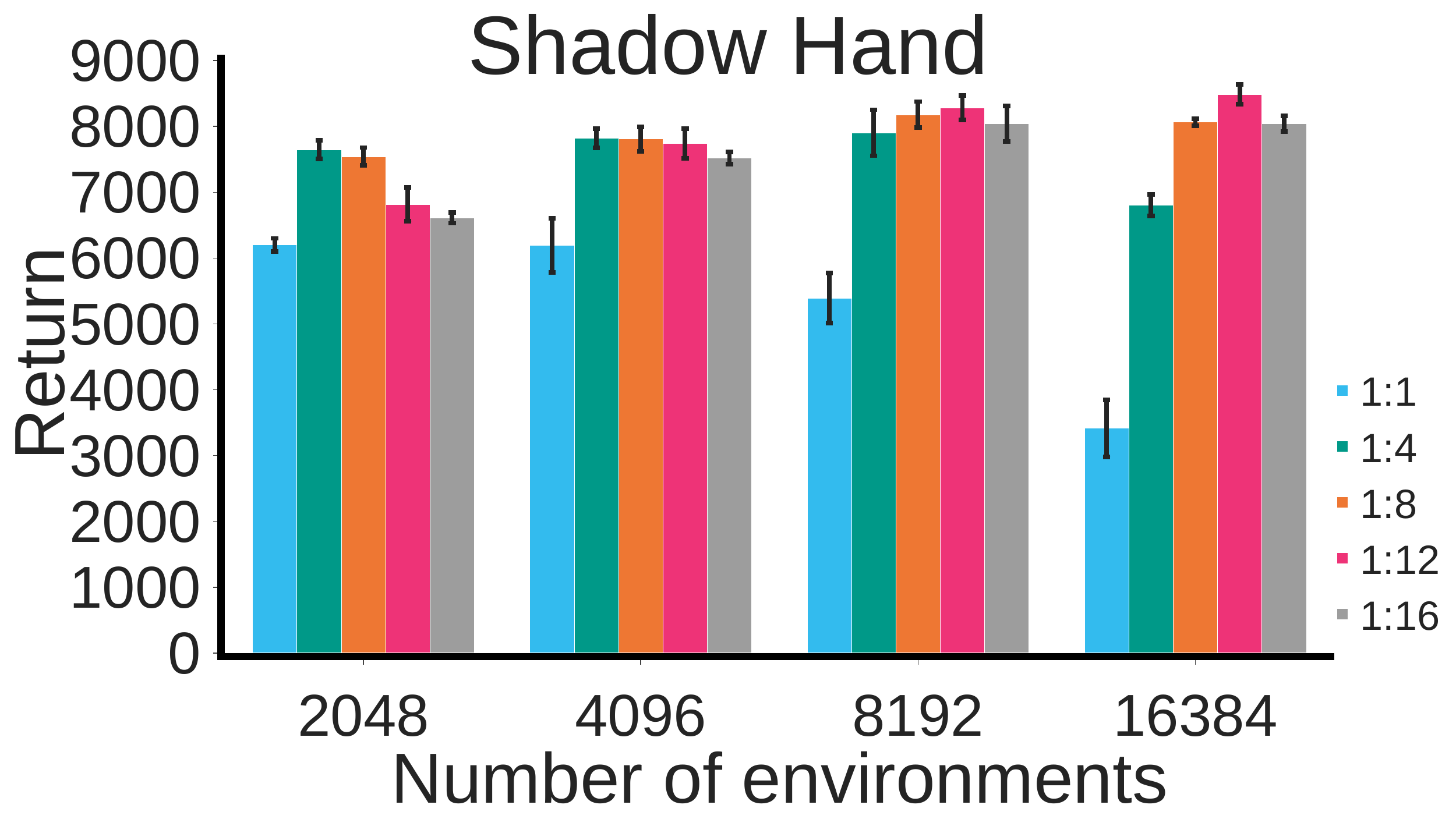}}
    \caption{Given different values of $N$, we show the effect of different $\beta_{a:v}$. An overall trend we observe is that as $N$ gets bigger, it's more beneficial to update the critic more frequently. We also found that $\beta_{a:v}=1:8$ generally works well given different $N$ values. So one can set $\beta_{a:v}=1:8$ as a good initial value, and tune it if necessary on new tasks.}
    \label{fig:worker_critic_ratio}
\end{figure}

To qualitatively assess the effects of different $\beta_{a:v}$ and $\beta_{p:v}$, we sweep over a range of values for these two hyper-parameters and compare them in \figref{fig:actor_critic_ratio} and \figref{fig:worker_critic_ratio}. \figref{fig:actor_critic_ratio} shows that PQL is relatively robust to a wide range of $\beta_{p:v}$ values, which means this hyper-parameter would require little tuning. We use $\beta_{p:v}=1:2$ as the default value in our experiments shown in the paper. This ratio value is consistent with prior works~\citep{fujimoto2018addressing, yang2022overcoming}, but our findings are in the context of parallel training of policy and value function. \figref{fig:worker_critic_ratio} shows that $\beta_{a:v}$ has a greater impact on the learning performance. An overall trend is that if we increase the number of environments, then we need to have \vlearner update the $Q$ functions more times. For example, on \shadow, $\beta_{a:v}=1:4$ performs the best when $N=2048$ and $N=4096$. But when $N=8192$ and $N=16384$, $\beta_{a:v}=1:12$ performs the best. We use $\beta_{a:v}=1:8$ by default as it achieves a good performance across different $N$ values. In summary, \figref{fig:actor_critic_ratio} and \figref{fig:worker_critic_ratio} show that $\beta_{p:v}$ and $\beta_{a:v}$ do affect the performance with a varied number of environments. We suggest setting $\beta_{p:v}=1:2, \beta_{a:v}=1:8$ as a good starting point for new tasks and tune them if necessary, as these are the values we found work well on six different tasks with different numbers of environments. In addition, in \secref{appsec:extra_exps}, we show that adding the speed ratio control ($\beta_{a:v}$ and $\beta_{p:v}$) is beneficial for balancing the computing resources used by each process when resources are limited.

\subsubsection{Effect of batch size}
With many parallel environments ($N$), a significant amount of data is generated quickly. While it is easy to increase $N$ from hundreds to tens of thousands in Isaac Gym on a single GPU, it is infeasible to increase the replay buffer size by 100 times due to limited GPU memory or CPU RAM (if the data is stored on the CPU). Consequently, the replay buffer is frequently overwritten, meaning that each collected sample may not be used efficiently. One way to efficiently utilize large amounts of changing data is to increase the batch size. To determine how much increase in batch size is necessary for Q-learning with a limited-capacity replay buffer to take advantage of the large amounts of incoming data, we investigated the relationship between performance and batch size. Many prior works have shown that using a large batch size can improve network performance, such as in contrastive learning settings~\citep{grill2020bootstrap, chen2020simple}. In our work, we found that large-batch training can notably improve the learning speed in off-policy RL for massively parallel simulations, as shown in \figref{fig:batch_size}. However, if the batch size is too big, the learning speed can be slowed down. This is because GPUs have a limited number of CUDA cores, and it takes more time to process a very big batch of data once the batch size is above some threshold value, which is another underlying trade-off.

\begin{figure}[!tb]
    \centering
    \subfigure[]{\label{fig:ant_bs}\includegraphics[width=0.49\linewidth]{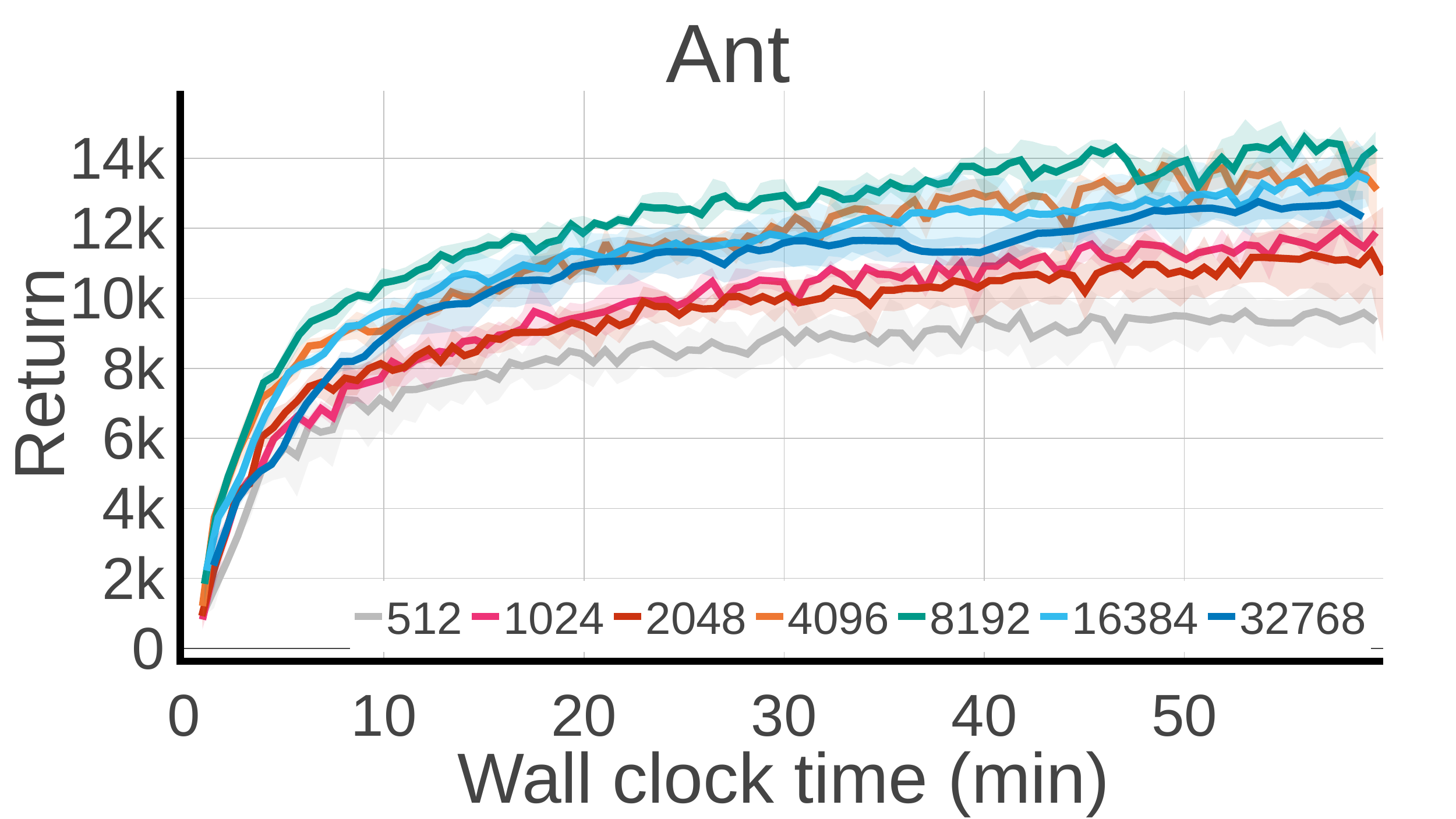}}
    \subfigure[]{\label{fig:shadow_bs}\includegraphics[width=0.49\linewidth]{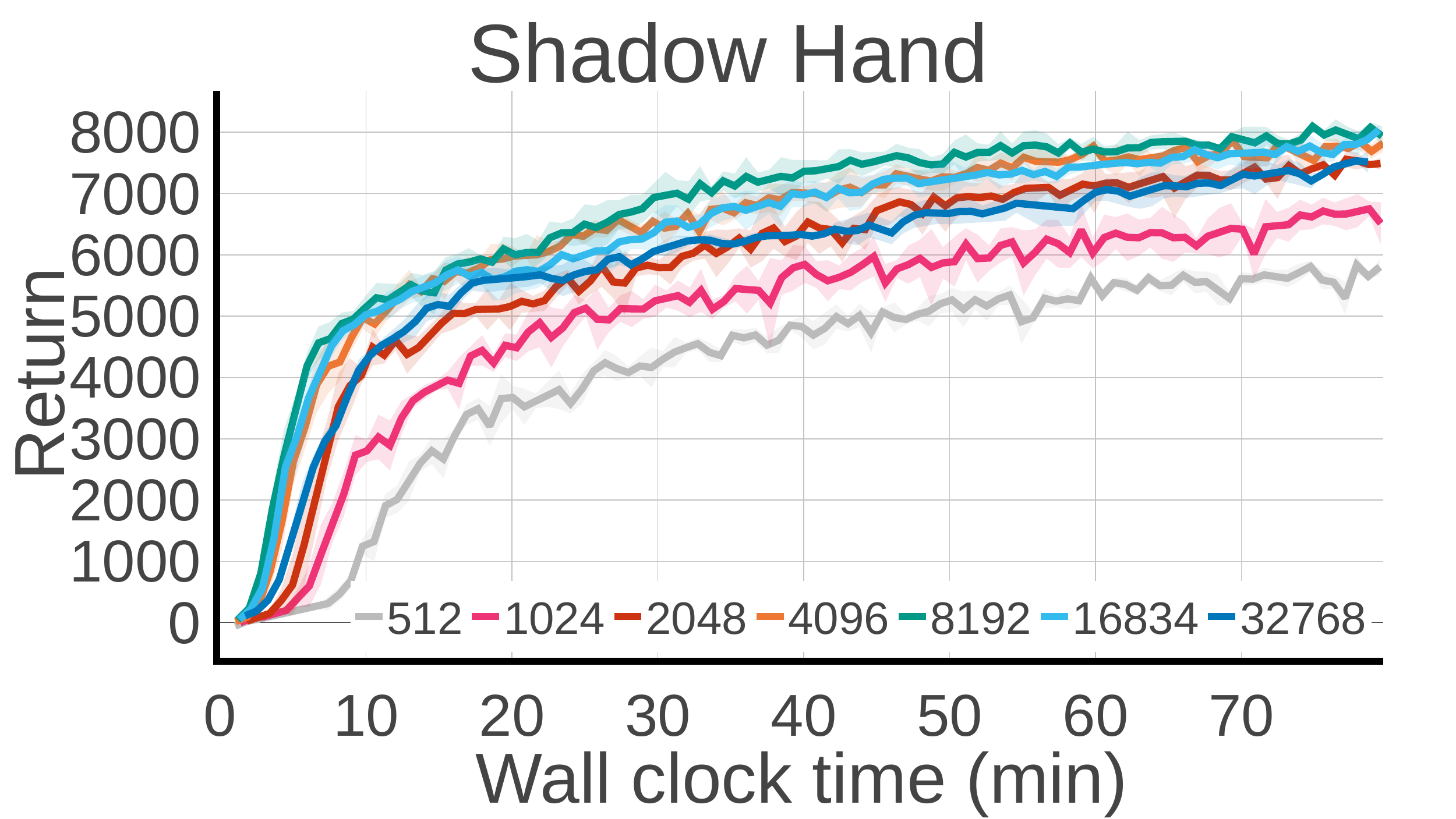}}
    \vspace{-2mm}
    \caption{Effect of different batch sizes. Small batch size usually leads to slower learning. If the batch size is too big, the policy learning can slow down because GPUs have limited cores and it takes more time to process a very big batch of data.}
    \label{fig:batch_size}
\end{figure}

\begin{figure}[t!]
    \centering
    \subfigure[]{\label{fig:ant_replay}\includegraphics[width=0.49\linewidth]{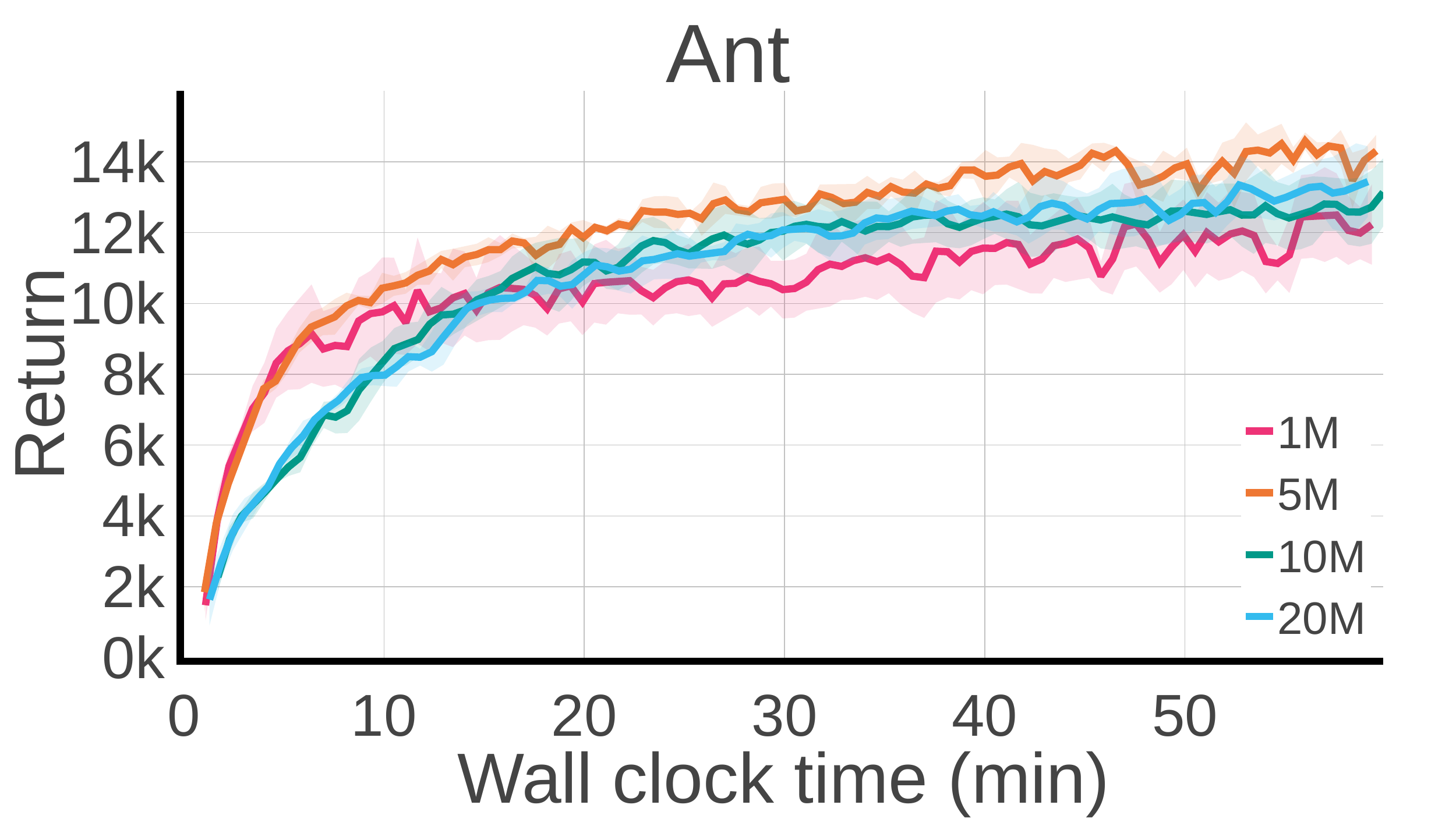}}
    \subfigure[]{\label{fig:shadow_replay}\includegraphics[width=0.49\linewidth]{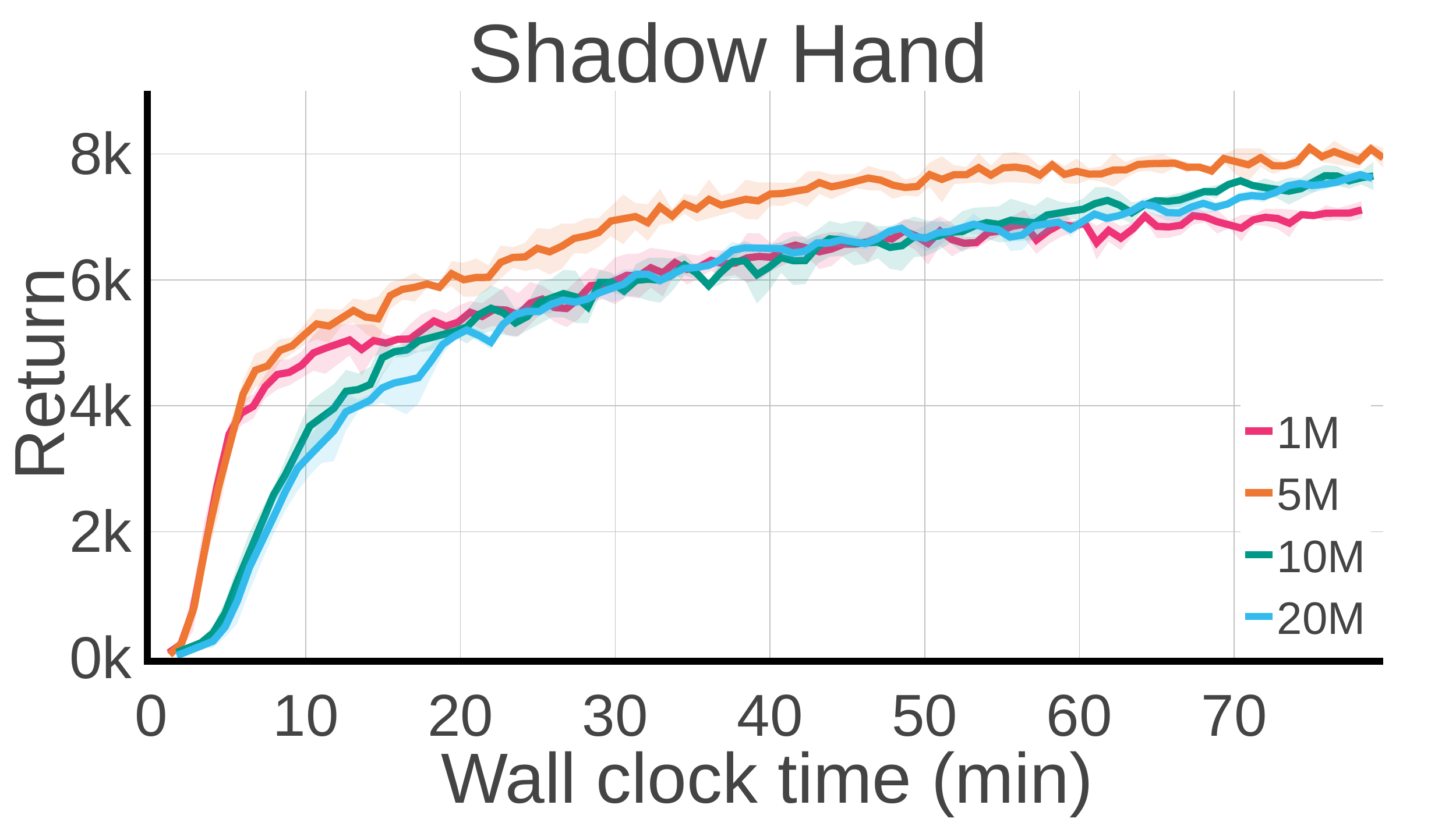}}\\
    \subfigure[]{\label{fig:ant_num_gpu}\includegraphics[width=0.49\linewidth]{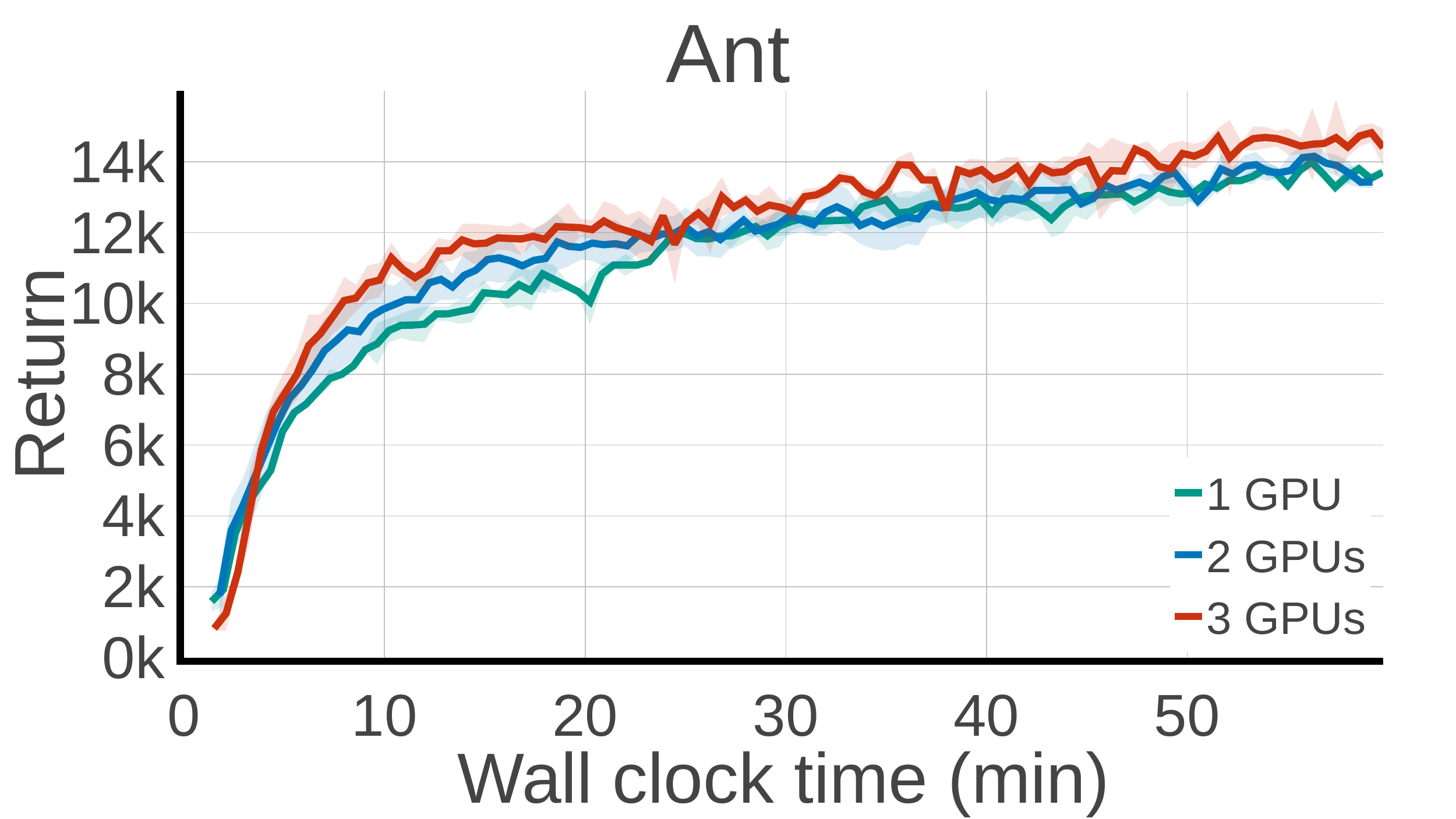}}
    \subfigure[]{\label{fig:shadow_num_gpu}\includegraphics[width=0.49\linewidth]{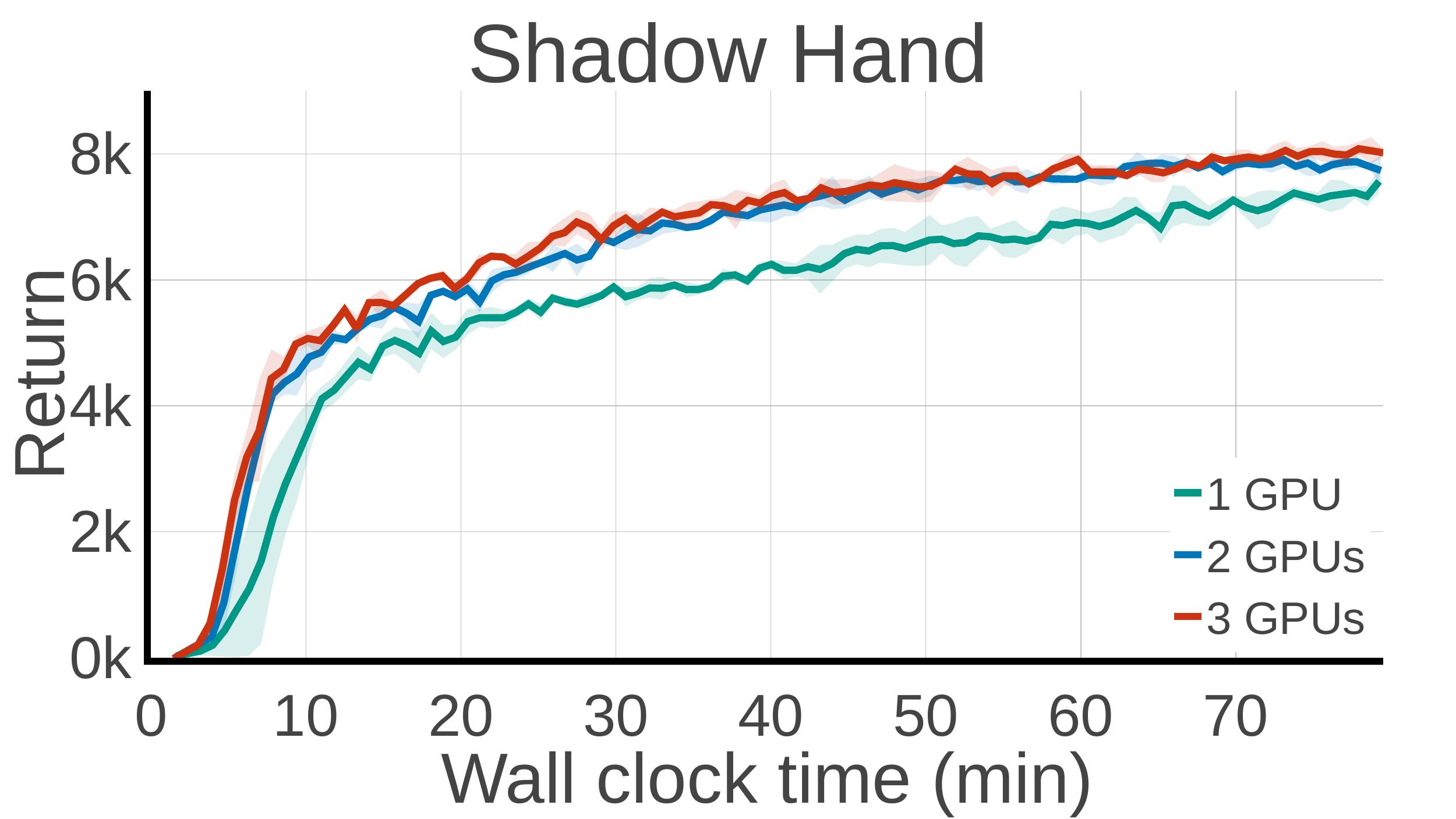}}
    \vspace{-2mm}
    \caption{\textbf{(a)} and \textbf{(b)}: effect of different replay buffer size. \textbf{(c)} and \textbf{(d)}: effect of number of GPUs used for running PQL. PQL can be deployed on a flexible number of GPUs. In complex tasks such as \shadow, it is beneficial to have at least 2 GPUs where the \actor runs on a separate GPU as the simulation itself consumes more GPU compute as the task complexity increases.}
    \label{fig:replay_gpu_num}
\end{figure}

\subsubsection{Effect of Replay buffer size}
As discussed in \secref{sec:intro}, when dealing with tens of thousands of parallel environments, the replay buffer with normal capacity (e.g. 1M) gets a full refresh for every several hundreds of environment steps. This means that the replay buffer will not contain too much historical data. Apriori, one might think off-policy methods will fail in this case and that we need to proportionally increase the replay buffer size to store more experience data. However, surprisingly, we empirically found that even with thousands of parallel environments, having a ``small" replay buffer (1M or 5M) can still lead to good performance. In \figref{fig:replay_gpu_num}, we show how the learning curves change as we vary the buffer capacity. We can see that, in all cases, the policies learn well.  We hypothesize that PQL still works well in this case because a large number of parallel environments can generate diverse enough data in a few environment steps. Moreover, $|B|\in\{1, 5\}$M leads to faster policy learning at the beginning of the training than $|B|=\{10, 20\}$M. We hypothesize that this is because a smaller replay buffer allows the old and less informative samples to be replaced much faster, which is more important in the early stages of training. The converged performances are similar when $|B|=\{5, 10, 20\}$M. However, the converged performance is slightly worse when $|B|=1$M. This may be because a small replay buffer, which discards new data too quickly, can have a negative impact on learning in the end.

\subsubsection{Number of GPUs} 
Nowadays, it is common to have workstations with multiple GPUs. Running different processes on separate GPUs can potentially speed up learning. Our PQL scheme can adapt to different numbers of GPUs available on a workstation. Specifically, \actor, \plearner, and \vlearner can be placed on any GPU. To investigate the performance variation when we distribute the three components across different numbers of GPUs, we conducted experiments with three scenarios on Tesla A100 GPUs: (1) place all three processes on the same GPU, (2) place the \actor on one GPU, \plearner and \vlearner on another GPU, (3) place the \actor, \plearner, \vlearner on a different GPU respectively. In the two-GPU case, we allocate \actor to a dedicated GPU because simulating many tasks with a large number of environments can cause high GPU utilization. As shown in \figref{fig:replay_gpu_num}, our PQL scheme works well in all three scenarios with one, two, or three GPUs. When the task becomes more complex like \shadow, the simulation takes much more computation and time. Putting all three processes will slow down each one of them due to full GPU utilization, which is why we see a bigger gap between the 2-GPU or 3-GPU training and 1-GPU training on \shadow. Therefore, it is beneficial to place the \actor on one GPU and \plearner and \vlearner on other GPUs.

\subsection{Additional Tasks}
\label{sec:vision-based}
\paragraph{Vision-based \ball task}
Simulating vision-based tasks is much slower and more demanding on the GPU as each simulation step involves both the physics simulation and image rendering. To demonstrate the generality of our scheme in operating in this practical setting of vision-based training, we consider a vision-based \ball task~\cite{makoviychuk2021isaac}. 

Since directly learning a vision-based policy with RL is time-consuming, we use the idea of asymmetric actor-critic learning~\citep{pinto2017asymmetric} to speed up vision policy learning. Image data is compressed using the \textit{lz4} library to reduce the bandwidth requirement and communication overhead. More setup details are in \appref{appsubsec:vision}. As shown in \figref{fig:vision}, PQL achieves better sample efficiency and higher final performance than PPO with $N=1024$ parallel environments. 

\begin{figure}[!tb]
    \centering
    \subfigure[]{\label{fig:dclaw_shot}\includegraphics[width=0.49\linewidth, height=0.25\linewidth]{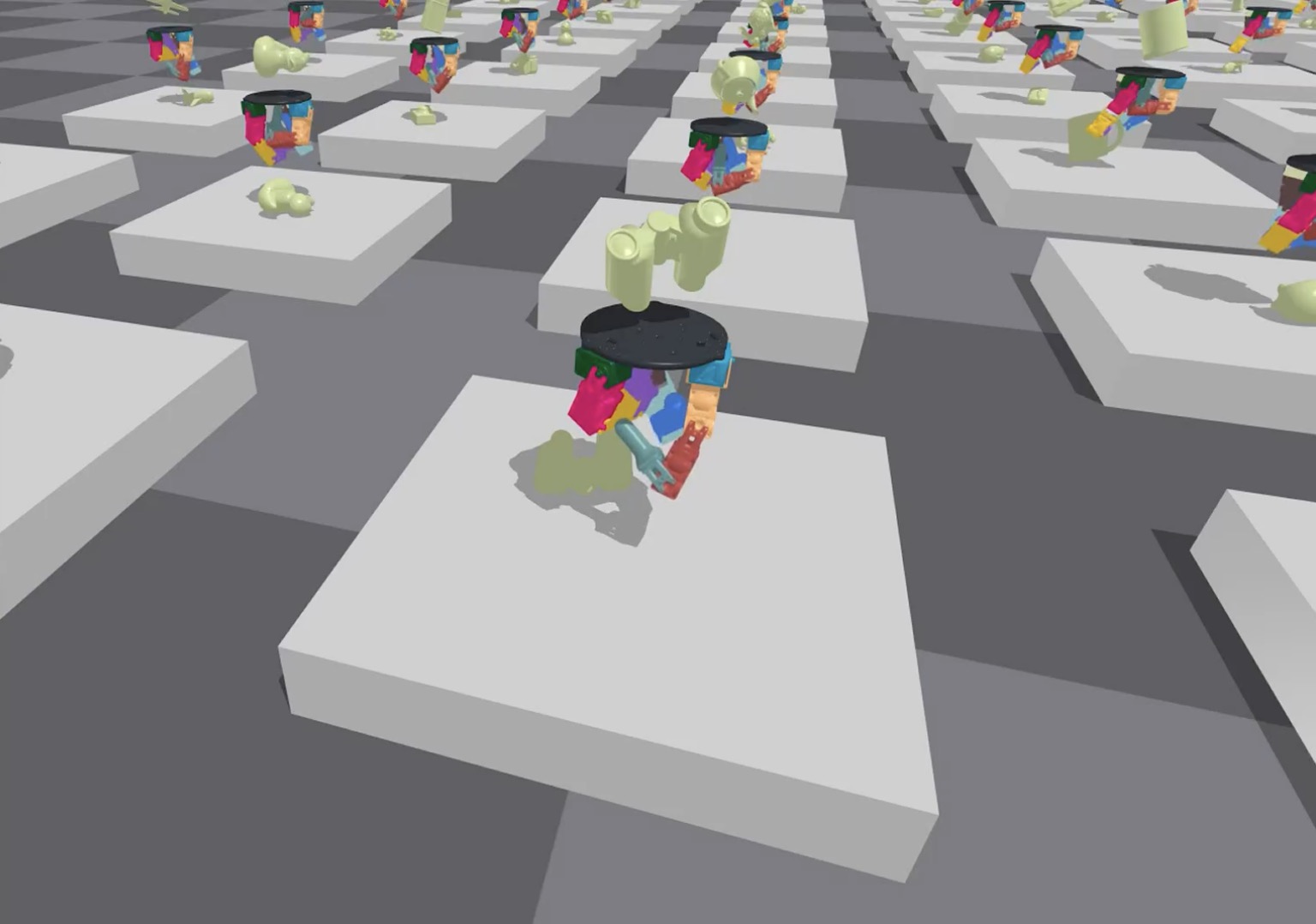}}
    \subfigure[]{\label{fig:dclaw}\includegraphics[width=0.49\linewidth]{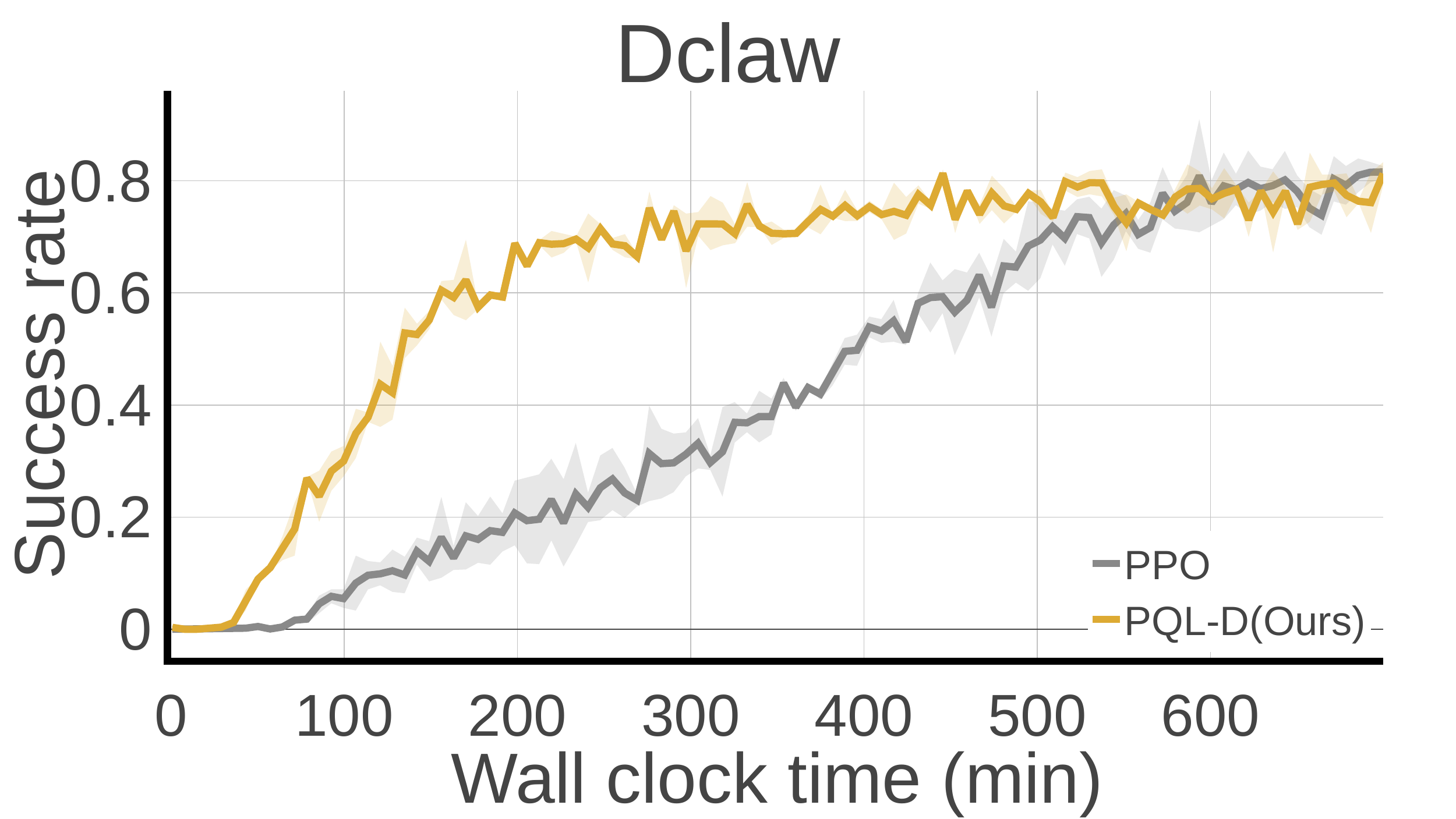}}
    \vspace{-2mm}
    \caption{\textbf{(a)}: \dcalw task. \textbf{(b)}: We compared our proposed PQL-D method with PPO.}
    \vspace{-6mm}
    \label{fig:batch_size}
\end{figure}

\paragraph{Reorient hundreds of objects with a \dcalw}
We conducted further experiments on a contact-rich dexterous manipulation task~\citep{chen2022visual}, \dcalw, as shown in \figref{fig:dclaw_shot}. This task is much more challenging than the \shadow task and \allegro task because it needs to learn to reorient hundreds of different objects with a single policy. Furthermore, the control frequency (12Hz) is much lower than the default control frequency (60Hz) used in all six proposed benchmark tasks. This means that the simulation takes much longer to run between each policy command step, and the \actor process will be much slower. In \figref{fig:dclaw}, we observe that our method reaches $70\%$ success rate around $200$ minutes, which is approximately $3$ times faster than PPO.

\section{Discussion and Future Work}
We present a scheme \method for scaling up off-policy methods with tens of thousands of parallel environments on a single workstation. Our method achieves state-of-the-art results on the Isaac Gym benchmark tasks in terms of the training wall clock time. The driving force behind this success is the parallelization of data collection, policy function learning, and value function learning. We provide a mechanism to balance and control the speed in different processes, which leads to better and more stable performance across different hardware conditions or when the GPU resource is limited. Although PPO requires a large number of environments to work on complex tasks such as \shadow, PQL is more lenient on the number of environments and works well on a wide range of different numbers of environments. With a large number of parallel environments, it is beneficial to use a big batch size for training agents, with the caveat that if the batch size is too big, it might take the GPU more time to process the batch data and lead to a slowdown in policy learning. We also found using different exploration scales in different environments achieves better or similar performance compared to a carefully-tuned exploration scale in all parallel environments, which means we need less hyper-parameter tuning. Even though the number of environments is $1000\times$ more, we did not find it necessary to use a replay buffer that is $1000\times$ bigger. In fact, a replay buffer with a capacity of $5$M transitions is sufficient for our experiments even with $16843$ parallel environments. Our scheme's hardware requirements are flexible and work well with different numbers of GPUs and various GPU models.

In this paper, we experimented with default task configurations for the Isaac Gym benchmark tasks. Therefore, the reported results are for tasks without extensive domain randomization. The investigation of how well PQL performs in the presence of extensive domain randomization is left for future work. Another interesting direction is to explore better sampling strategies for the replay buffer. PQL does not use techniques such as prioritized experience replay~\citep{schaul2015prioritized}, which could improve sample efficiency but significantly hurt wall-clock time efficiency due to massive amount of collected data. Therefore, new strategies should be considered, such as rejecting samples given the massive amount of data. It would also be practical to study different exploration strategies that can take advantage of parallel environments. When using PQL in other tasks, we have observed that if the agent is in joint position control mode, mixed exploration strategy tends to work better when the action space is defined on relative joint position control rather than on absolute joint position control. We hypothesize that exploring the full action space ($[-1, 1]$) in absolute joint position control leads to many useless explorations, and thus data from a significant portion of the parallel environments may not provide much value. In such cases, it would be interesting to investigate how to adaptively change the maximum range of exploration noise throughout the training process. Lastly, our scheme can be easily extended to a system with multiple parallel learners or value learners given the decoupling of policy and value learning. In this case, it would be interesting to apply ensemble methods or evolutionary strategies to further exploit the massive amount of data.

\section{Acknowledgements} 
\label{sec:acknowledgments}

We thank the members of the Improbable AI Lab for their helpful feedback and insightful discussions. The authors acknowledge the MIT SuperCloud and Lincoln Laboratory Supercomputing Center for providing HPC resources that have contributed to the research results reported within this paper. This research was supported in part by the Hyundai Motor Company, an Amazon Research Award, Google cloud credits provided as part of Google-MIT support, DARPA Machine Common Sense Program, ARO MURI under Grant Number W911NF-21-1-0328, ONR MURI under Grant Number N00014-22-1-2740.

\section{Contributions} 
\label{sec:contribution}
\textbf{Zechu Li} and \textbf{Tao Chen} jointly developed the framework. \textbf{Zechu Li} implemented the framework and the baselines, and ran all the experiments. \textbf{Tao Chen} reviewed and notably improved the code quality and played a primary role in paper writing. \textbf{Zhang-Wei Hong} and \textbf{Anurag Ajay} were involved in research discussions. \textbf{Pulkit Agrawal} conceived the project, contributed to some of the main research ideas, provided feedback on the writing, and participated in research discussions.

\bibliography{ref}
\bibliographystyle{icml2023}

\newpage
\appendix
\onecolumn
\begin{appendices}
\setcounter{figure}{0}
\setcounter{table}{0}
\setcounter{footnote}{0}
\renewcommand{\thefigure}{\Alph{section}.\arabic{figure}}
\renewcommand{\thetable}{\Alph{section}.\arabic{table}}

\section{Pseudo Code}
\label{appsec:code}

\begin{algorithm}
\caption{\actor Process (main process)}
\label{alg:actor}
\begin{algorithmic}
    \FOR{$n=1:W_a$}
    \STATE $\pi \gets$ policy network from \plearner process
    \STATE Initialize an empty buffer $B=\phi$
    \FOR{$t=1:H$}
    \STATE $\bm{a}_t\gets \pi(\bm{s}_t)$ with mixed exploration noise
    \STATE $(\bm{r}_t, \bm{s}_{t+1})\gets$ \textbf{envs}.step($\bm{a}_t$)
    \STATE $B=B\cup{\{\bm{s}_t, \bm{a}_t, \bm{r}_t, \bm{s}_{t+1}\}}$
    \ENDFOR
    \STATE $Q_1, Q_2\gets$ Q functions from \vlearner process
    \STATE send $B, \pi$ to \vlearner, send $\{s_t\}$ in $B$, $Q_1, Q_2$ to \plearner
    \STATE sleep for $t_a$ seconds to satisfy $\beta_{a:v}$ 
    \ENDFOR
\end{algorithmic}
\end{algorithm}

\begin{algorithm}
\caption{\plearner Process}
\label{alg:plearner}
\begin{algorithmic}
    \STATE Initialize an empty buffer $B_p=\phi$
    \FOR{$n=1:W_p$}
    \IF {new data received}
    \STATE ${\{s_t\}}\gets$ from \actor process
    \STATE $Q_1, Q_2\gets$ from \actor process
    \STATE $B=B\cup{\{s_t\}}$
    \ENDIF
    \STATE sample a batch of $\{s_t\}$
    \STATE update $\pi$ by maximizing the $\min_{i=1,2}Q_i(s_t, \pi(s_t))$
    \STATE sleep for $t_p$ seconds to satisfy $\beta_{p:v}$ 
    \ENDFOR
\end{algorithmic}
\end{algorithm}

\begin{algorithm}[H]
\caption{\vlearner Process}
\label{alg:vlearner}
\begin{algorithmic}
    \STATE Initialize an empty buffer $B_v=\phi$
    \FOR{$n=1:W_v$}
    \IF {new data received}
    \STATE ${\{s_t, a_t, r_t, s_{t+1}\}}\gets$ from \actor process
    \STATE $\pi\gets$ from \actor process
    \STATE $Q_1, Q_2\gets$ from \actor process
    \STATE $B=B\cup{\{s_t\}}$
    \ENDIF
    \STATE sample a batch of $\{s_t, a_t, r_t, s_{t+1}\}$
    \STATE update $Q_1, Q_2$ by minimizing the mean-squared Bellman error (with Double Q-learning)
    \STATE sleep for $t_v$ seconds to satisfy $\beta_{p:v}, \beta_{a:v}$ 
    \ENDFOR
\end{algorithmic}
\end{algorithm}

\section{Training setups}
\subsection{Hyper-parameters}
\label{appsubsec:hyper}

We use the hyper-parameter values shown in \tblref{tbl:hyper} and the reward scaling shown in \tblref{tbl:reward_scale} for all the experiments unless otherwise specified. As for PPO, we use the same hyperparameter setup in ~\citet{makoviychuk2021isaac}.

\begin{table}[!h]
\centering
\caption{Hyper-parameter setup for six Isaac Gym benchmark tasks}
\label{tbl:hyper}
\begin{tabular}{llll} 
\toprule
Hyper-parameter                          & PQL(ours) & DDPG     & SAC       \\ 
\midrule
Num. Environments                        & 4,096      & 4,096     & 4,096      \\
Critic Learning Rate                     & $5 \times 10^{-4}$  & $5 \times 10^{-4}$ & $5 \times 10^{-4}$  \\
Actor Learning Rate                      & $5 \times 10^{-4}$  & $5 \times 10^{-4}$ & $5 \times 10^{-4}$  \\
Learnable Entropy Coefficient            & - & - & True \\
Optimizer                                & Adam      & Adam     & Adam      \\
Target Update Rate ($\tau$) & $5 \times 10^{-2}$  & $5 \times 10^{-2}$ & $5 \times 10^{-2}$  \\
Batch Size                               & 8,192      & 8,192     & 8,192      \\
Num. Epochs ($\beta_{a:v}$)              & 8         & 8        & 8         \\
Discount Factor($\gamma$)                & 0.99      & 0.99     & 0.99      \\
Normalized Observations                  & True      & True     & True      \\
Gradient Clipping                        & 0.5       & 0.5      & 0.5    \\
Exploration Policy                       & Mix       & Mix      & -      \\
$N$-step target                          & 3         & 3        & 3         \\
Warm-up Steps                            & 32        & 32       & 32        \\
Replay Buffer Size                       & $5 \times 10^6$ & $5 \times 10^6$ & $5 \times 10^6$  \\
\bottomrule
\end{tabular}
\end{table}

\begin{table}[!h]
\centering
\caption{Reward scale}
\label{tbl:reward_scale}
\begin{tabular}{cc} 
\hline
                     & Reward scale  \\ 
\hline
Ant                  & 0.01          \\
Humanoid             & 0.01          \\
ANYmal               & 1.0           \\
Franka Cube Stacking & 0.1           \\
Allegro Hand         & 0.01          \\
Shadow Hand          & 0.01          \\
Ball Balance         & 0.1           \\
DClaw Hand         & 0.01           \\
\hline
\end{tabular}
\end{table}

\subsection{Hardware Configurations}

\tblref{tbl:hardware} lists the hardware configurations of the workstations we used for the experiments. We use the machines with GeForce RTX $3090$ for experiments by default. We also measure how much time it takes for the simulator to generate $1$M interaction data with $4096$ parallel environments on \ant and \shadow. We generate $1$M data via the following command.
\begin{verbatim}
for i in range(244):
    action = torch.randn((4096, 
                          envs.action_space.shape[0]), 
                          device='cuda')
    out = envs.step(action)
\end{verbatim}

\renewcommand{\arraystretch}{1.5}
\begin{table}[!htb]
\centering
\caption{Hareware configurations on different workstations}
\label{tbl:hardware}
\resizebox{\columnwidth}{!}{
\begin{tabular}{c|ccccc} 
\hline
\multicolumn{2}{c}{}                                                                                                        & Workstation 1                & Workstation 2              & Workstation 3      & Workstation 4            \\ 
\hline
\multicolumn{2}{c}{CPU}                                                                                                     & AMD Threadripper 3990X & Intel Xeon Gold 6248 & AMD Rome 7742 & Intel Xeon W-2195  \\
\multicolumn{2}{c}{GPU}                                                                                                     & GeForce RTX 3090             & Tesla V100                 & Tesla A100         & GeForce RTX~2080 Ti      \\
\multicolumn{2}{c}{GPU CUDA Cores}                                                                                          & 10496                        & 5120                       & 6912               & 4352                     \\
\multicolumn{2}{c}{GPU FP32 TFLOPs}                                                                                         & 35.58                        & 16.4                       & 19.5               & 13.45                    \\ 
\hline
\multirow{2}{*}{\begin{tabular}[c]{@{}c@{}}Time for generating \\ 1M data ($N=4096$) (s)\end{tabular}} & Ant & $1.678\pm 0.006$  &  $2.117\pm0.038$  &   $1.999\pm0.004$   &   $3.397\pm0.014$   \\
   & Shadow Hand &   $6.706\pm0.028$  &$9.051\pm0.035$ & $8.653\pm0.101$   & $10.885\pm0.025$\\
\hline
\end{tabular}
}
\end{table}
\renewcommand{\arraystretch}{1.}

\subsection{Vision experiment setup}
\label{appsubsec:vision}

We render the RGB camera image in a resolution of $48\times48$. The CNN part of our vision network $g(o_t)$ is as follows: 
\begin{verbatim}
Conv(3,32,3,2)-BN(32)-ReLU-3x(Conv(32,32,3,2)-BN(32)-ReLU)
\end{verbatim}
where \verb|Conv(a,b,k,s)| is a Convolutional layer with input channels $a$, output channels $b$, kernel size $k$, stride $s$.

\begin{figure*}[!htbp]
    \centering
    \subfigure[]{\label{fig:ball_env}\includegraphics[width=0.247\linewidth]{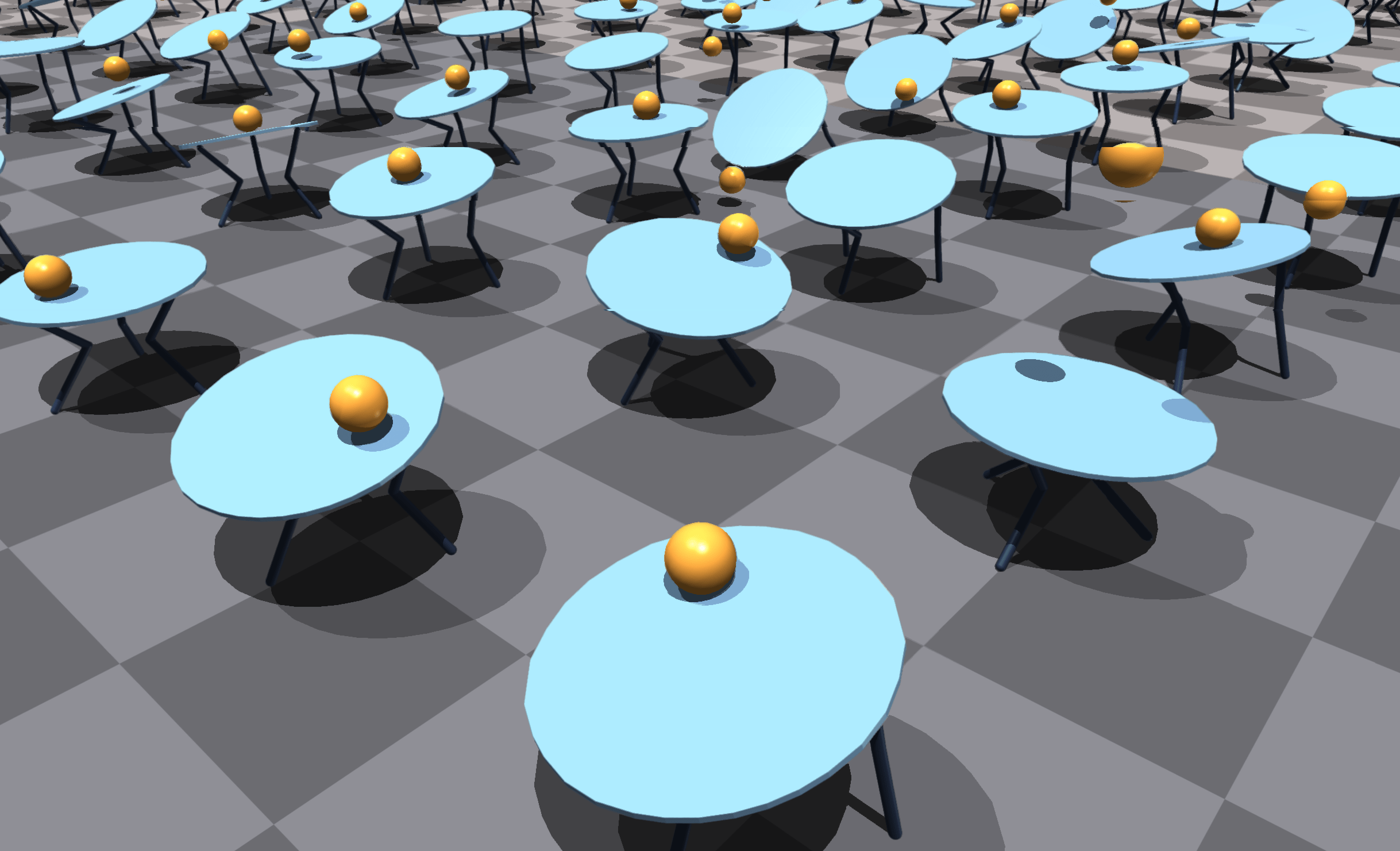}}
    \subfigure[]{\label{fig:ball_render}\includegraphics[width=0.15\linewidth]{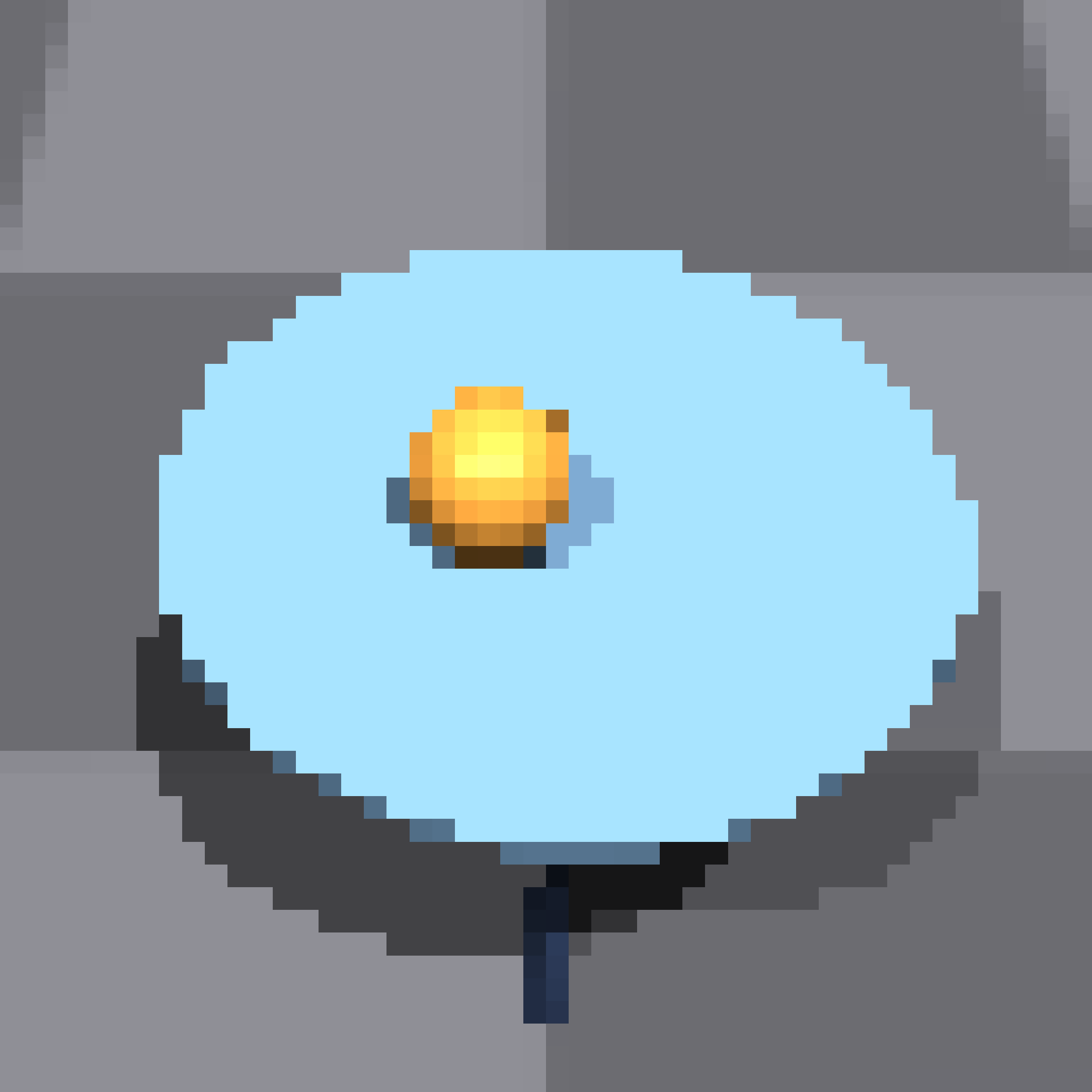}}
    \subfigure[]{\label{fig:vision_sample}\includegraphics[width=0.29\linewidth]{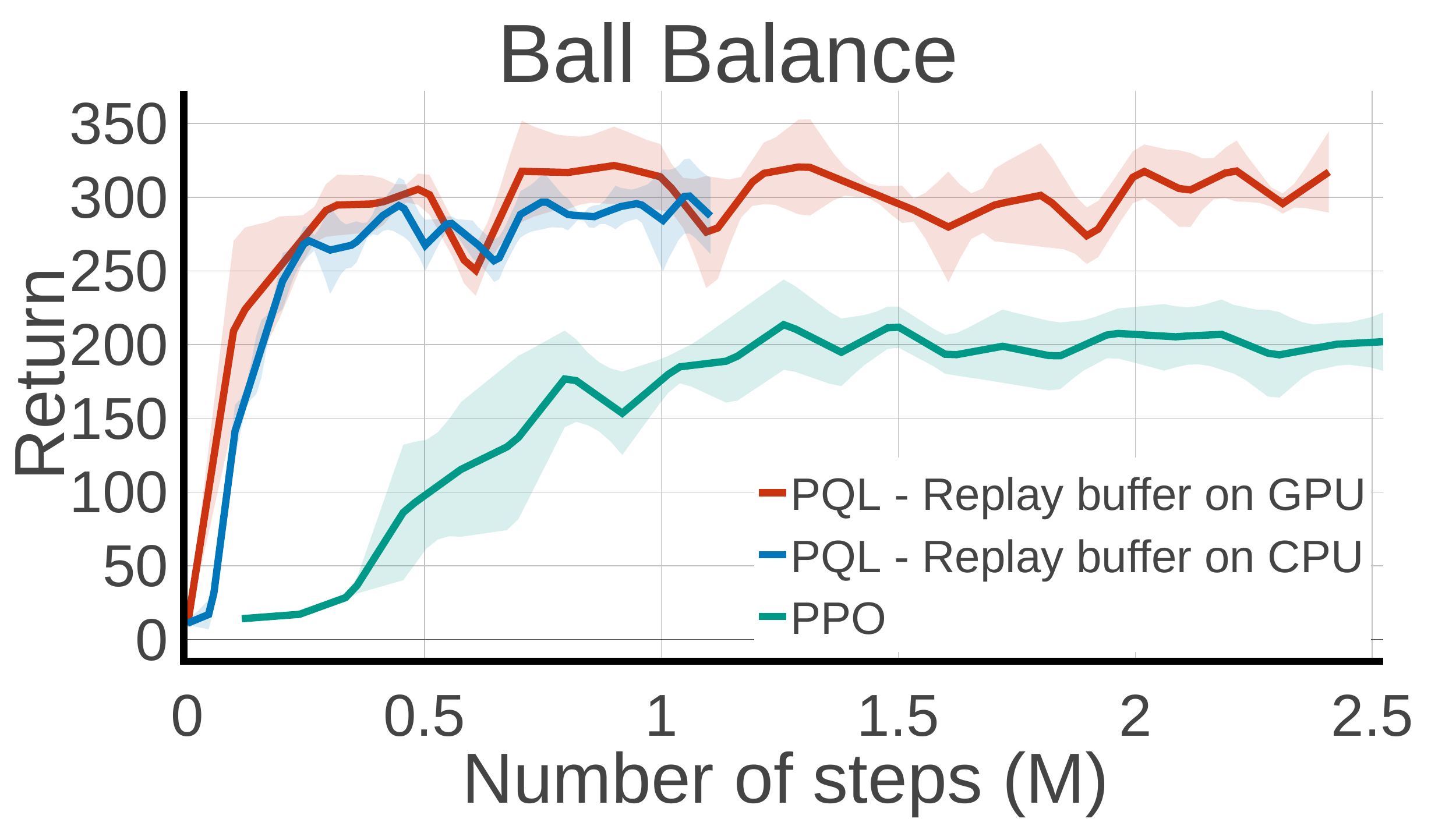}}
    \subfigure[]{\label{fig:vision_time}\includegraphics[width=0.29\linewidth]{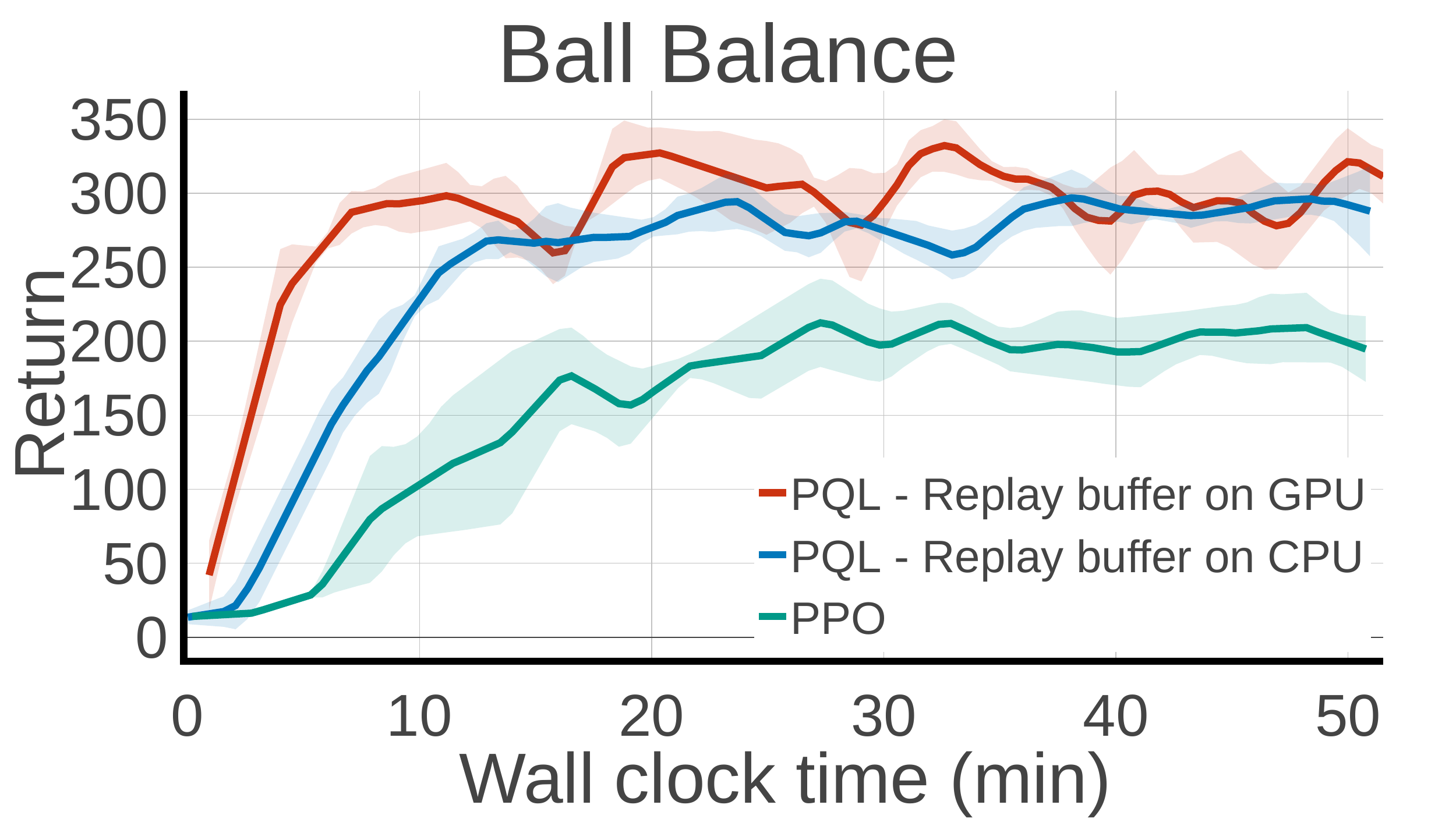}}
    \caption{\textbf{(a)}: the \ball task in Isaac Gym. \textbf{(b)}: the rendered RGB image from the simulated camera. \textbf{(c)} shows the learning curves regarding the number of environment steps. \textbf{(d)} shows the training wall-clock time. We can see that PQL achieves both better sample efficiency and higher final performance than PPO.}
    \label{fig:vision}
\end{figure*}

Since our policy input contains a history of observations $(o_{t-2}, o_{t-1}, o_t)$, we use the same CNN to extract the feature of each observation and then concatenate all the embeddings. Then, the concatenated embedding goes through an MLP network $h$:
\begin{verbatim}
    FC(256)-ReLU-FC(63)-ReLU-FC(3)
\end{verbatim}

In summary, at each time step $t$, the policy output is $h[\text{cat}(g(o_{t-2}), g(o_{t-1}), g(o_t))]$. Storing images in a replay buffer can take up a lot of memory. Therefore, we experiment with different placements of the replay buffer: (1) put the replay buffer on a GPU with a big memory, (2) put the replay buffer on CPU RAM. We use the same A100 GPUs for all these image-based experiments. \figref{fig:vision} shows that our method (PQL) works with either the replay buffer on the GPU or CPU, and it achieves much faster learning and better performance than PPO.

\begin{table}[!h]
\centering
\caption{Hyper-parameter setup for the \textit{Ball Balancing} task.}
\label{tbl:hyper-ball}
\begin{tabular}{lll} 
\toprule
Hyper-parameter                          & PQL(ours) & PPO  \\ 
\midrule
Num. Environments                        & 1,024      & 1,024   \\
Critic Learning Rate                     & $5 \times 10^{-4}$  & $5 \times 10^{-4}$ \\
Actor Learning Rate                      & $5 \times 10^{-4}$  & $5 \times 10^{-4}$ \\
Optimizer                                & Adam      & Adam  \\
Target Update Rate ($\tau$) & $5 \times 10^{-2}$  & -  \\
Batch Size                               & 4,096     & 4,096  \\
Horizon length                           & 1         & 16      \\
Num. Epochs                              & 12         & 5      \\
Discount Factor($\gamma$)                & 0.99      & 0.99  \\
Normalized Observations                  & True      & True  \\
Gradient Clipping                        & True      & True  \\
Exploration Policy                       & Mix       & -      \\
$N$-step target                          & 3         & -      \\
Warm-up Steps                            & 32        & -      \\
Replay Buffer Size                       & $10^6$ & -  \\
Clip Ratio                               & -         & 0.2 \\
GAE                                      & -         & True   \\
$\lambda$                                & -         & 0.95   \\
\bottomrule
\end{tabular}
\end{table}

\section{Additional Experiments}
\label{appsec:extra_exps}

\begin{figure}[!h]
    \centering
    \subfigure[]{\label{fig:ant_nstep}\includegraphics[width=0.49\linewidth]{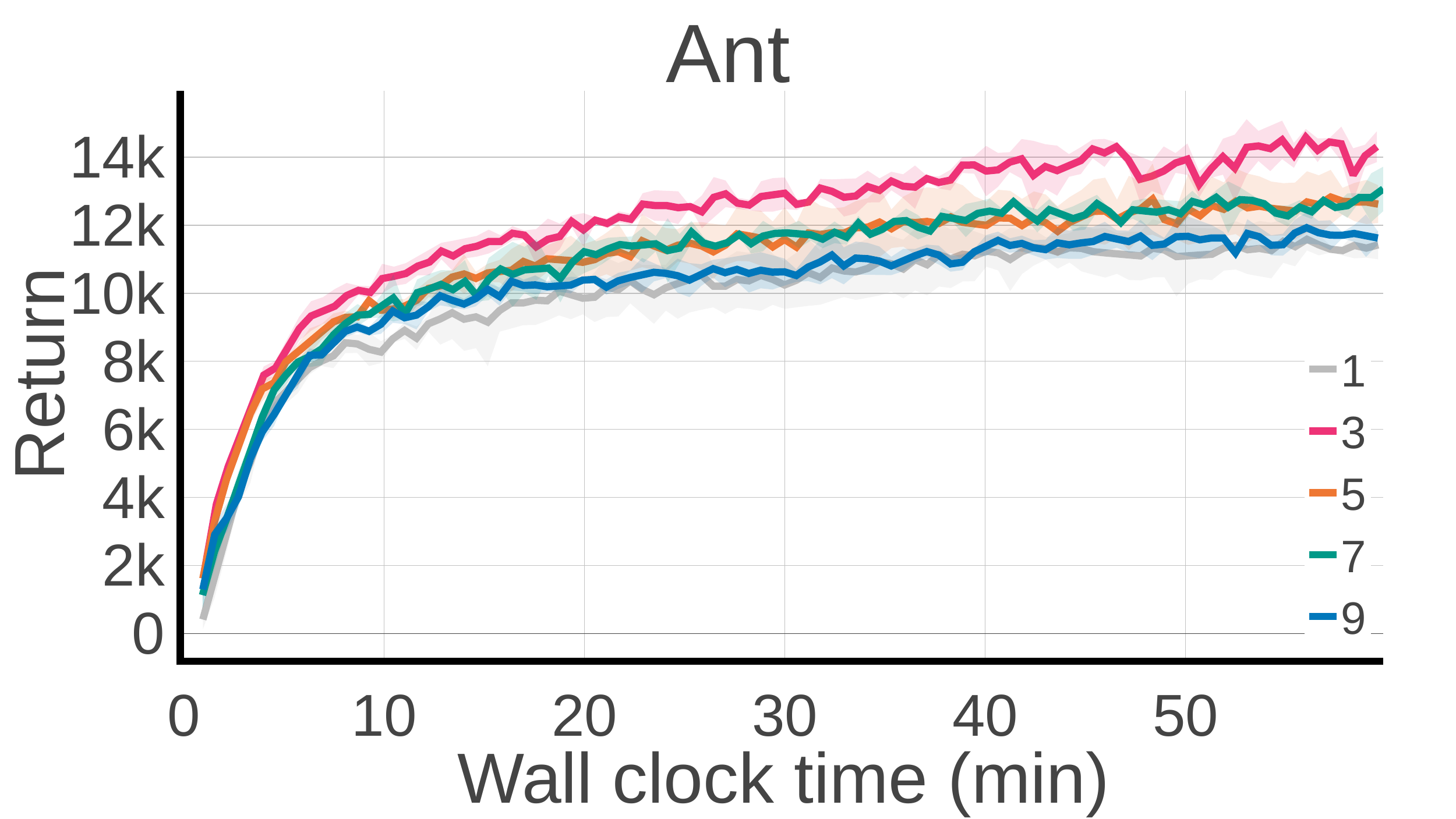}}
    \subfigure[]{\label{fig:shadow_nstep}\includegraphics[width=0.49\linewidth]{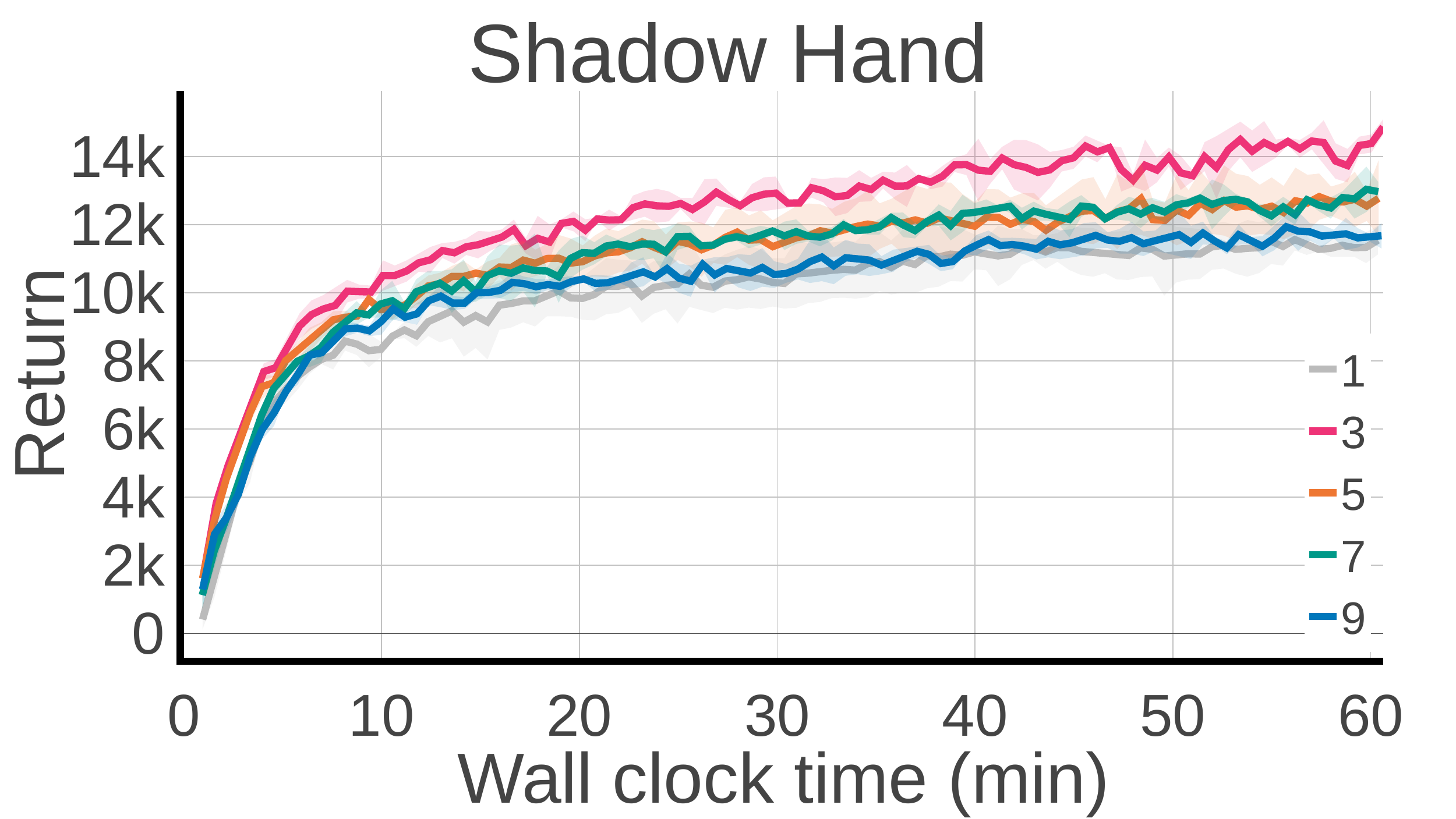}}\\
        \subfigure[]{\label{fig:ant_gpu_type}\includegraphics[width=0.49\linewidth]{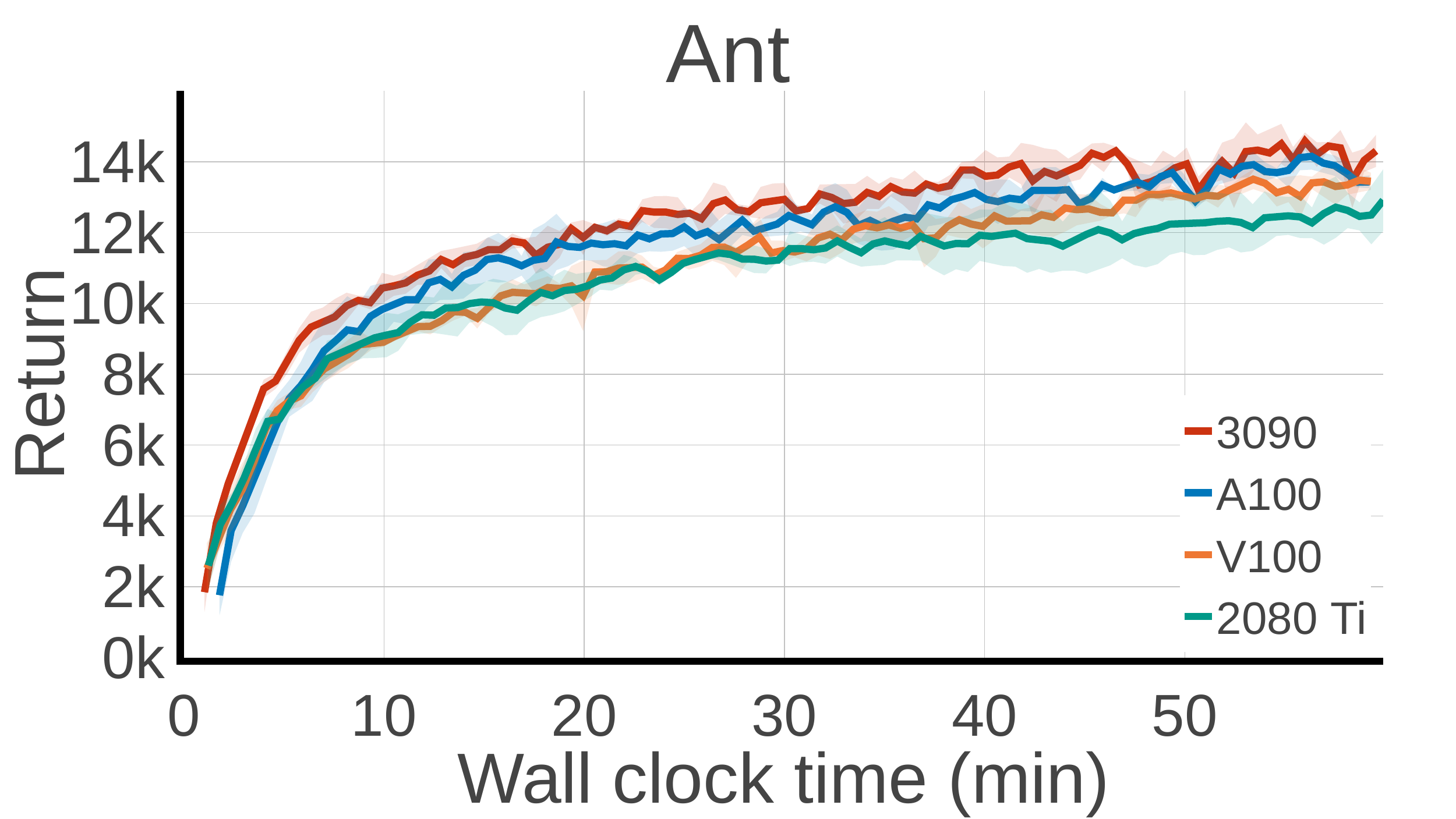}}
    \subfigure[]{\label{fig:shadow_gpu_type}\includegraphics[width=0.49\linewidth]{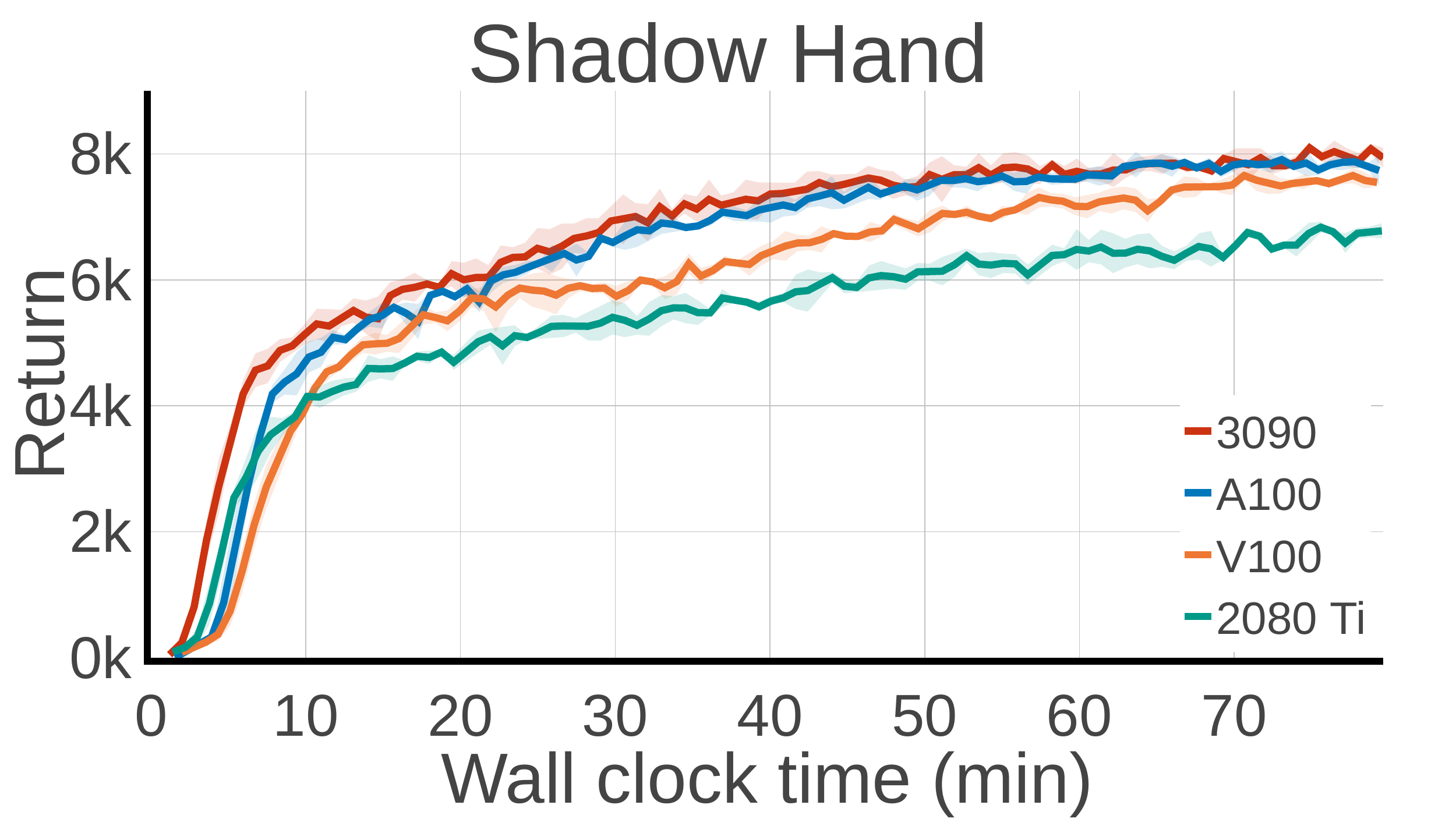}}
    \caption{\textbf{(a)} and \textbf{(b)}: effect of $n$-step return. $n=3$ performs the best. \textbf{(c)} and \textbf{(d)}: effect of GPU models used for running PQL. Overall, we see that PQL works robustly across different GPU models, and running on newer GPUs tends to give faster learning. }
    \label{fig:nstep_replay}
\end{figure}

\paragraph{$n$-step returns} We investigate how much does $n$-step returns help for PQL. As shown in \figref{fig:nstep_replay}, adding $n$-step return leads to faster learning than not using $n$-step return ($n=1$). However, using a big $n$ value hurt the learning. Empirically we found that $n=3$ gives us the best performance.

\begin{figure}[!h]
    \centering
    \subfigure[]{\label{fig:ant_sac_baseline_time}\includegraphics[width=0.33\linewidth]{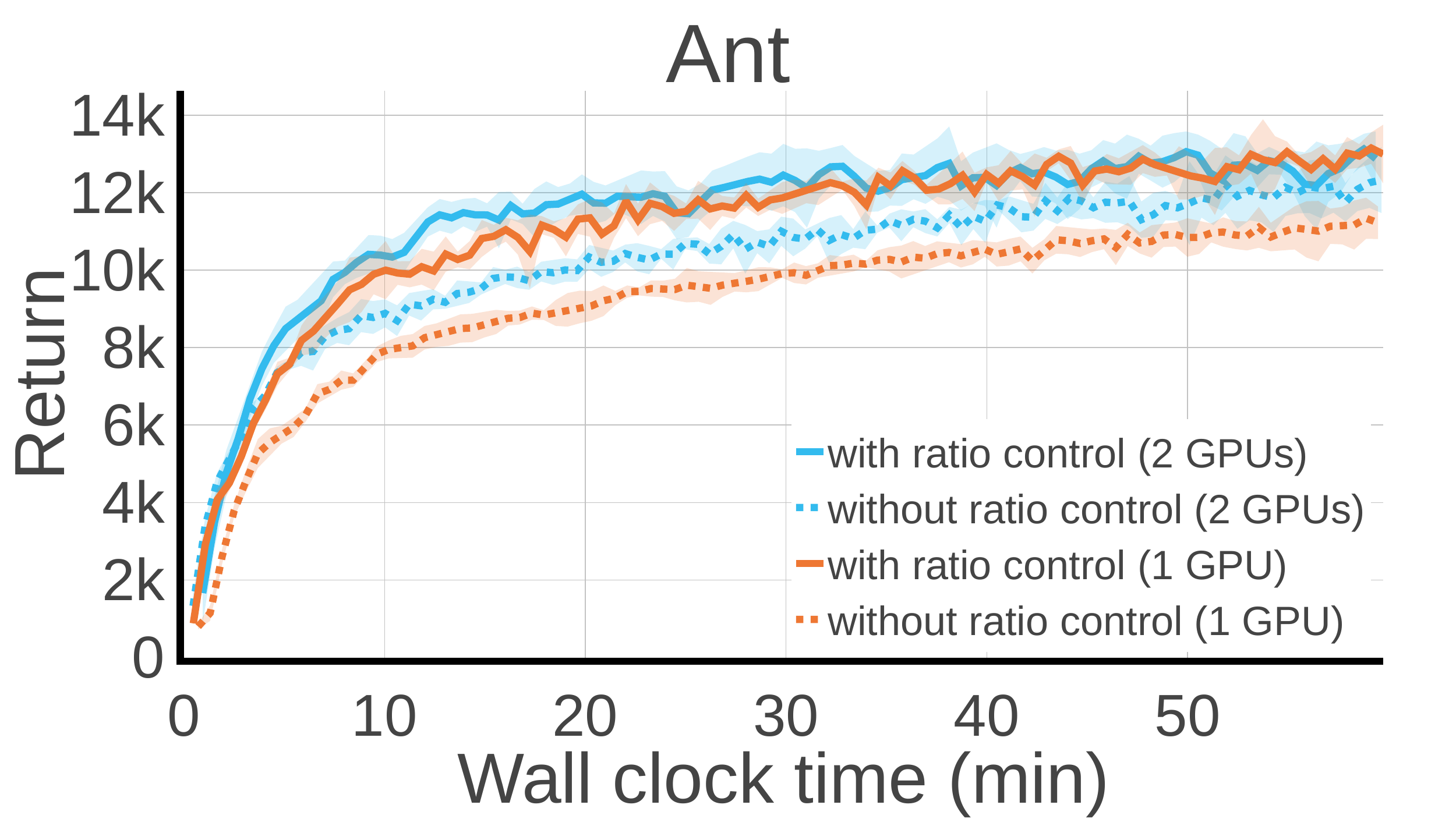}}
    \subfigure[]{\label{fig:humanoid_sac_baseline_time}\includegraphics[width=0.33\linewidth]{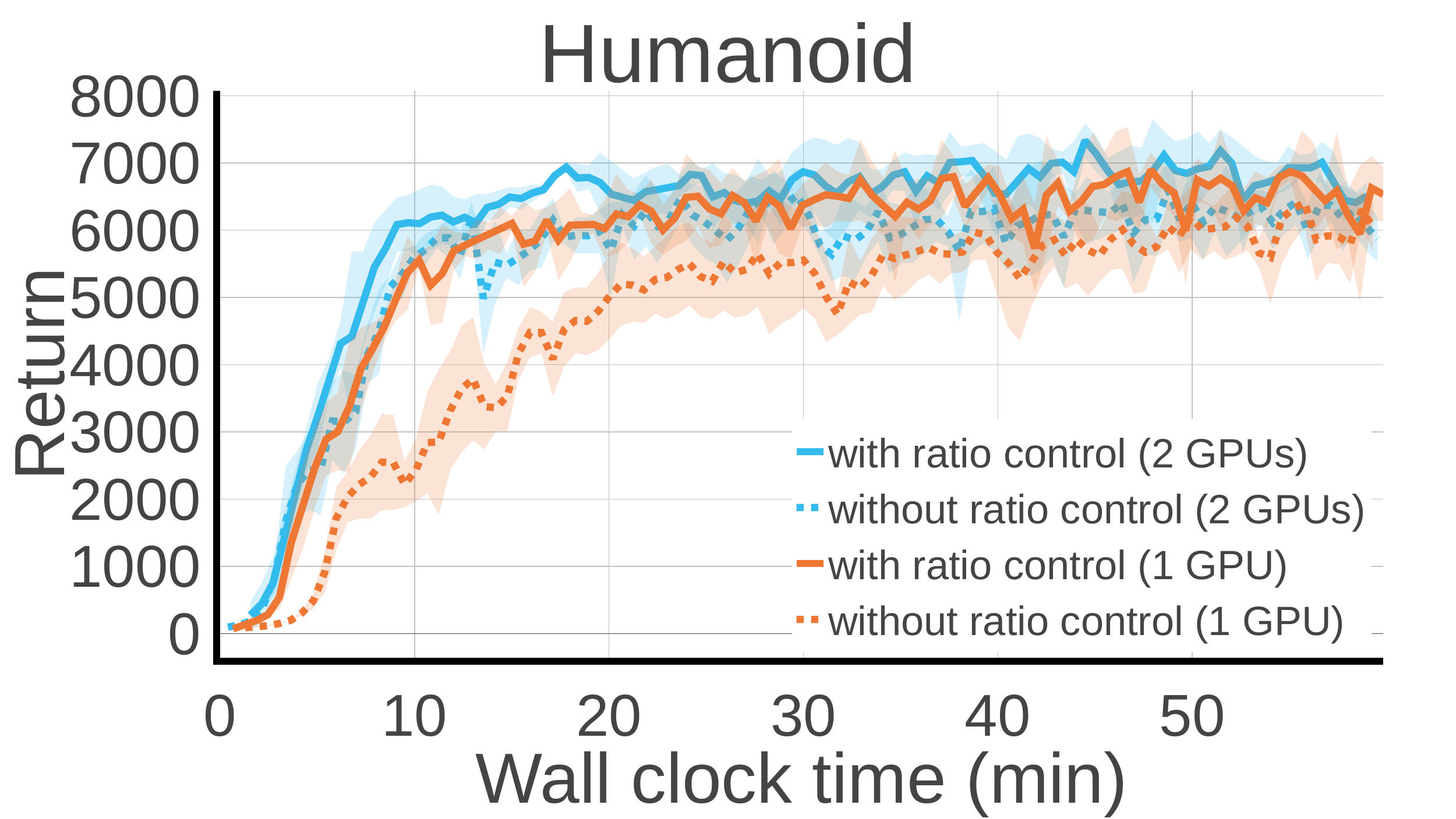}}
    \subfigure[]{\label{fig:anymal_sac_baseline_time}\includegraphics[width=0.33\linewidth]{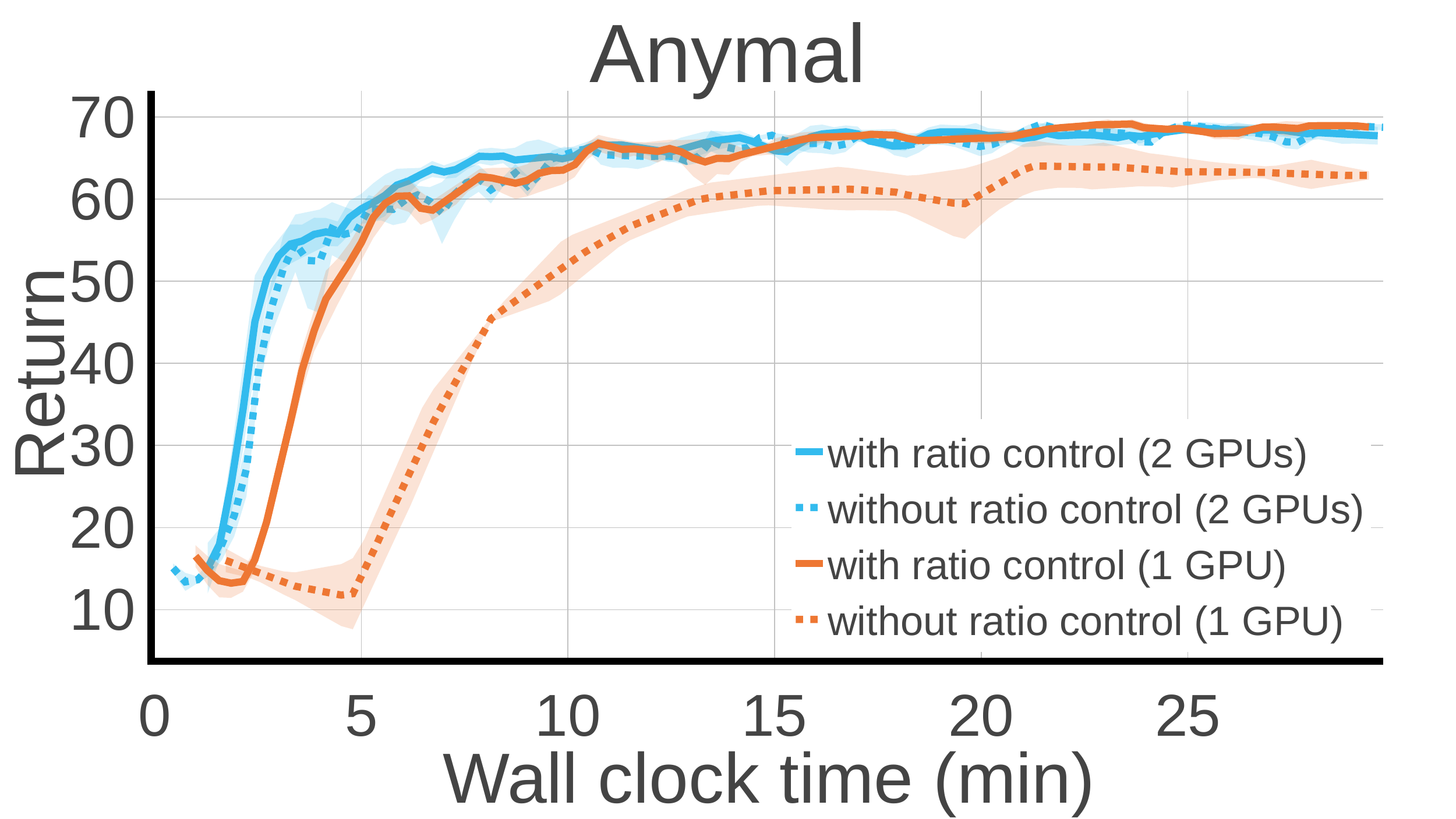}}
    \subfigure[]{\label{fig:franka_sac_baseline_time}\includegraphics[width=0.33\linewidth]{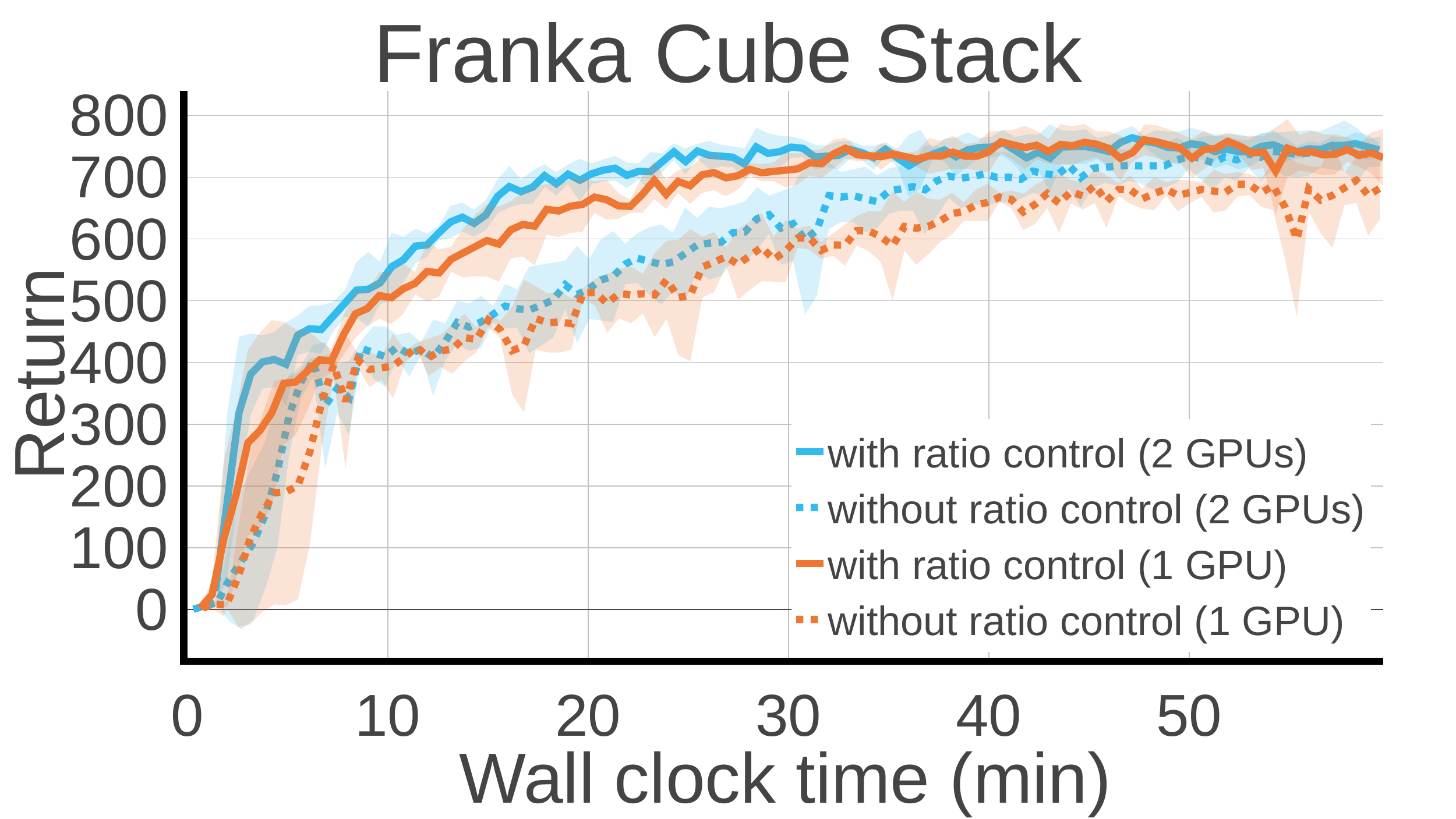}}
    \subfigure[]{\label{fig:allegro_sac_baseline_time}\includegraphics[width=0.33\linewidth]{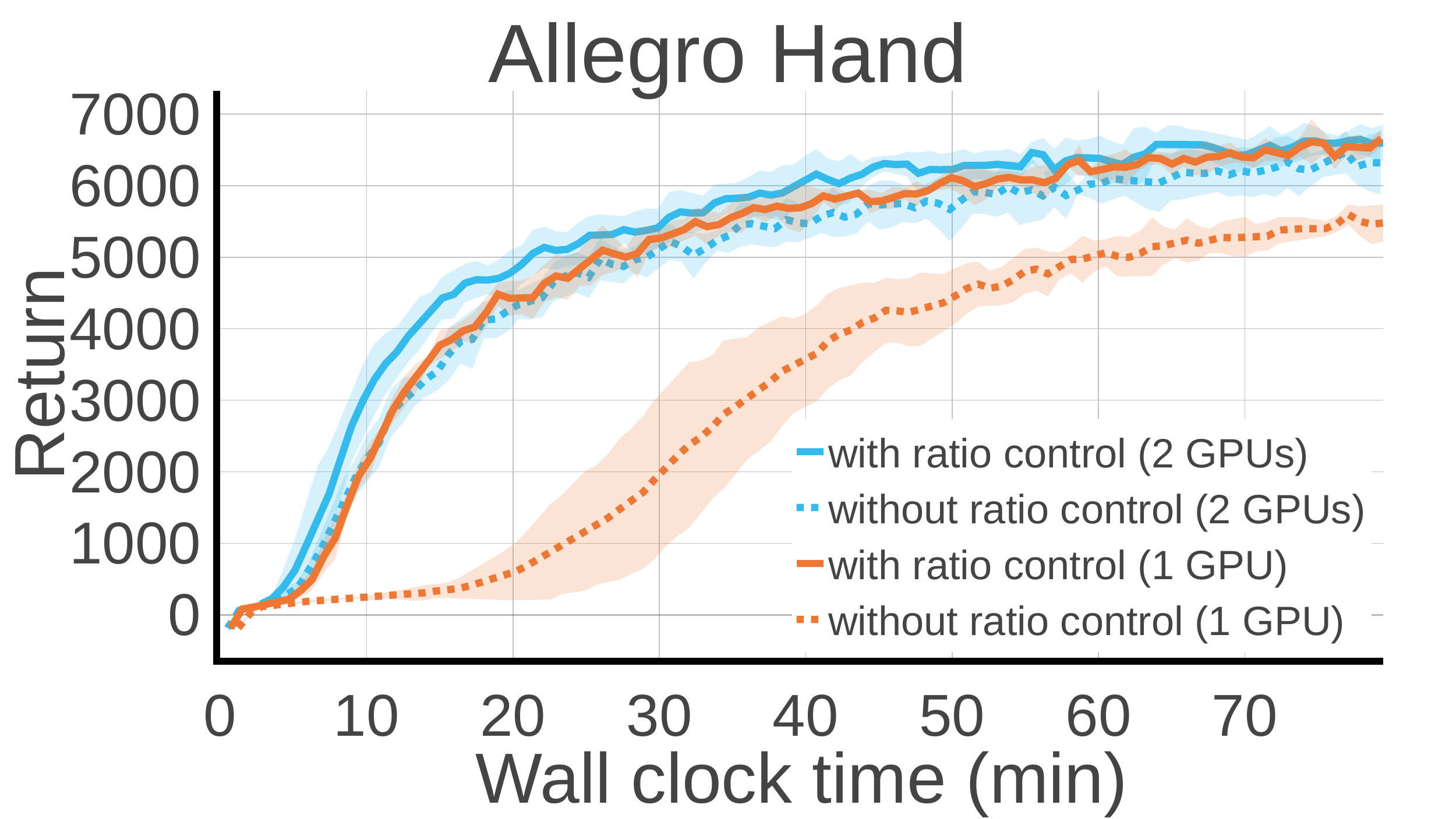}}
    \subfigure[]{\label{fig:shadow_sac_baseline_time}\includegraphics[width=0.33\linewidth]{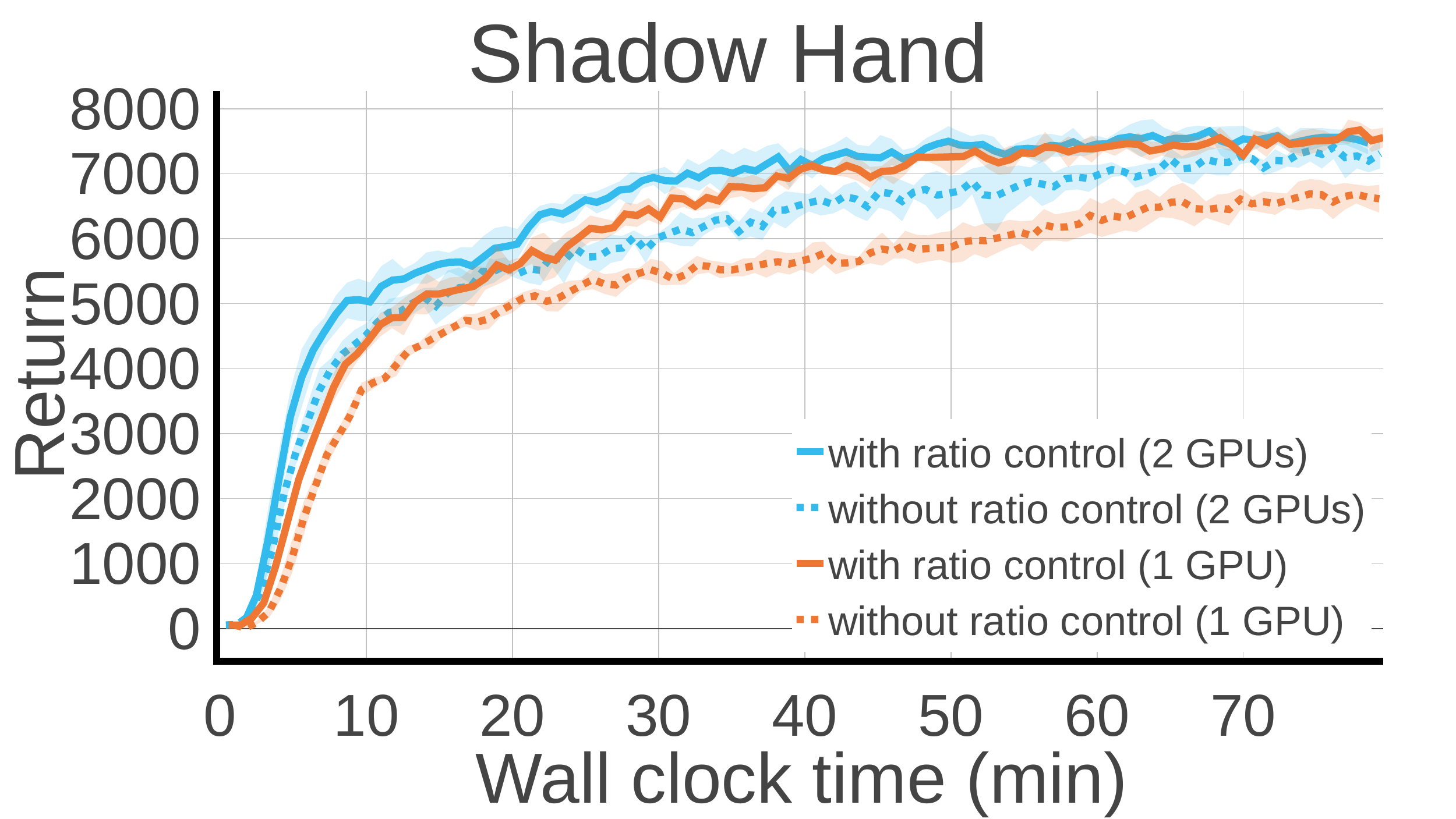}}
    \caption{Comparison of the learning performance with and without the speed ratio control ($\beta_{p:v}, \beta_{a:v}$) on two RTX3090 GPUs and on 1 RTX3090 GPU, respectively.}
    \label{fig:beta_benefit}
\end{figure}

\paragraph{Benefit of adding speed control ($\beta_{p:v}, \beta_{a:v})$ on different processes} As we mentioned in \secref{subsec:ratio_balance}, adding speed control using $\beta_{p:v}, \beta{a:v}$ can help reduce the variance of training when the amount of computation resources changes. To provide more insights, we ran experiments without speed control, i.e., each process could run as fast as possible without any waiting. As shown in \figref{fig:beta_benefit}, when there are sufficient compute resources available (with two GPUs), the benefit of having the speed ratio control is not significant. However, when only one GPU is available for running all three processes (\actor, \plearner, \vlearner), we can see that without the ratio control, the learning curves on all six benchmark tasks slow down. We believe this is because all three processes are trying to run as fast as possible, resulting in competition for GPU utilization, which slows overall learning. Adding the ratio control helps balance GPU resource utilization among the three processes. Thus, even with one GPU, the learning performance with ratio control is quite similar to that with two GPUs.

\paragraph{GPU hardware}
The simulation speed and network training speed vary across different GPU models. In \tblref{tbl:hardware}, we list how much time it takes for the simulator to generate $1$M environment interaction data with $4096$ parallel environments on four machines with different GPU models. In our test, the simulation speed on different GPU models is as follows: GeForce 3090 $>$ Tesla A100 $>$ Tesla V100 $>$ GeForce 2080Ti. We test PQL performance on all these four different machine configurations (\tblref{tbl:hardware}). \figref{fig:ant_gpu_type} and \figref{fig:shadow_gpu_type} show that different GPU models affect the policy learning speed, especially on complex tasks like \shadow which takes more simulation time.

\begin{figure}[!h]
    \centering
    \subfigure[]{\label{fig:ant_sac_baseline_time}\includegraphics[width=0.33\linewidth]{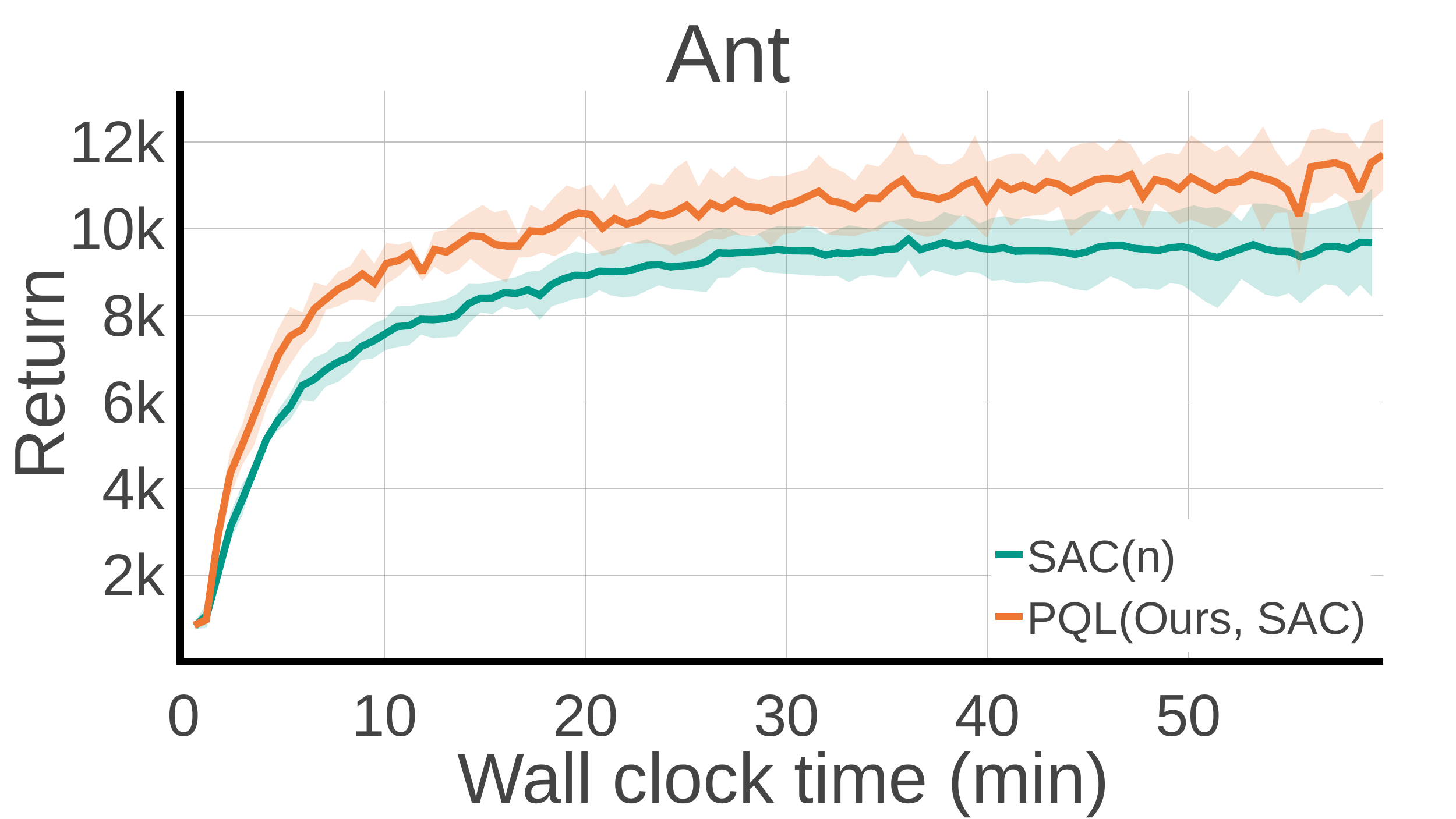}}
    \subfigure[]{\label{fig:humanoid_sac_baseline_time}\includegraphics[width=0.33\linewidth]{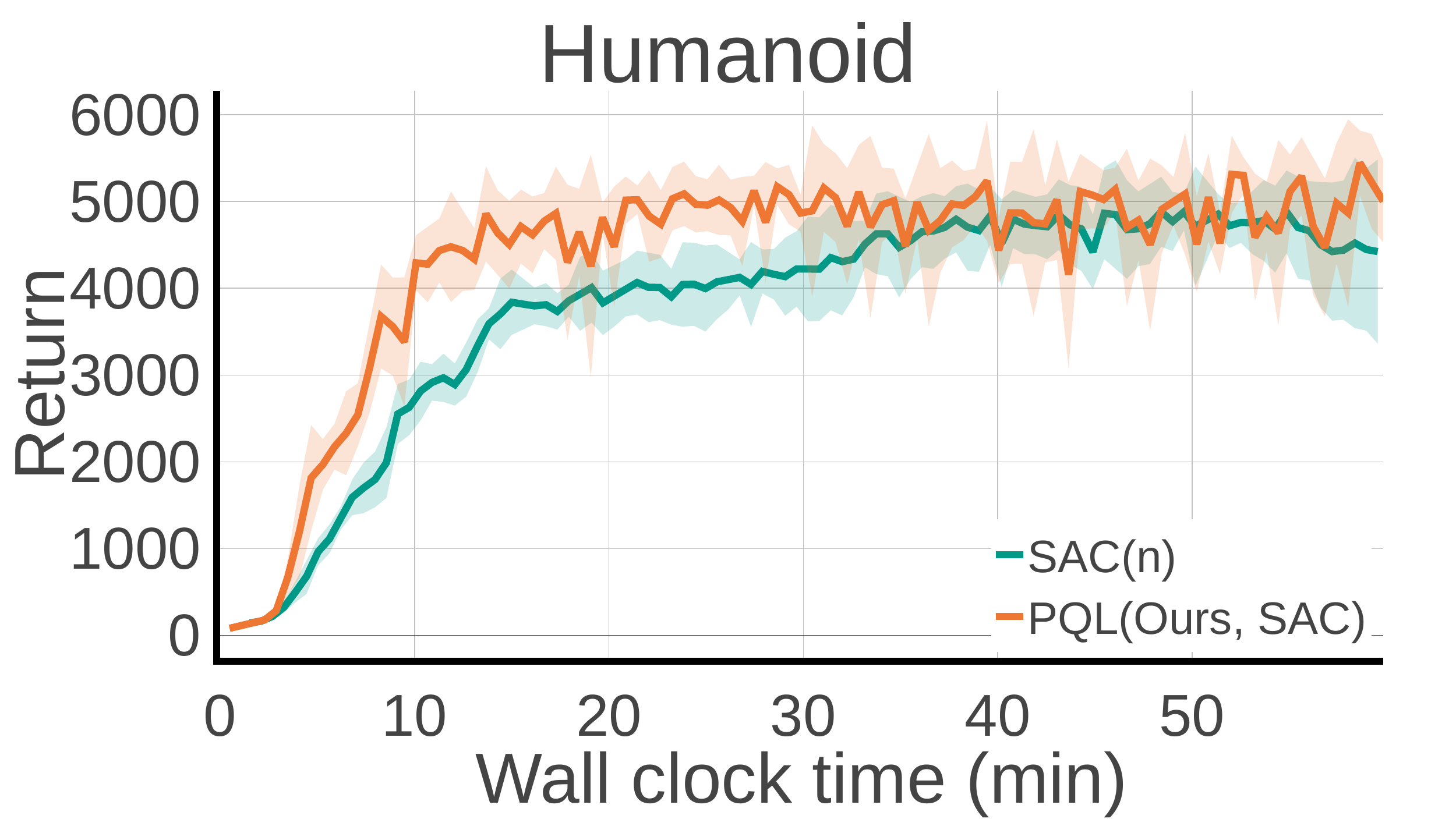}}
    \subfigure[]{\label{fig:anymal_sac_baseline_time}\includegraphics[width=0.33\linewidth]{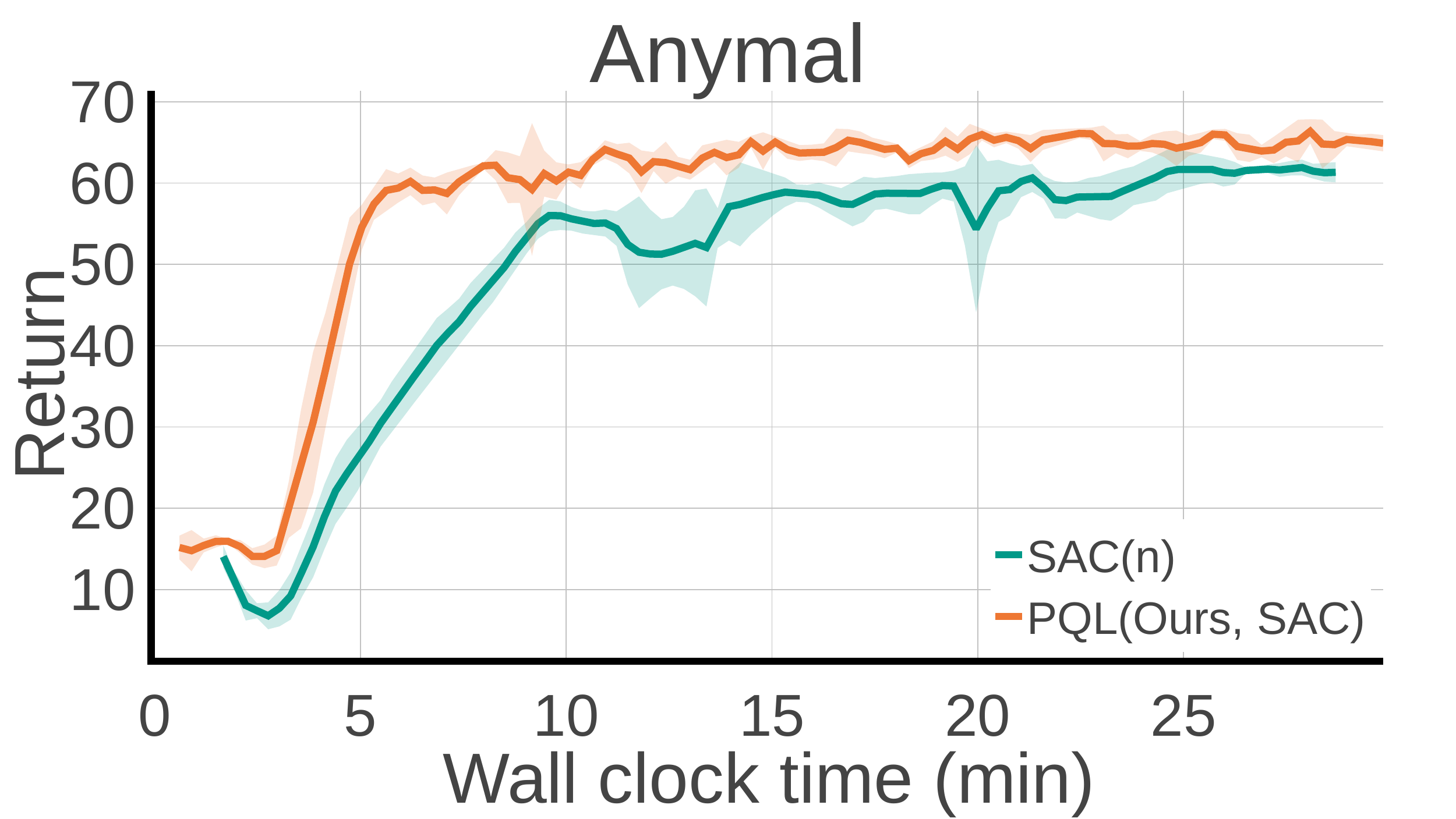}}
    \subfigure[]{\label{fig:franka_sac_baseline_time}\includegraphics[width=0.33\linewidth]{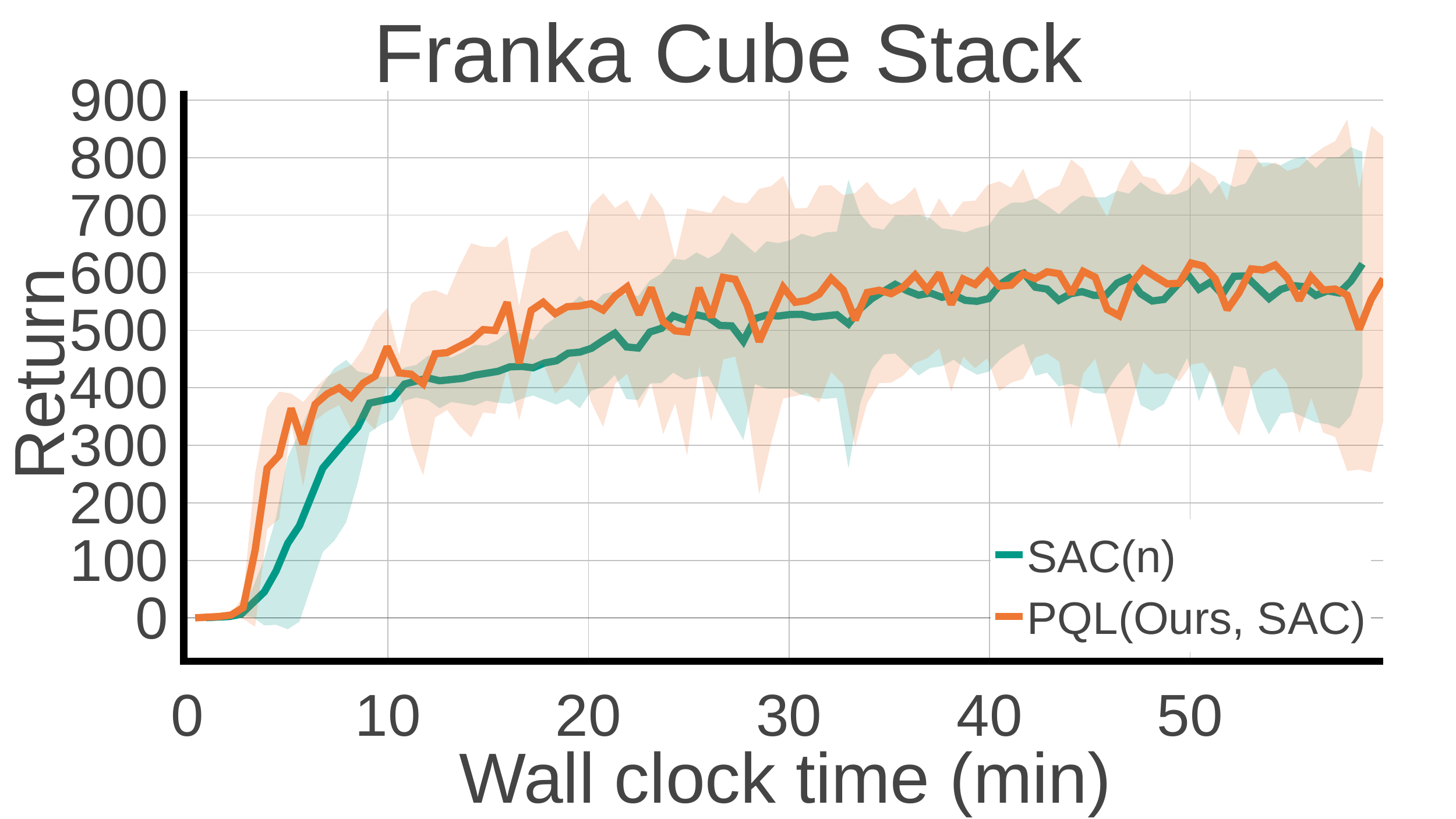}}
    \subfigure[]{\label{fig:allegro_sac_baseline_time}\includegraphics[width=0.33\linewidth]{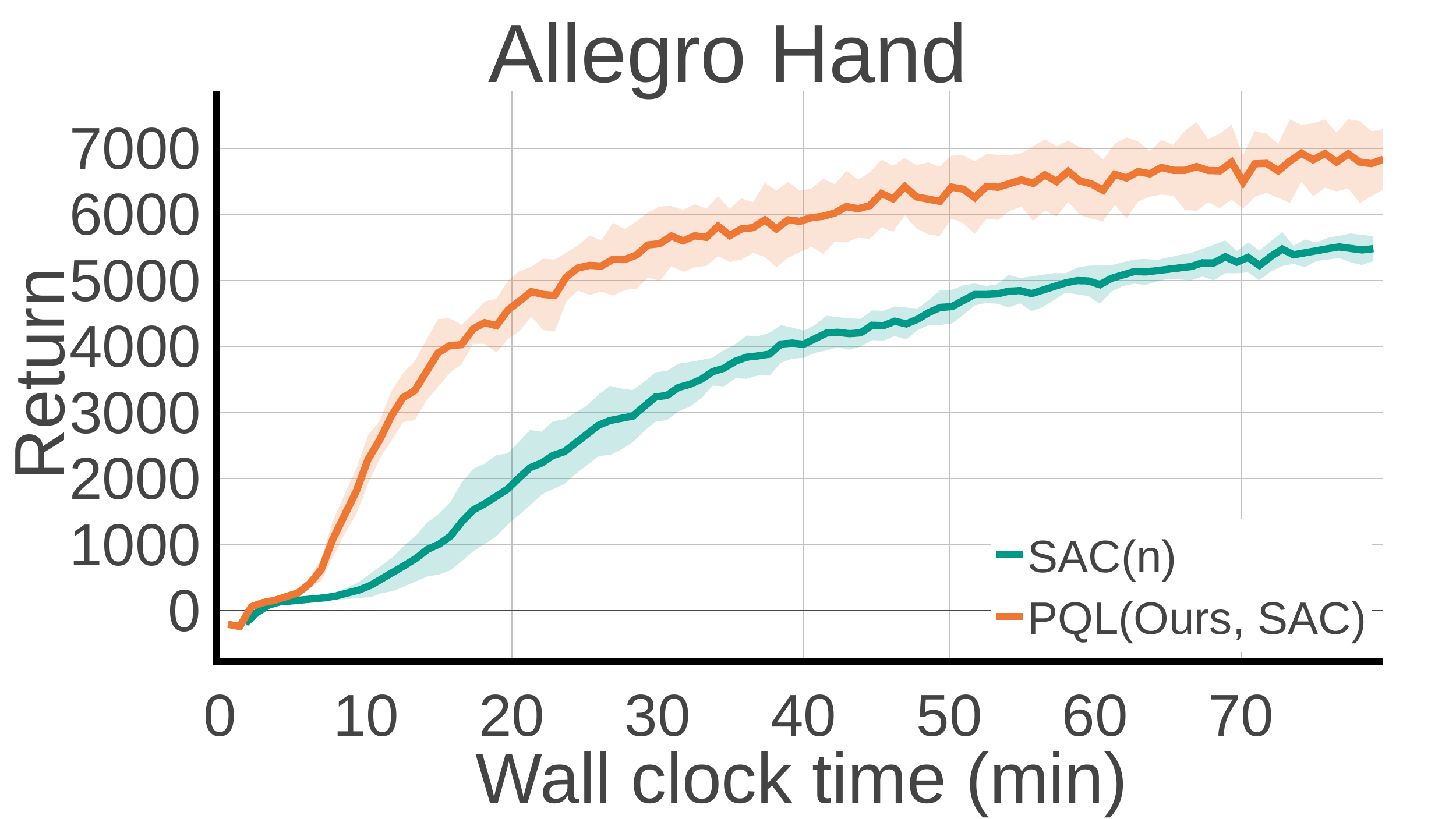}}
    \subfigure[]{\label{fig:shadow_sac_baseline_time}\includegraphics[width=0.33\linewidth]{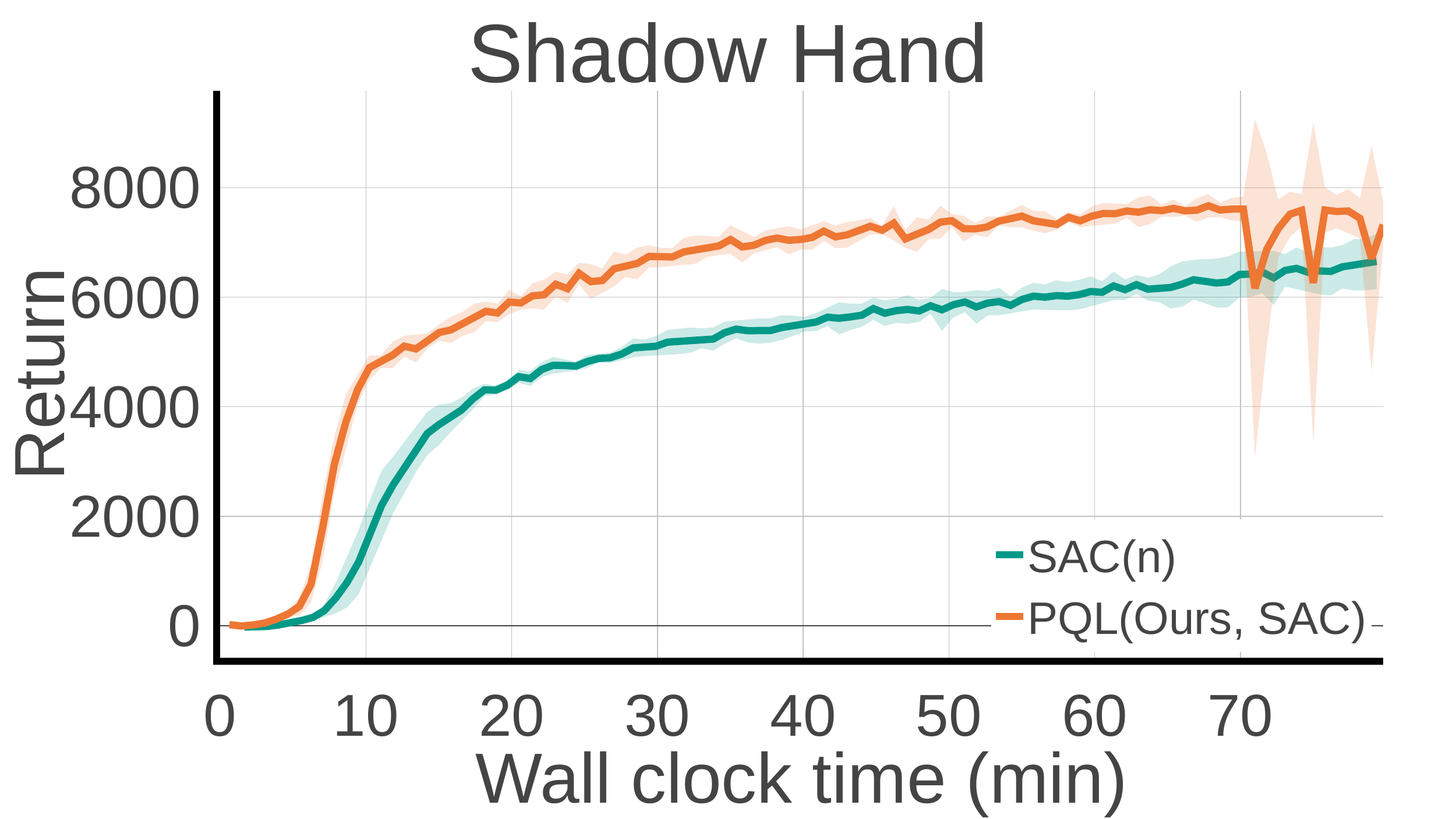}}
    \caption{We apply our parallel $Q$-learning to SAC. PQL + SAC achieves faster learning than SAC itself.}
    \label{fig:sac_baselines_time}
\end{figure}

\paragraph{PQL for SAC} As discussed above, PQL framework is flexible and can be combined with different $Q$-learning methods. Here, we show that PQL can be combined with SAC as well. \figref{fig:sac_baselines_time} shows that adding the PQL framework to SAC substantially speeds up the learning speed of SAC.

\begin{figure}[!h]
    \centering
    \subfigure[]{\label{fig:ant_baseline_sample}\includegraphics[width=0.33\linewidth]{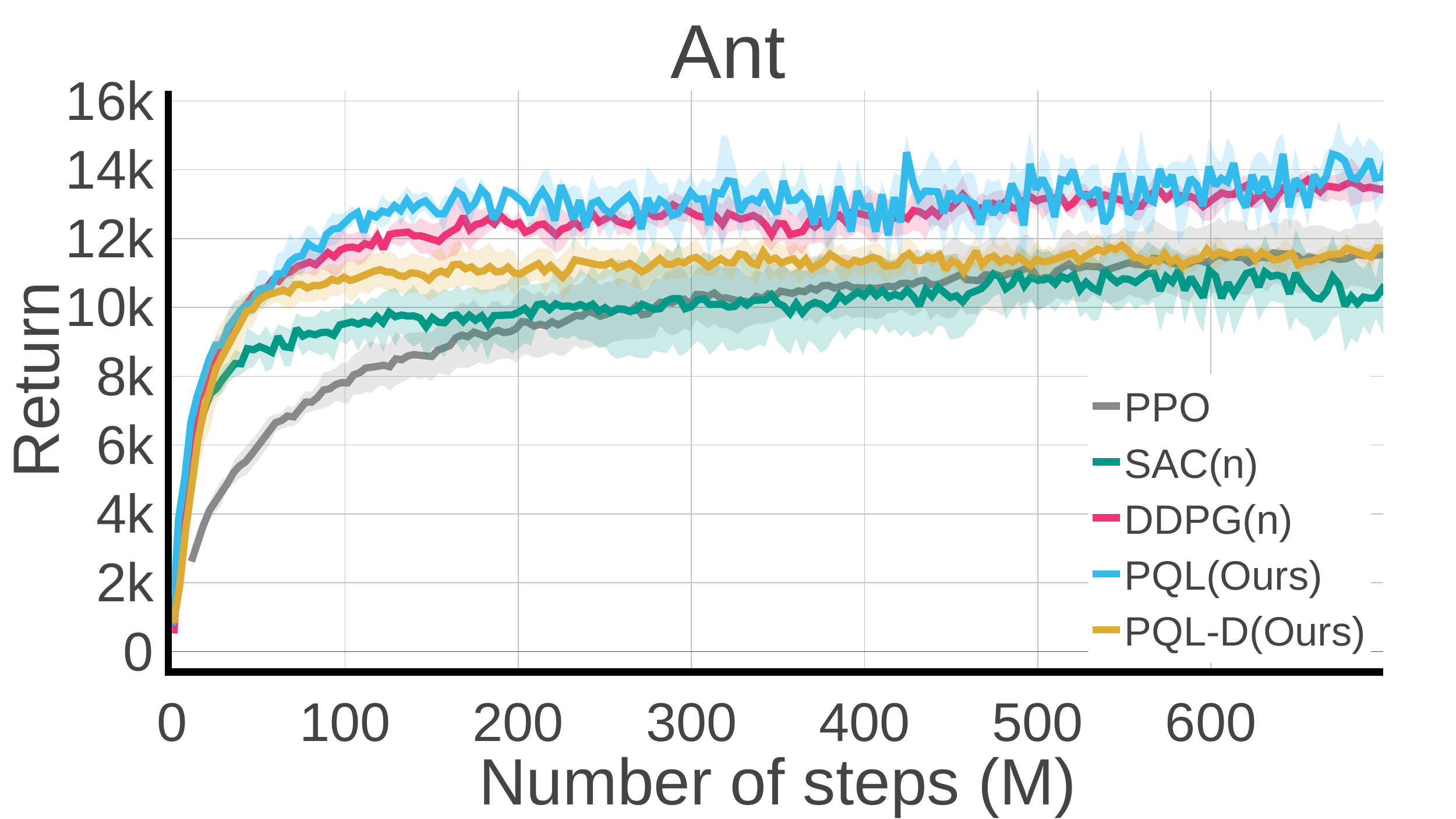}}
    \subfigure[]{\label{fig:humanoid_baseline_sample}\includegraphics[width=0.33\linewidth]{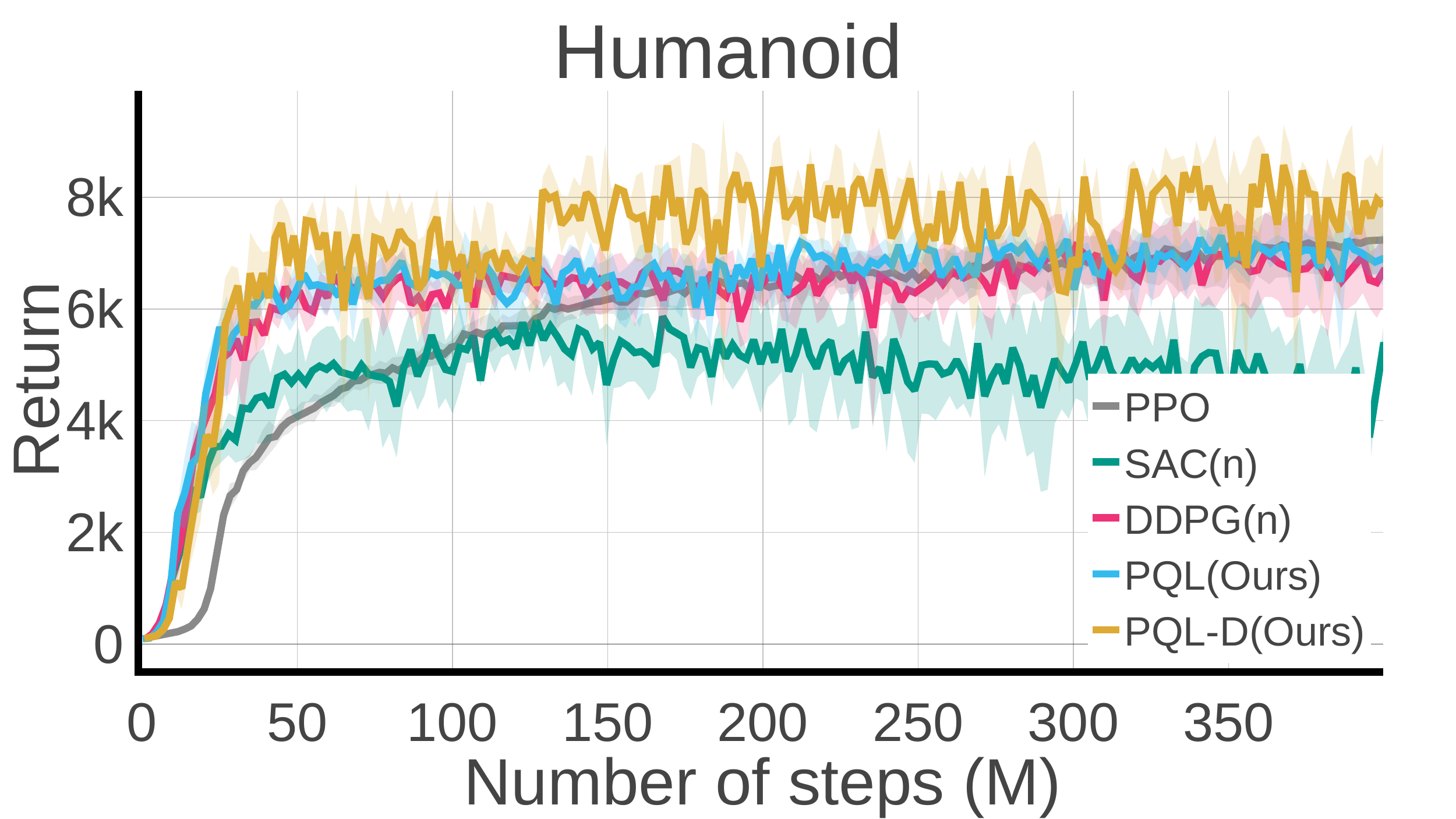}}
    \subfigure[]{\label{fig:anymal_baseline_sample}\includegraphics[width=0.33\linewidth]{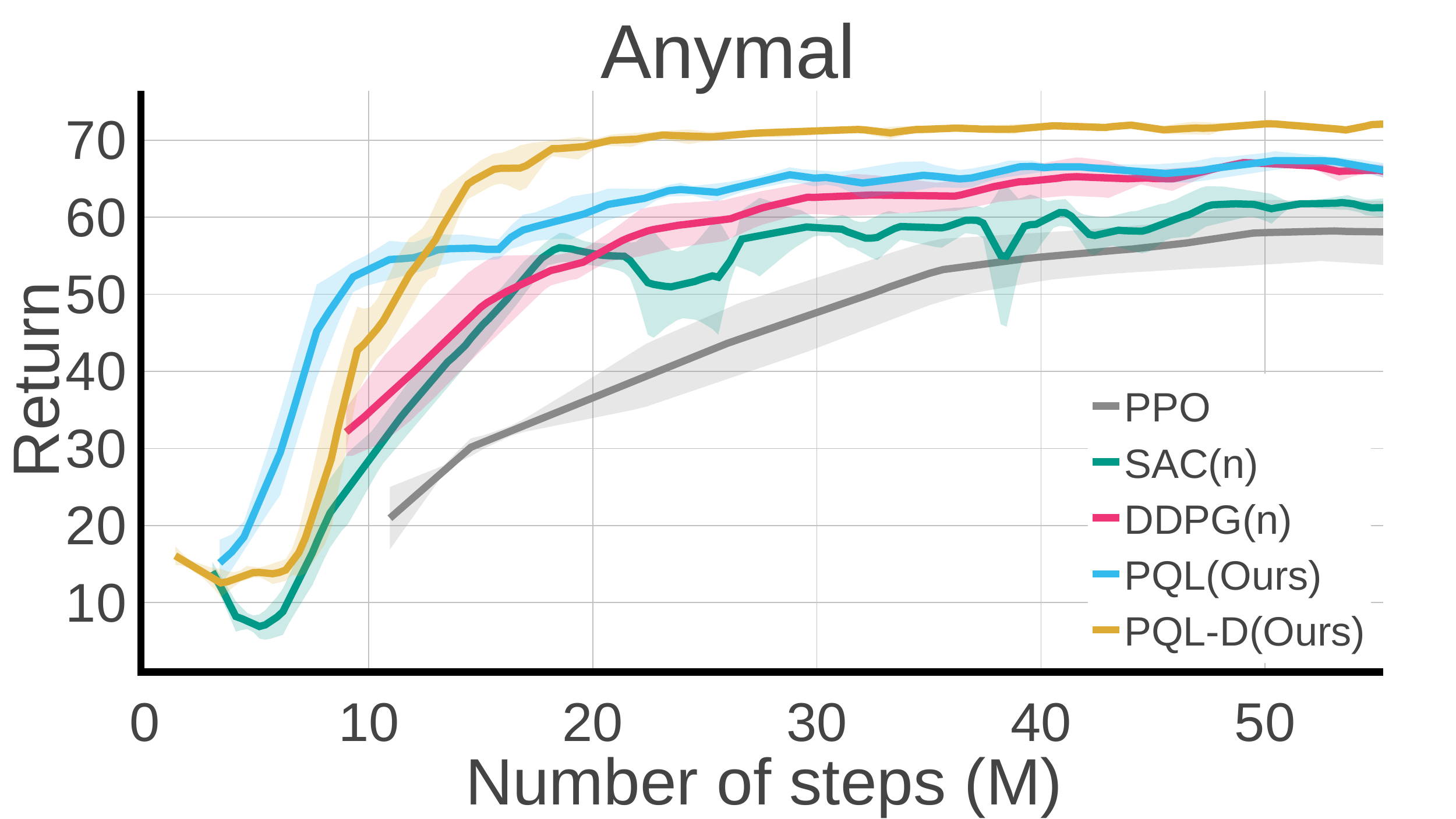}}
    \subfigure[]{\label{fig:franka_baseline_sample}\includegraphics[width=0.33\linewidth]{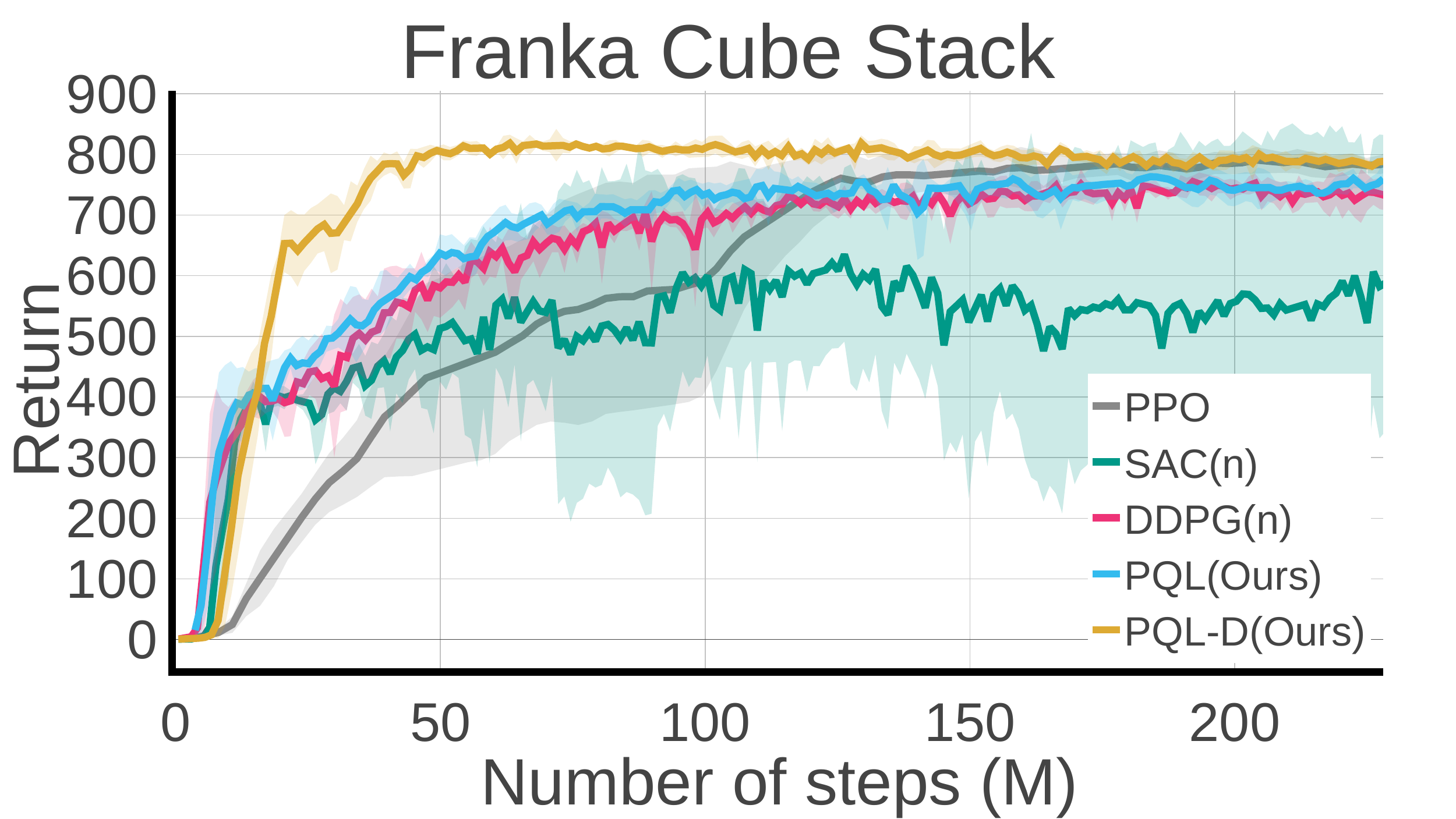}}
    \subfigure[]{\label{fig:allegro_baseline_sample}\includegraphics[width=0.33\linewidth]{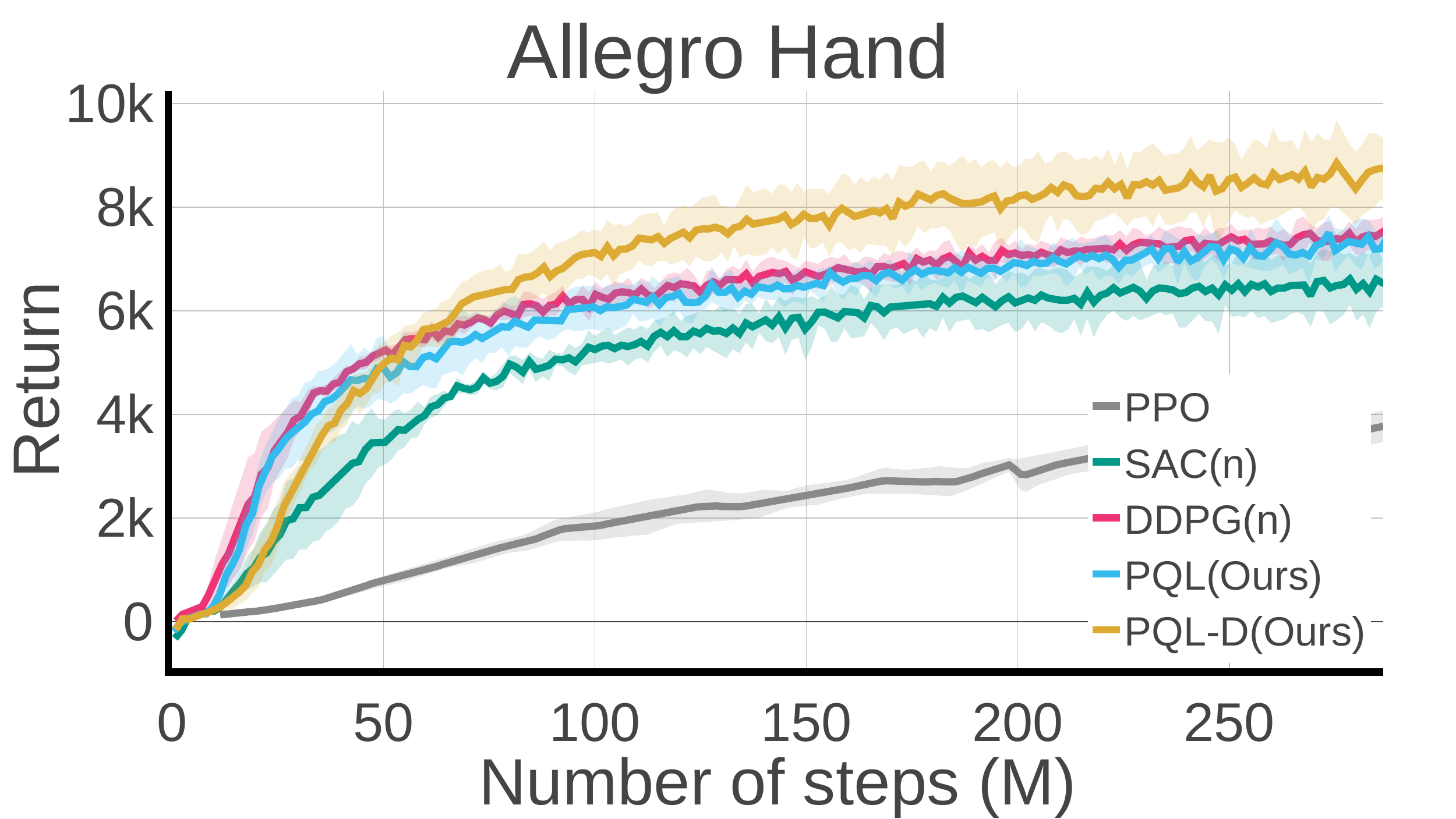}}
    \subfigure[]{\label{fig:shadow_baseline_sample}\includegraphics[width=0.33\linewidth]{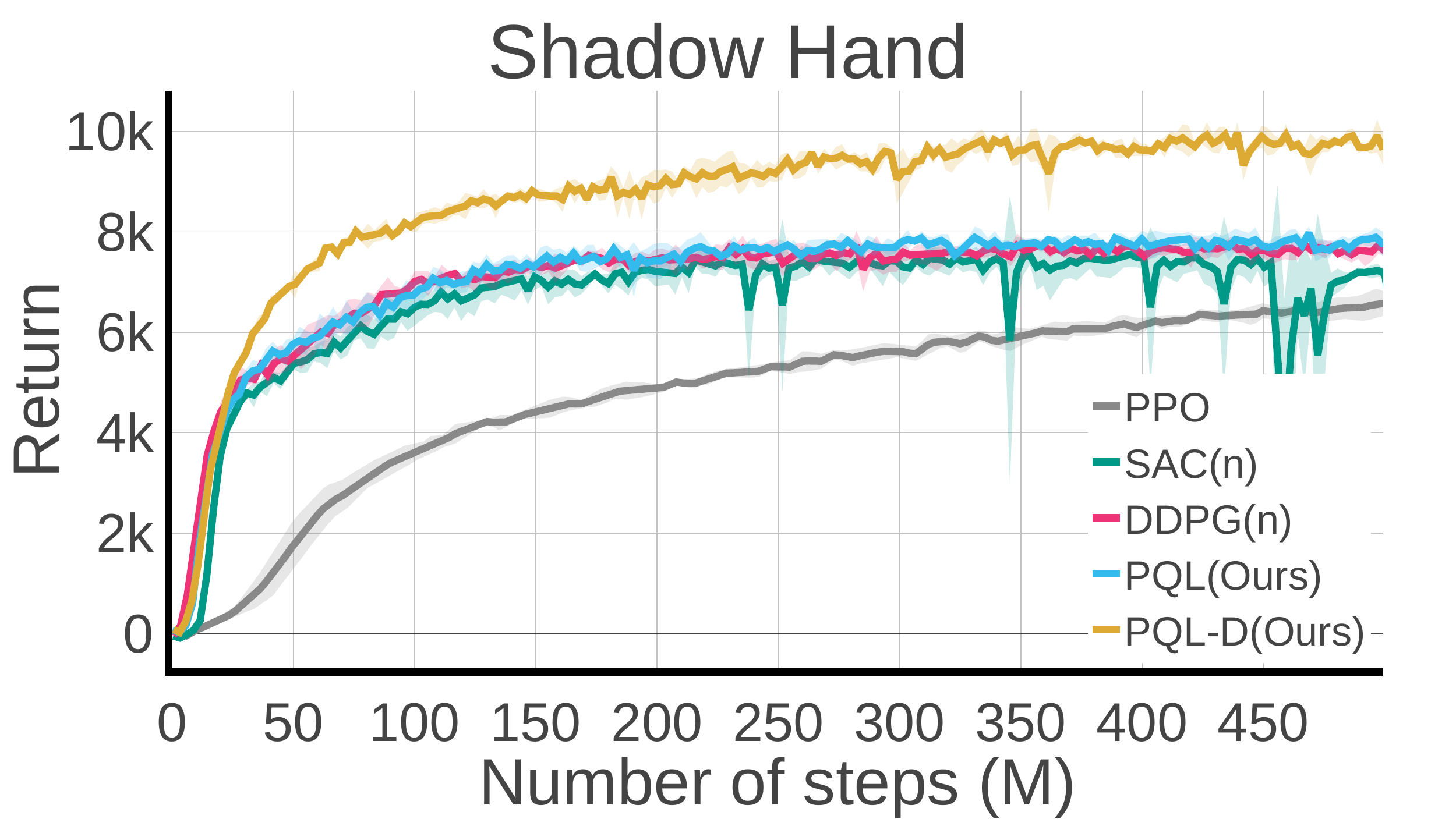}}
    \caption{Similar to \figref{fig:baselines}, we show that our method (PQL) also achieves better sample efficiency than baselines.}
    \label{fig:baselines_sample}
\end{figure}

\paragraph{Sample efficiency compared to baselines}
\figref{fig:baselines_sample} shows the sample efficiency of each algorithm on different environments. Overall, we see that PQL achieves the best sample efficiency. In addition, DDPG(n) also outperforms SAC(n) in terms of sample efficiency on these tasks.

\paragraph{Sweep over different $\beta_{a:v}$ and $\beta_{p:v}$}

\begin{figure}[!tb]
    \centering
    \subfigure[]{\label{fig:ant_2048_ac}\includegraphics[width=0.49\linewidth]{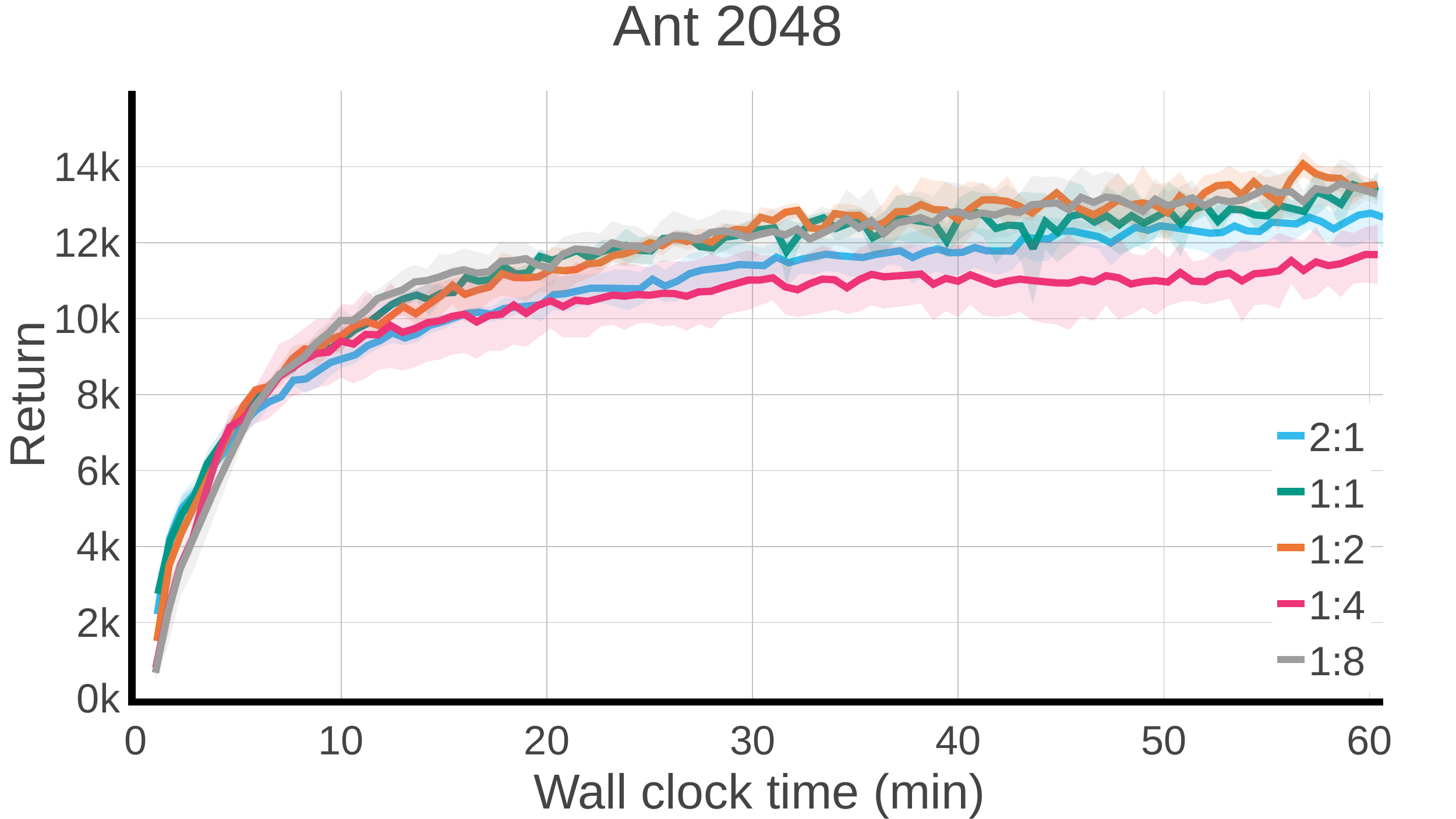}}
    \subfigure[]{\label{fig:ant_4096_ac}\includegraphics[width=0.49\linewidth]{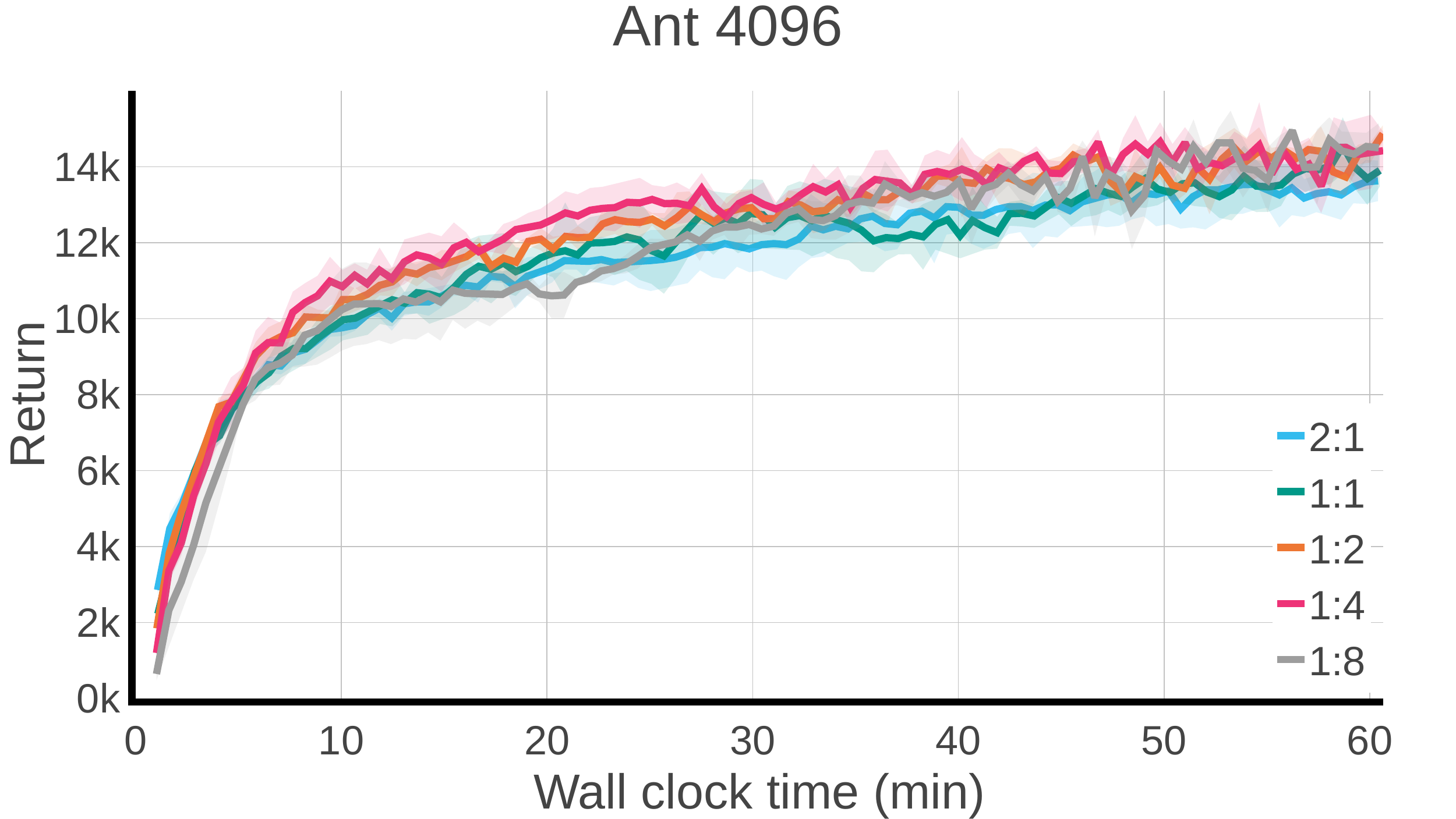}}
     \\
    \subfigure[]{\label{fig:ant_8192_ac}\includegraphics[width=0.49\linewidth]{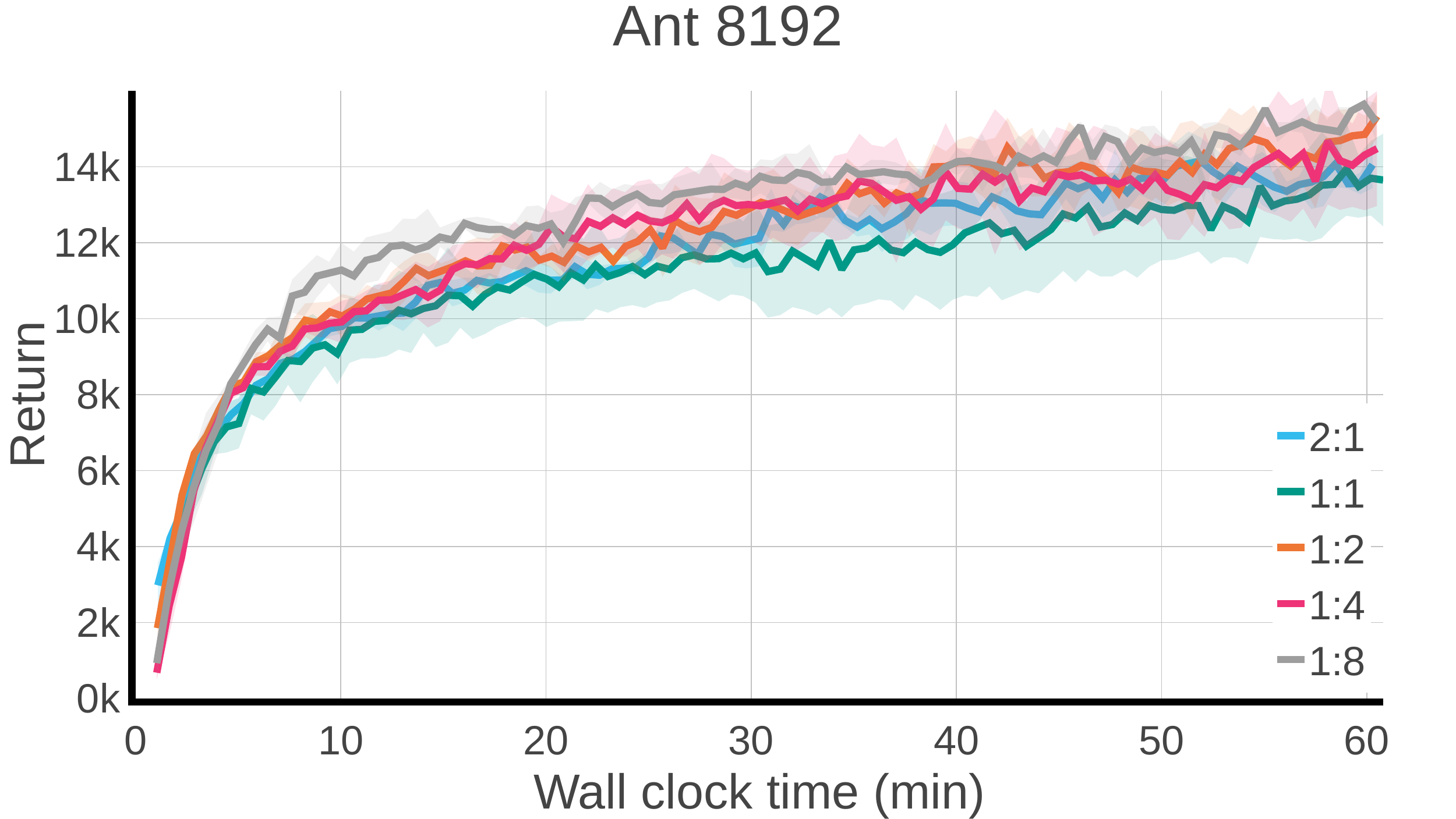}}
    \subfigure[]{\label{fig:ant_16384_ac}\includegraphics[width=0.49\linewidth]{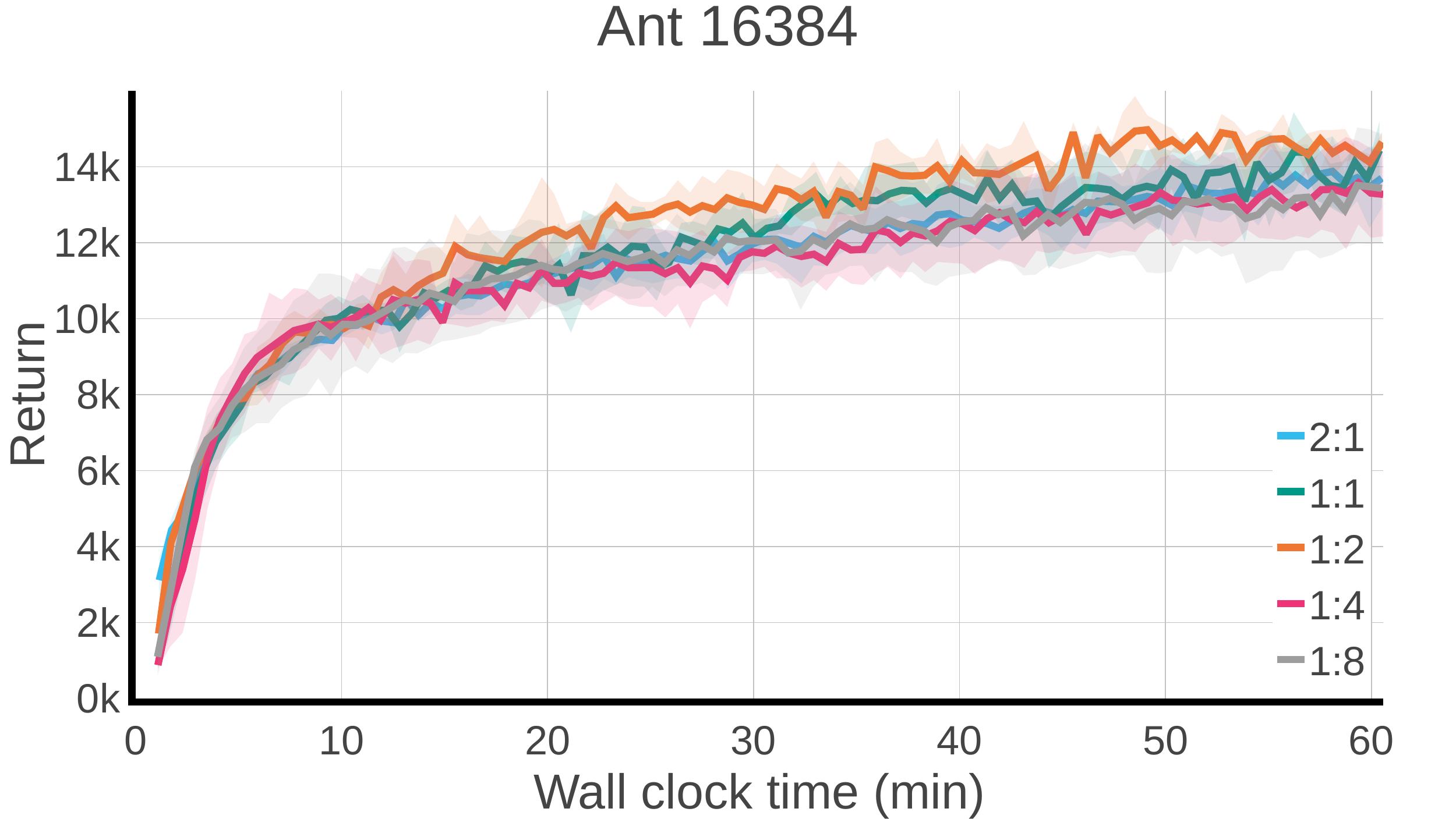}}
     \\
    \subfigure[]{\label{fig:shadow_2048_ac}\includegraphics[width=0.49\linewidth]{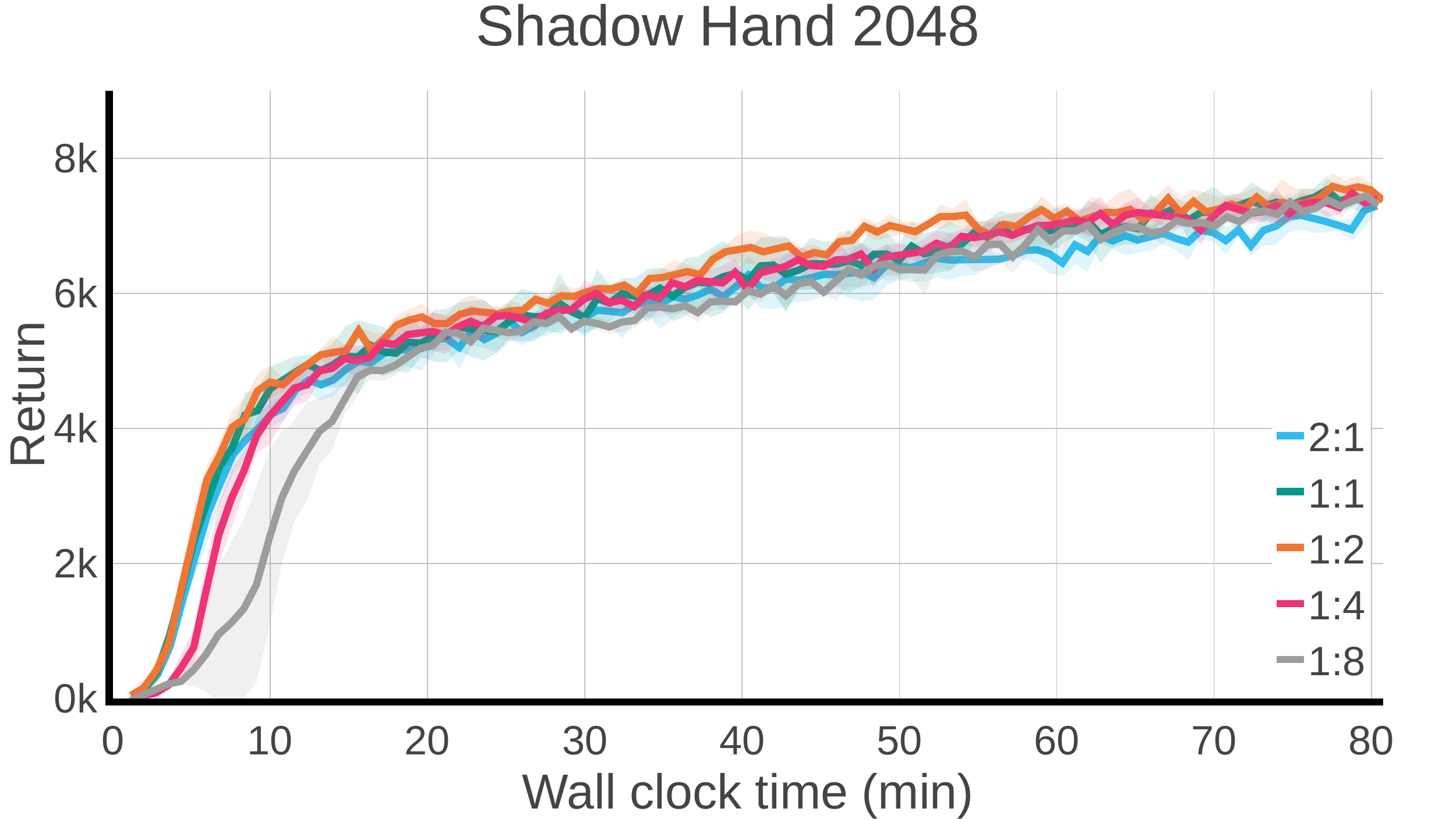}}
    \subfigure[]{\label{fig:shadow_4096_ac}\includegraphics[width=0.49\linewidth]{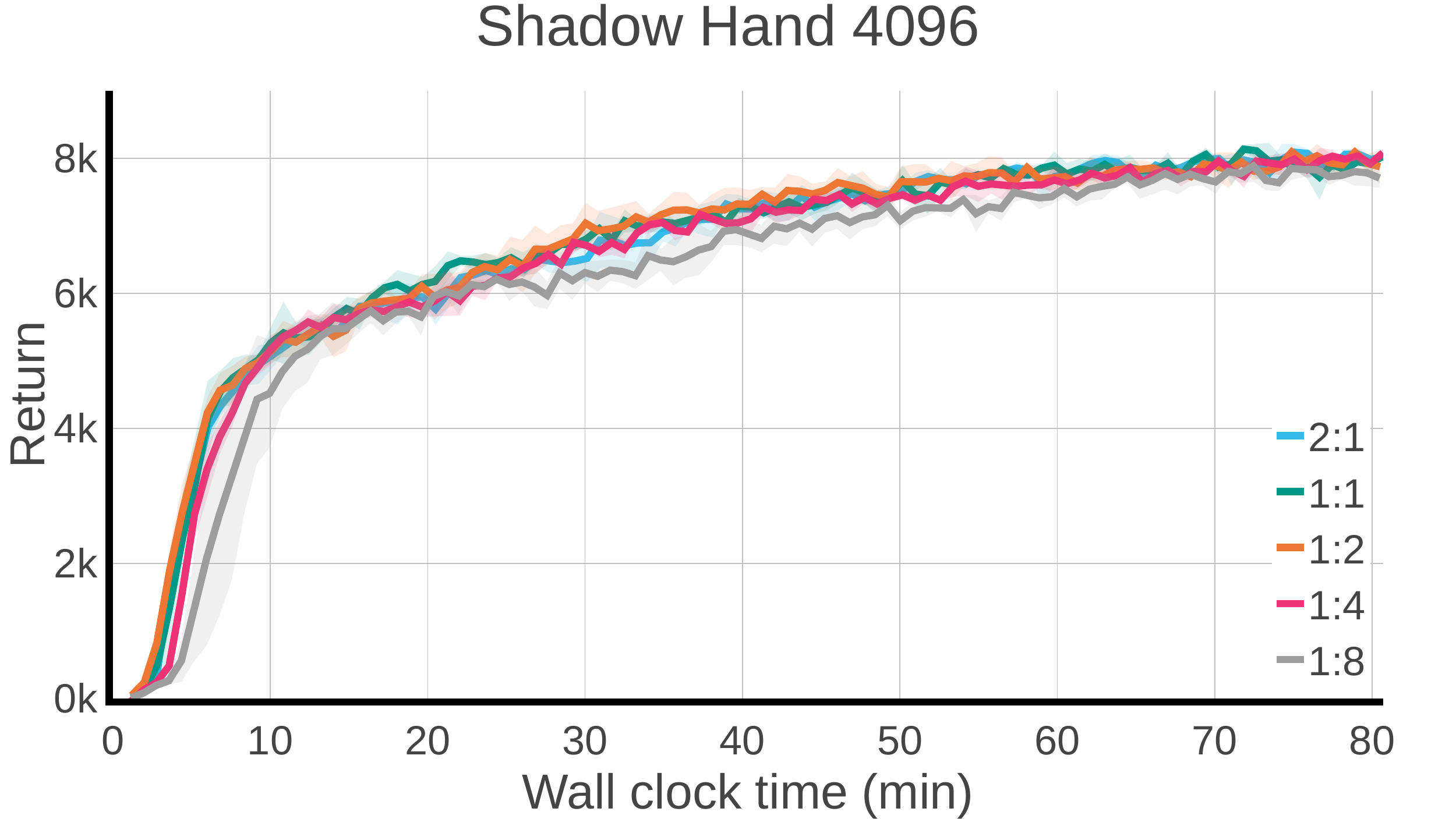}}
\\
    \subfigure[]{\label{fig:shadow_8192_ac}\includegraphics[width=0.49\linewidth]{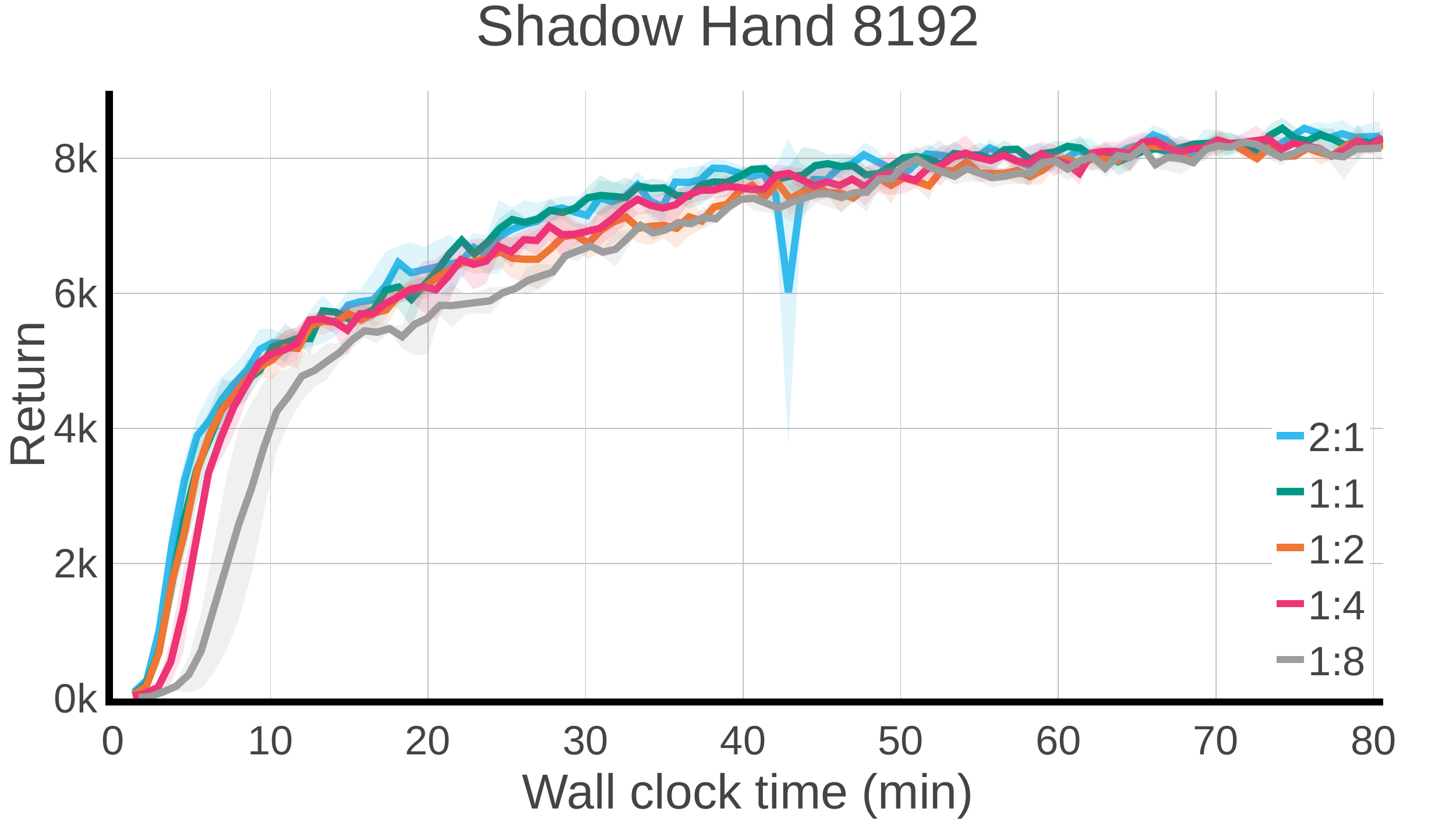}}
    \subfigure[]{\label{fig:shadow_16384_ac}\includegraphics[width=0.49\linewidth]{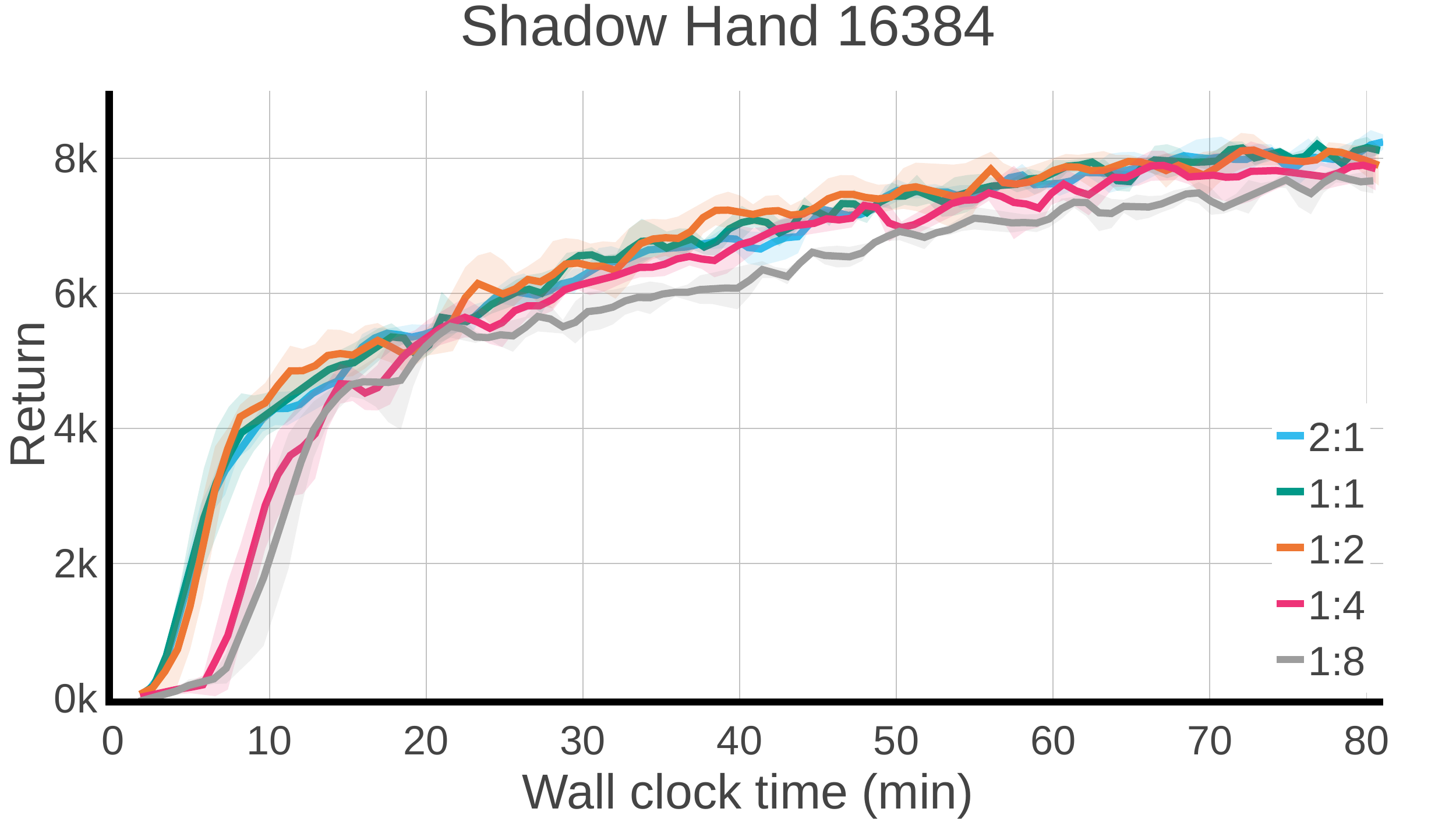}}
    \caption{Learning curves for different $\beta_{p:v}$.}
    \label{fig:app_actor_critic_ratio}
\end{figure}

\begin{figure}[!htb]
    \centering
    \subfigure[]{\label{fig:ant_2048_wc}\includegraphics[width=0.49\linewidth]{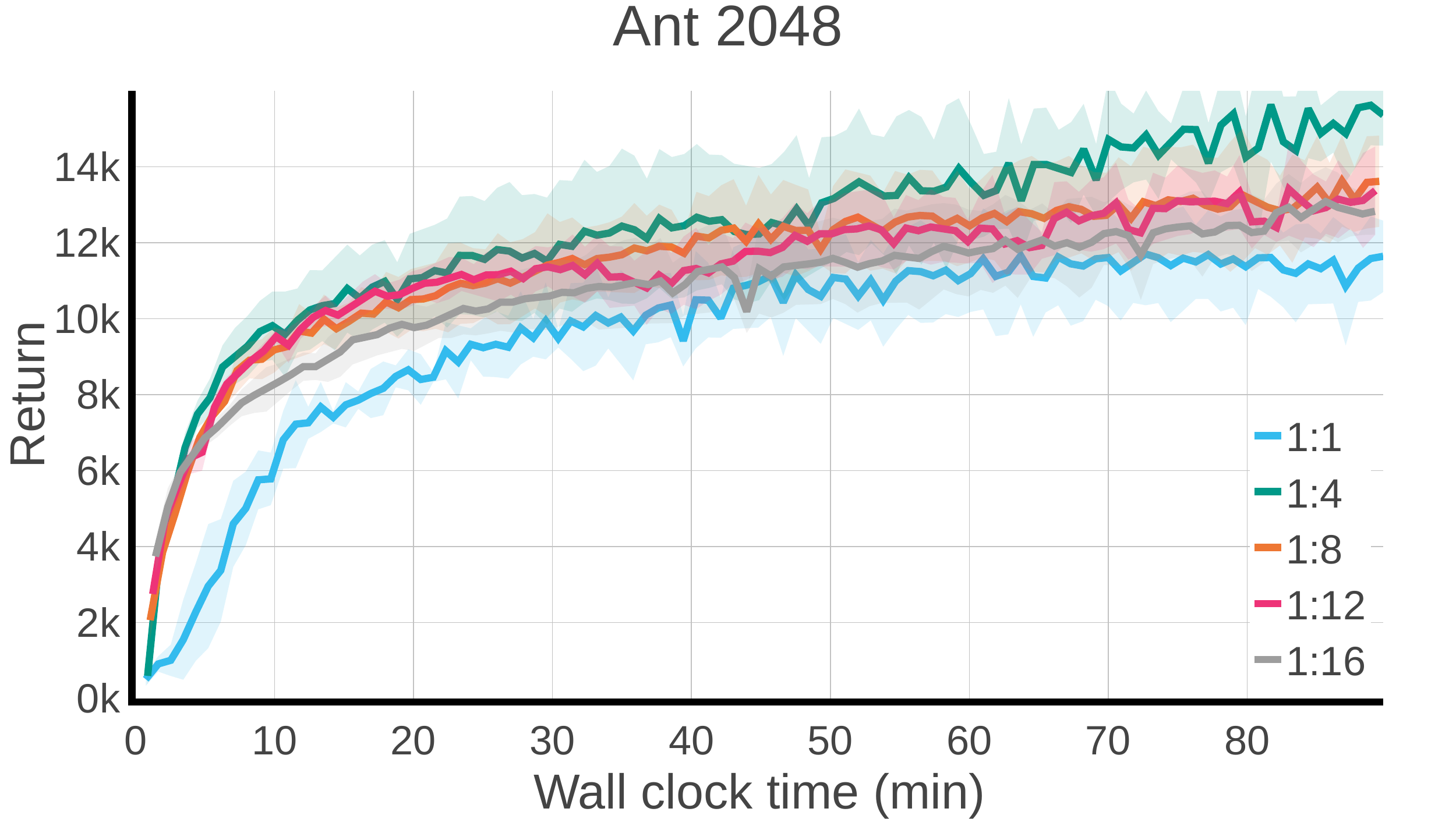}}
    \subfigure[]{\label{fig:ant_4096_wc}\includegraphics[width=0.49\linewidth]{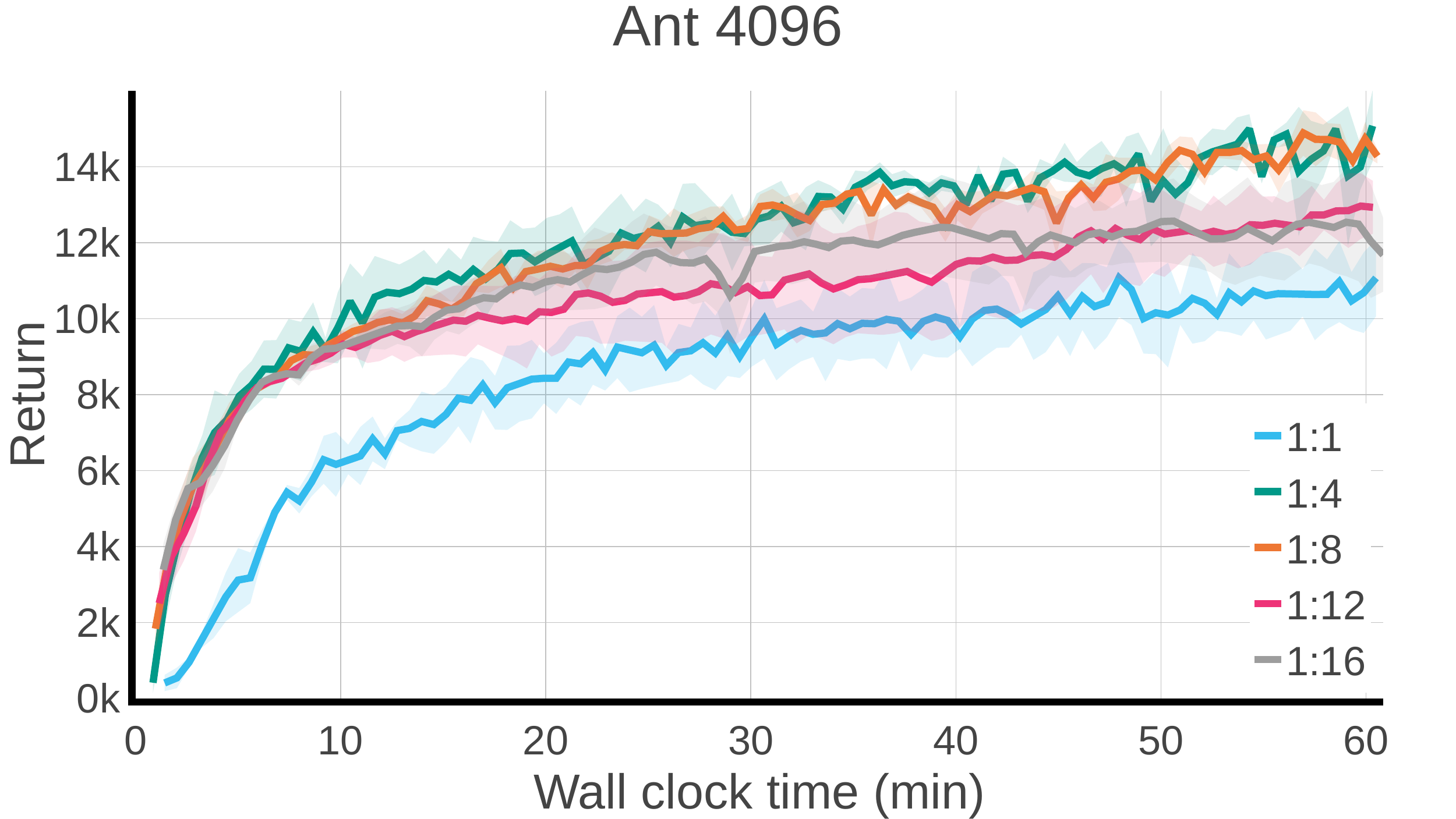}}
     \\
    \subfigure[]{\label{fig:ant_8192_wc}\includegraphics[width=0.49\linewidth]{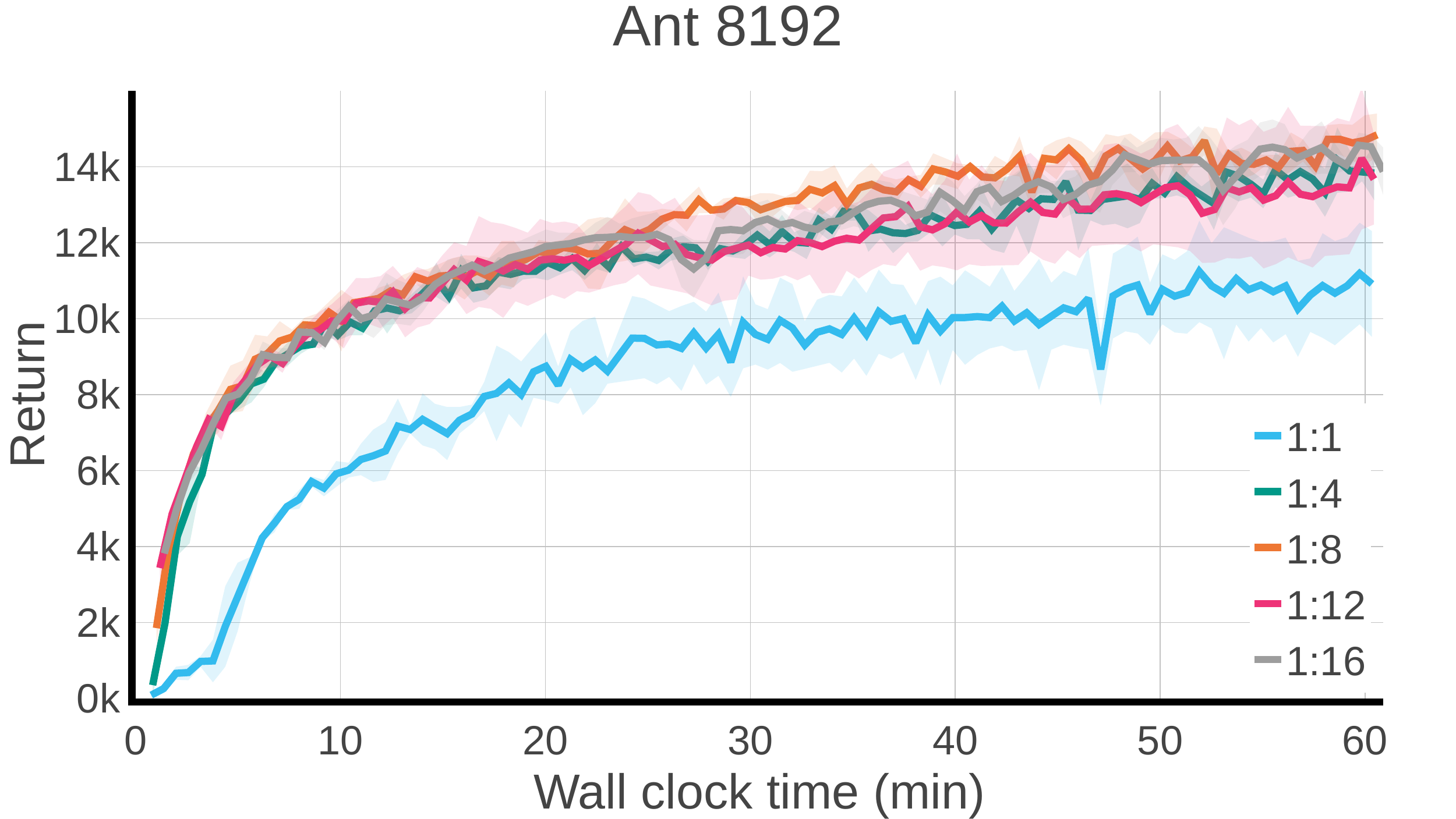}}
    \subfigure[]{\label{fig:ant_16384_wc}\includegraphics[width=0.49\linewidth]{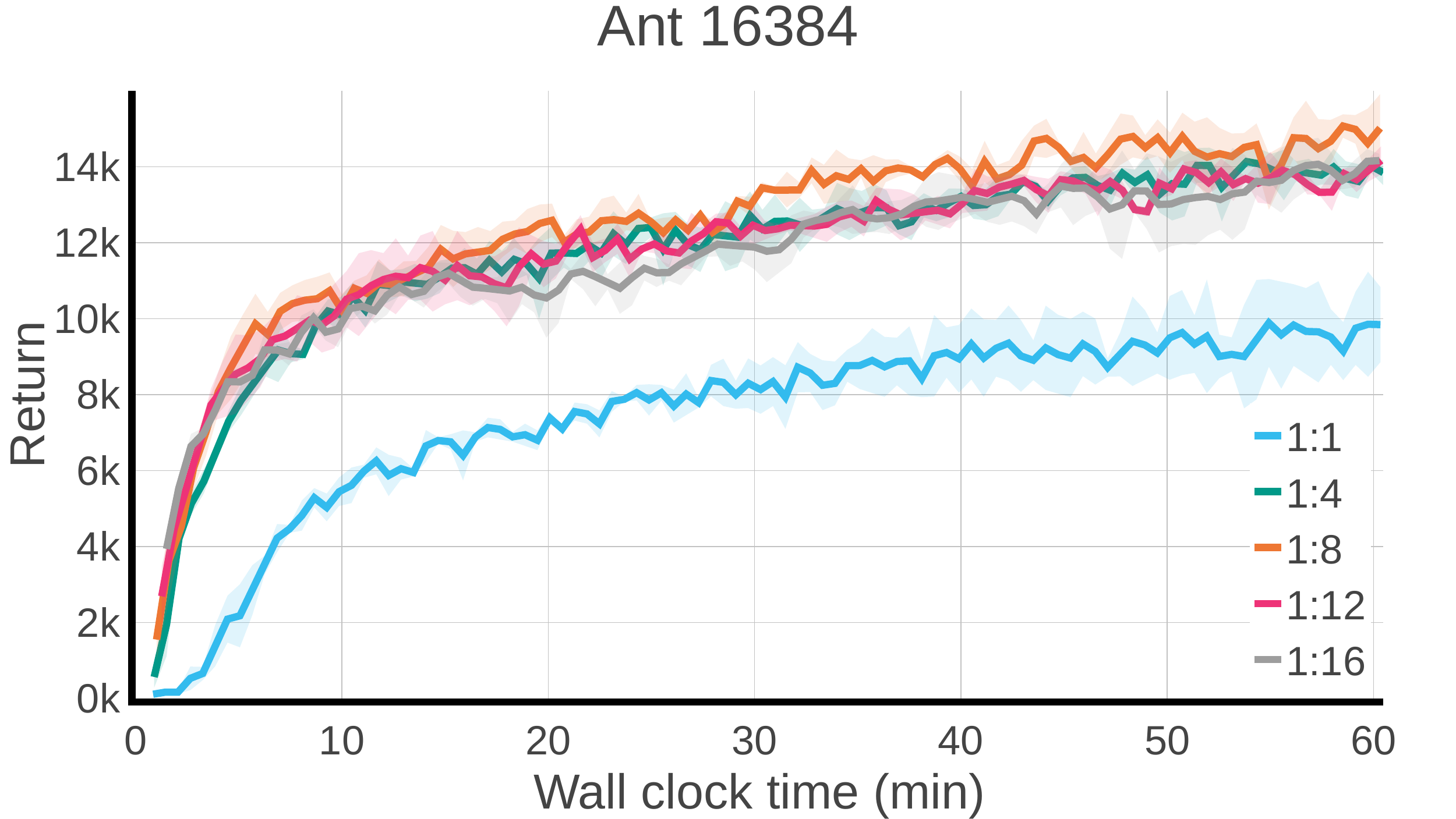}}
     \\
    \subfigure[]{\label{fig:shadow_2048_wc}\includegraphics[width=0.49\linewidth]{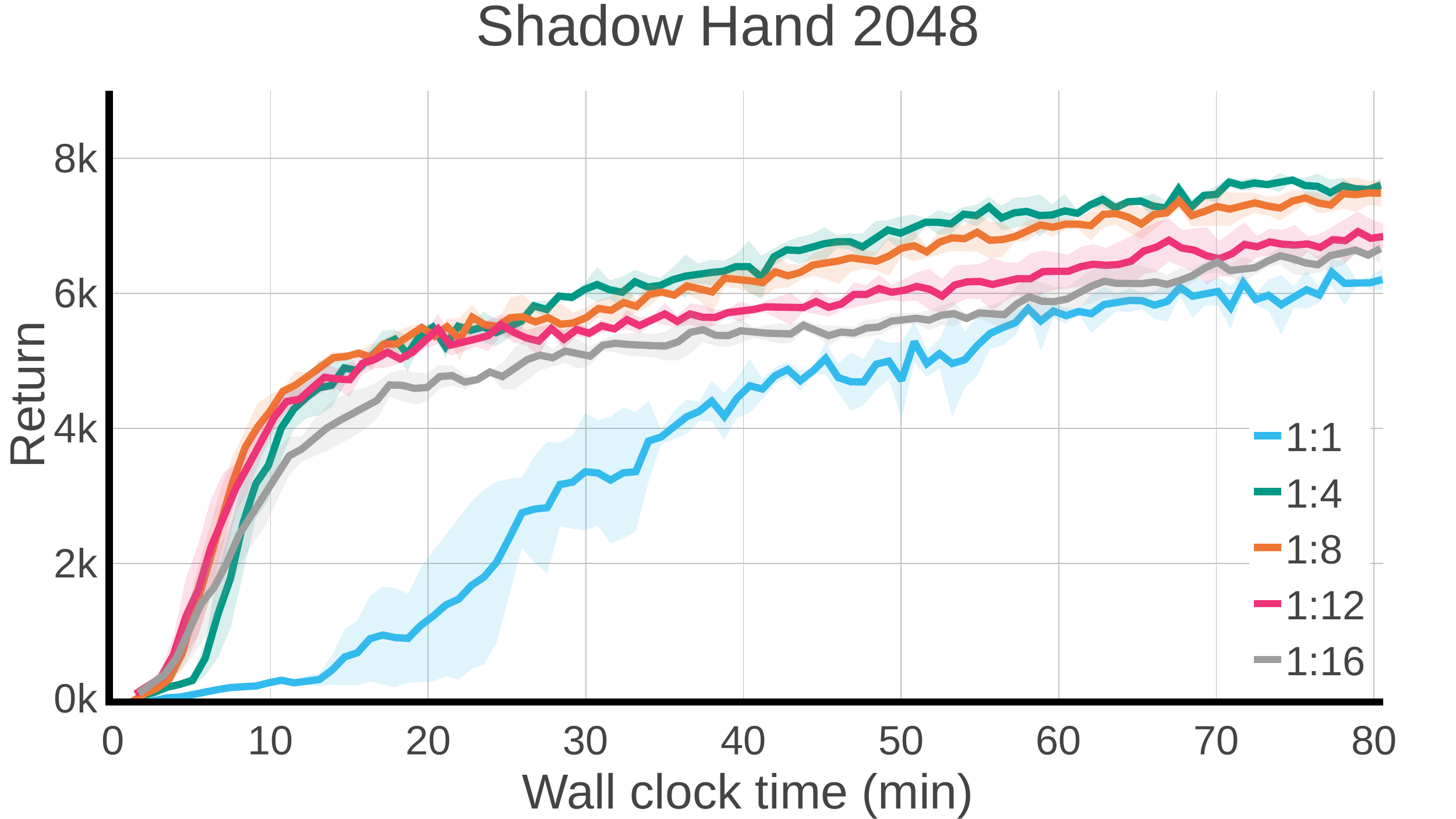}}
    \subfigure[]{\label{fig:shadow_4096_wc}\includegraphics[width=0.49\linewidth]{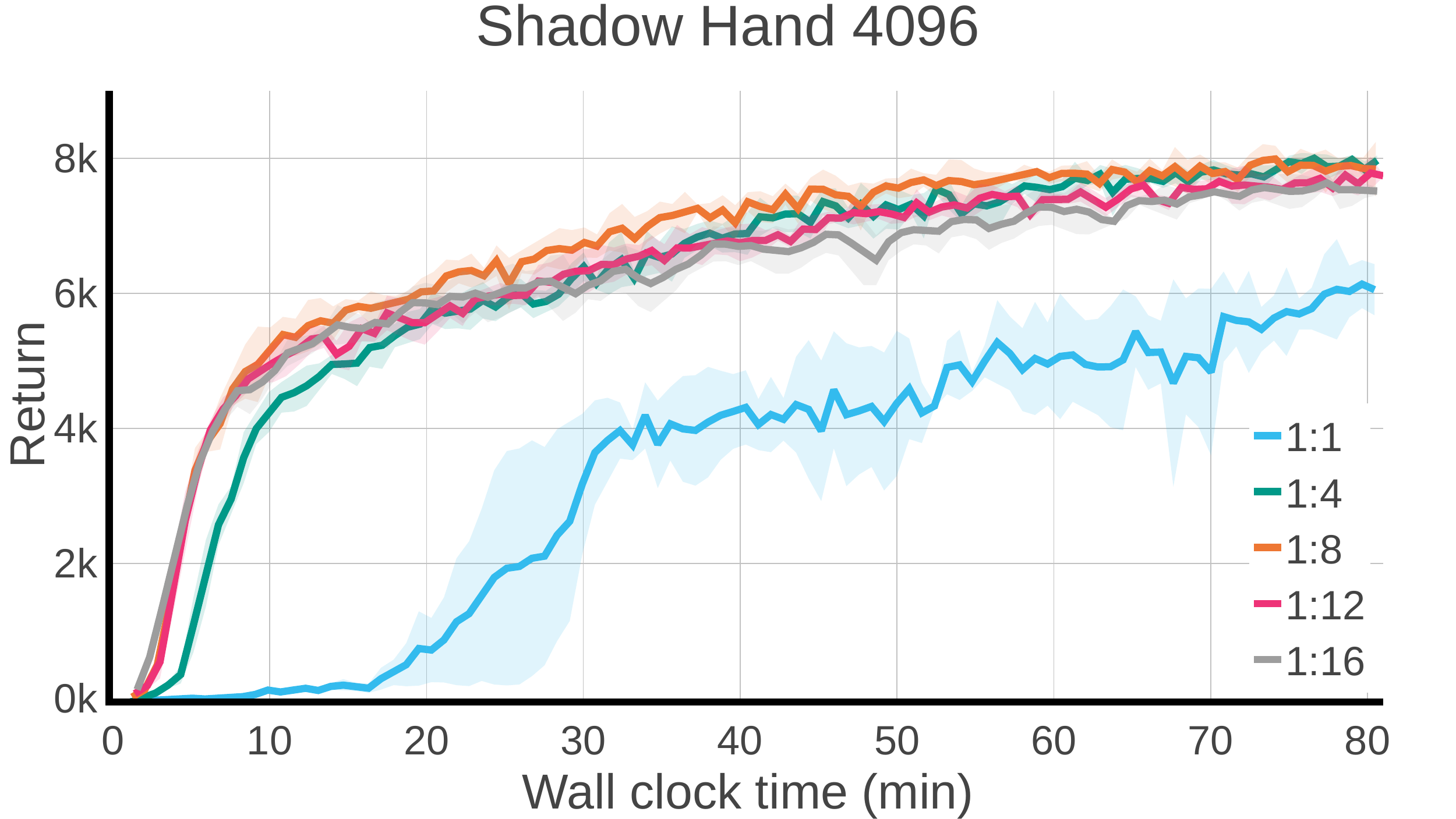}}
\\
    \subfigure[]{\label{fig:shadow_8192_wc}\includegraphics[width=0.49\linewidth]{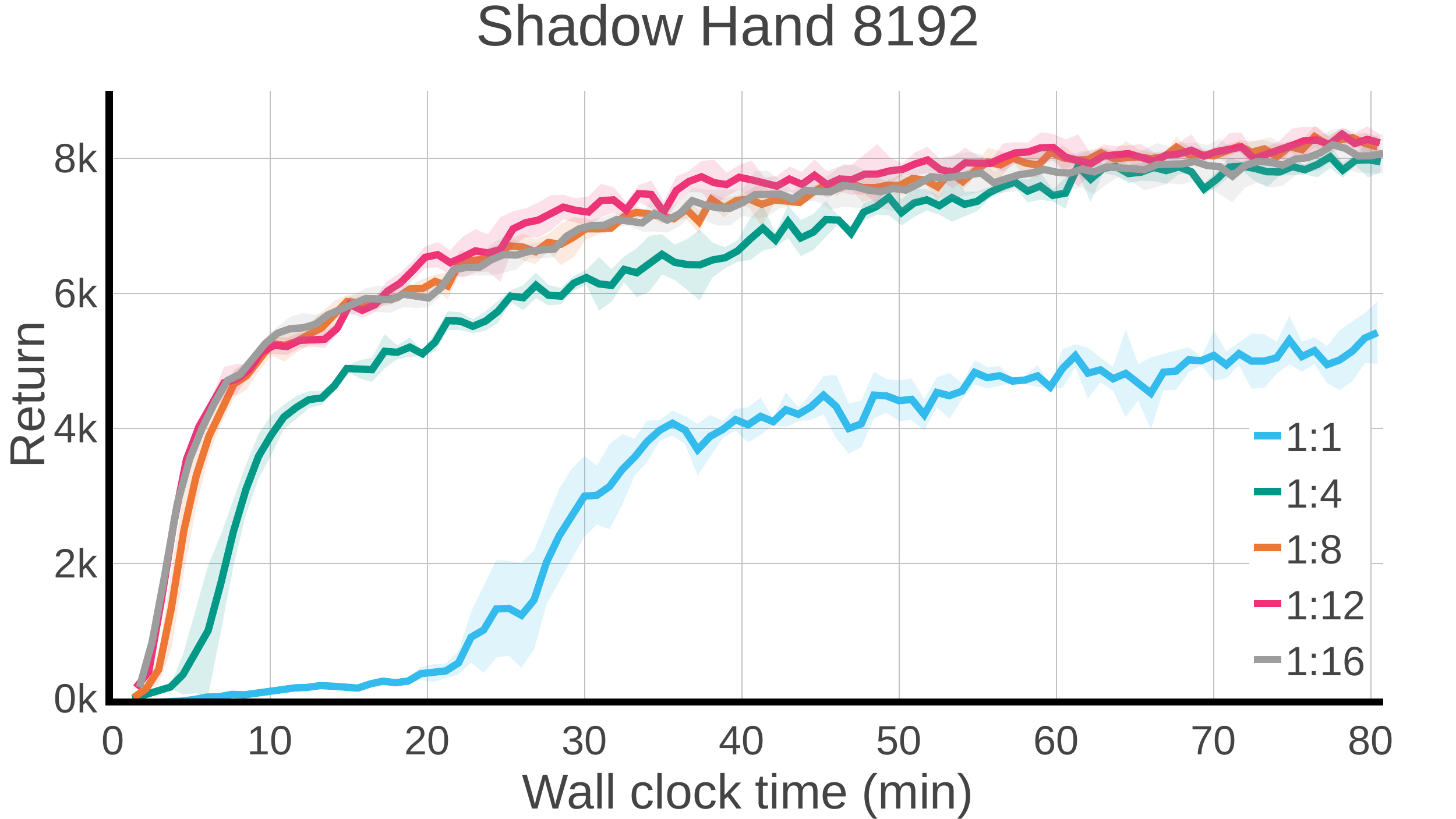}}
    \subfigure[]{\label{fig:shadow_16384_wc}\includegraphics[width=0.49\linewidth]{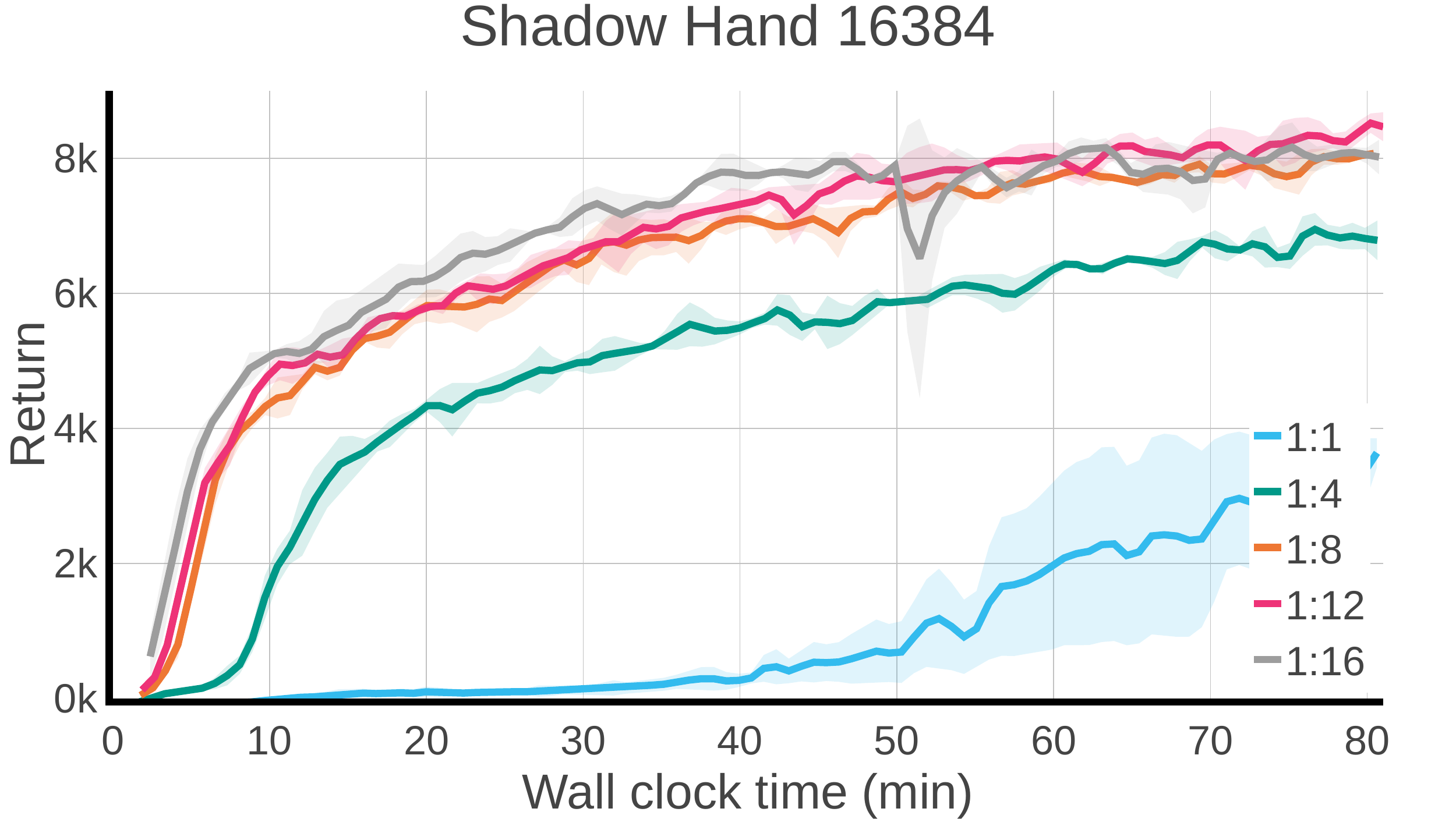}}
    \caption{Learning curves for different $\beta_{a:v}$.}
    \label{fig:app_worker_critic_ratio}
\end{figure}

\figref{fig:app_actor_critic_ratio} shows the complete learning curves with different $\beta_{p:v}$ values and different number of environments. Similarly, \figref{fig:app_worker_critic_ratio} shows the learning curves for different $\beta_{a:v}$.

\paragraph{Comparison of our implementation with RL-games}
In this work, we implemented all the algorithms (PQL and all the baselines) from scratch, as it gives us the most flexibility in exploring different design choices that can affect learning performance. To show that our codebase provides good performance, we compare it against the most commonly used RL codebase used for Isaac Gym, which is RL-games~\citep{rl-games2022}. However, RL-games only support PPO and SAC. Hence, we compare our implementations of PPO and SAC against the ones in RL-games.

\begin{figure}[!htb]
    \centering
    \subfigure[]{\label{fig:ant_rlgames_time}\includegraphics[width=0.49\linewidth]{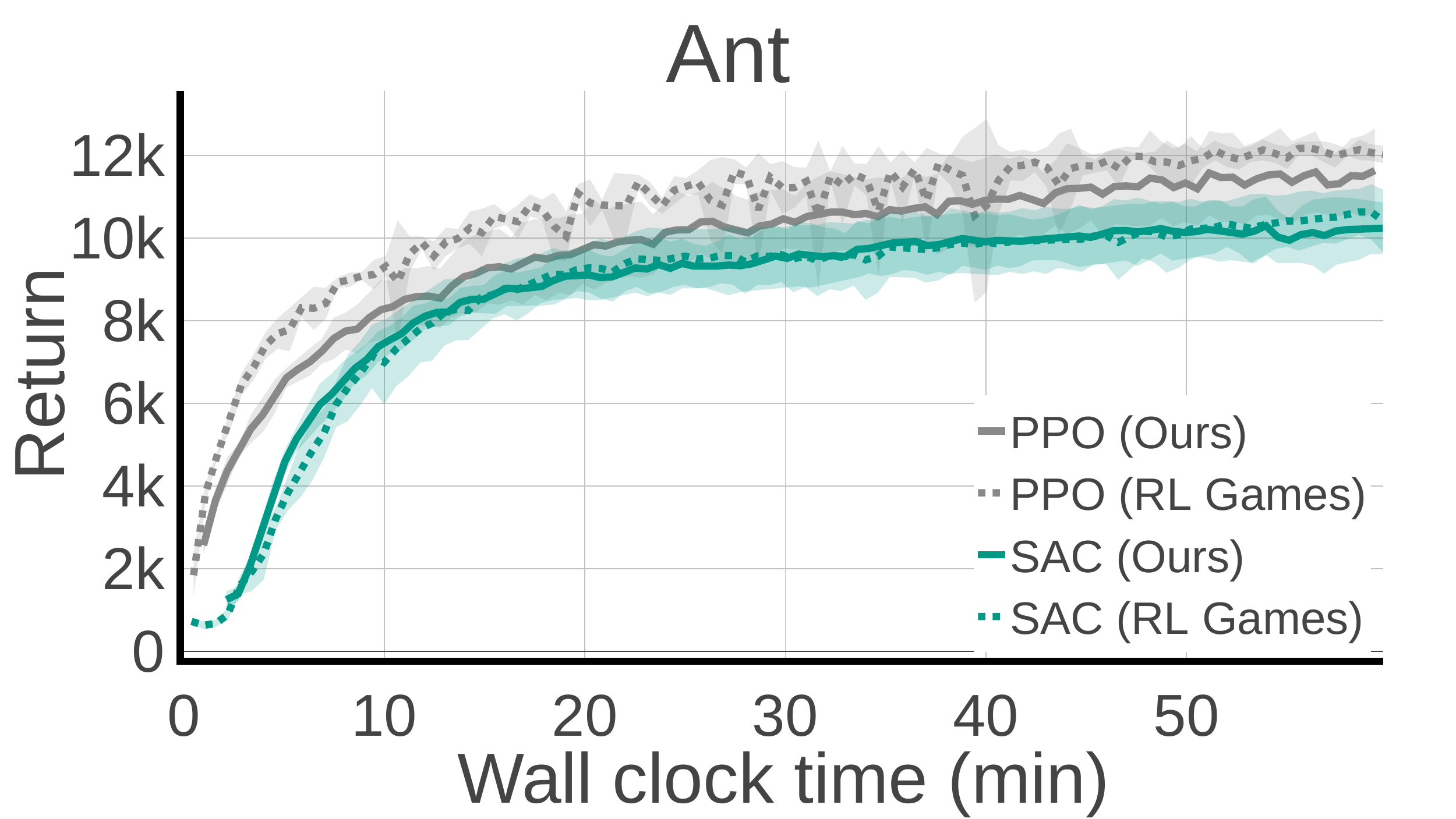}}
    \subfigure[]{\label{fig:shadow_rlgames_time}\includegraphics[width=0.49\linewidth]{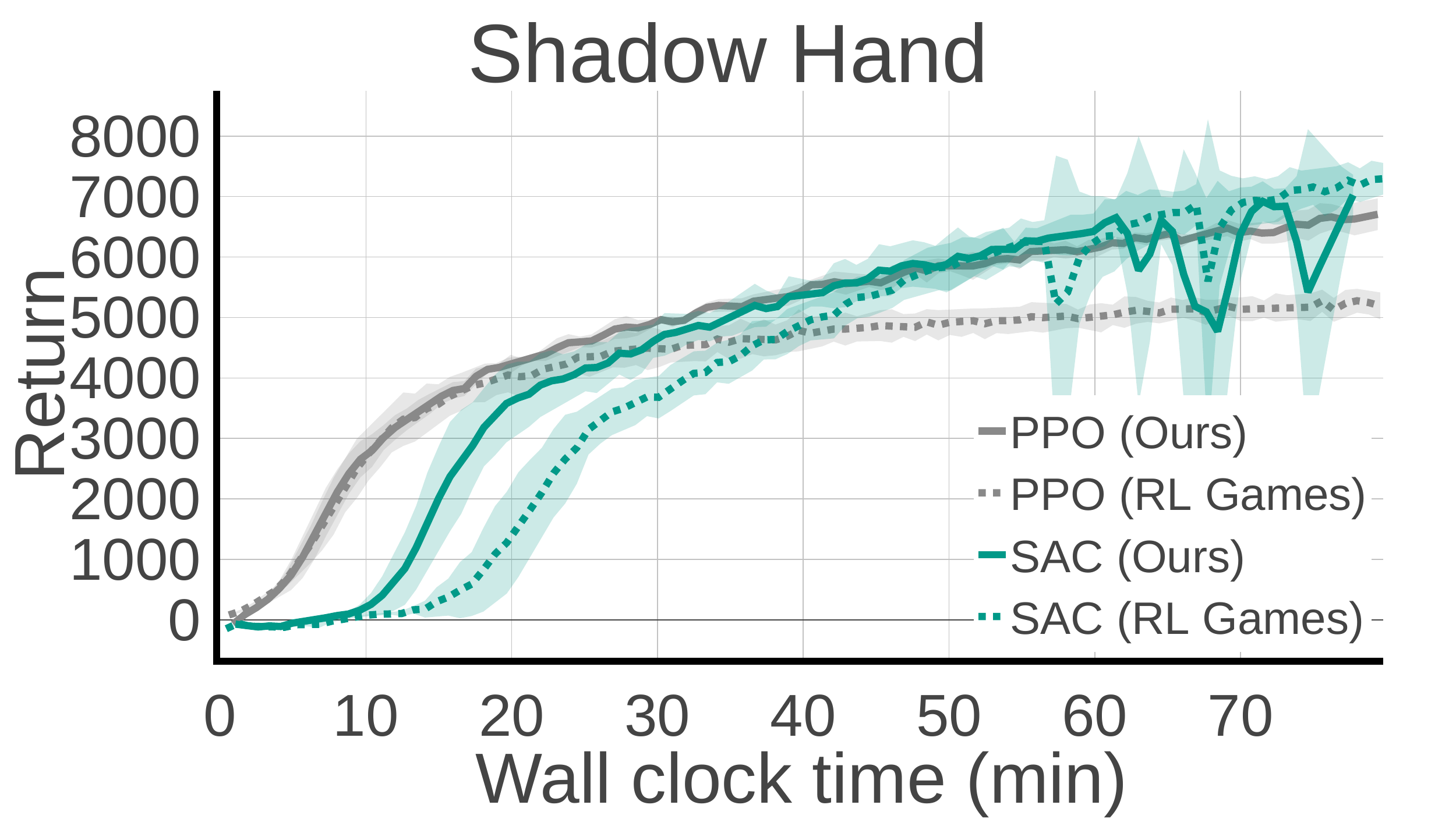}}
    \caption{Comparison between our implementations of PPO and SAC against the ones provided in RL-games. We can see that both codebases provide similar performance, showing that our implementation is good and reliable. On \shadow, our PPO learns even faster and better than the PPO in RL-games.}
    \label{fig:rl-games}
\end{figure}

\paragraph{Distributional critic update}
We investigate how a distributional version of the critic update affects the policy learning performance. Here, we utilize categorical parameterization that outputs a discrete-value distribution defined over a fixed set of atoms $z_i$ \citep{bellemare2017distributional}. We use the same hyper-parameters across the six tasks, where the number of atoms $l = 51$ and the bounds on the support from ($-10, 10$). To make sure the values lie on the support defined by the atoms, we scale the reward into a similar range via different scaling factors shown in \tblref{tbl:reward_scale} and apply the categorical projection operator before minimizing the cross-entropy.

\end{appendices}

\end{document}